\documentclass[journal]{IEEEtran}
\usepackage{cite}
\usepackage{esvect}
\usepackage{mathrsfs}
\usepackage{amsmath}
\allowdisplaybreaks[4]
\newcommand{\norm}[1]{\left\lVert#1\right\rVert}
\usepackage{multirow}
\usepackage{algorithm}
\usepackage{algpseudocode}
\renewcommand{\algorithmicrequire}
{\textbf{Input:}}  
\renewcommand{\algorithmicensure}
{\textbf{Output:}} 
\usepackage{graphicx}
\usepackage{caption}
\usepackage{subcaption}
\usepackage{amsfonts,amssymb}
\usepackage{color}
\usepackage{bm}

\ifCLASSINFOpdf

\else

\fi

\begin{document}

\title{Multi-view Hybrid Embedding:\\ A Divide-and-Conquer Approach}

\author{Jiamiao~Xu$^*$,
        Shujian~Yu$^*$,~\IEEEmembership{Student Member,~IEEE,}
        Xinge~You$^\dagger$,~\IEEEmembership{Senior Member,~IEEE,}\\
        Mengjun~Leng,
        Xiao-Yuan~Jing,
        and~C. L. Philip Chen,~\IEEEmembership{Fellow,~IEEE}
\thanks{$^*$The first two authors contributed equally to this work and should be regarded as co-first authors.}
\thanks{$^\dagger$Corresponding author.}
\thanks{J. Xu and X. You are with the School of Electronic Information
	and Communications, Huazhong University of Science and Technology,
	Wuhan 430074, China.(e-mail: youxg@hust.edu.cn)}
\thanks{S. Yu is with the Department of Electrical and
	Computer Engineering, University of Florida, Gainesville, FL 32611, USA.}
\thanks{M. Leng is with the Department of Computer Science, University of Houston, Houston, TX 77204, USA.}
\thanks{X.-Y. Jing is with the State Key Laboratory of Software Engineering, School of Computer, Wuhan University, China.}
\thanks{C. L. P. Chen is with the Department of Computer and Information Science,
	Faculty of Science and Technology, University of Macau, Macau 99999,
	China; with Dalian Maritime University, Dalian 116026, China; and also with
	the State Key Laboratory of Management and Control for Complex Systems,
	Institute of Automation, Chinese Academy of Sciences, Beijing 100080, China.}
}

\markboth{IEEE TRANSACTIONS ON CYBERNETICS}
{Shell \MakeLowercase{\textit{et al.}}: Bare Demo of IEEEtran.cls for IEEE Journals}

\maketitle

\begin{abstract}
We present a novel cross-view classification algorithm where the gallery and probe data come from different views. A popular approach to tackle this problem is the multi-view subspace learning~(MvSL) that aims to learn a latent subspace shared by multi-view data. Despite promising results obtained on some applications, the performance of existing methods deteriorates dramatically when the multi-view data is sampled from nonlinear manifolds or suffers from heavy outliers. To circumvent this drawback, motivated by the \textit{Divide-and-Conquer} strategy, we propose Multi-view Hybrid Embedding (MvHE), a unique method of dividing the problem of cross-view classification into three subproblems and building one model for each subproblem. Specifically, the first model is designed to remove view discrepancy, whereas the second and third models attempt to discover the intrinsic nonlinear structure and to increase discriminability in intra-view and inter-view samples respectively. The kernel extension is conducted to further boost the representation power of MvHE. Extensive experiments are conducted on four benchmark datasets. Our methods demonstrate overwhelming advantages against the state-of-the-art MvSL based cross-view classification approaches in terms of classification accuracy and robustness.
\end{abstract}

\begin{IEEEkeywords}
cross-view classification, Multi-view Hybrid Embedding~(MvHE), multi-view learning, divide-and-conquer.
\end{IEEEkeywords}

\IEEEpeerreviewmaketitle

\section{Introduction}
\label{sec_intro}
\IEEEPARstart{T}{he} rapid development of imaging technology allows us to capture a single object with different sensors or from different views. Consequently, a single object may have multi-view representations~\cite{li2016multi,xu2013survey,zhao2018multi}. Although more information is provided, there is a tremendous variation and diversity among different views~(i.e., within-class samples from different views might have less similarity than between-class samples from the same view). This unfortunate fact poses an emerging yet challenging problem - the classification of an object when the gallery and probe data come from different views, also known as cross-view classification~\cite{Kan2012,sharma2012generalized,cao2017generalized,jing2017super}.



To tackle this problem, substantial efforts have been made in recent years. These include, for example, the multi-view subspace learning~(MvSL) based methods (e.g.,~\cite{ diethe2008multiview,diethe2010constructing,lin2006inter,lei2009coupled,klare2011matching,chen2010predictive,li2013locally,tzimiropoulos2012subspace,wang2013learning,xu2017learning,chen2012continuum,sim2009simultaneous,moutafis2017overview}) that attempt to remove view discrepancy by projecting all samples into a common subspace. As one of the best-known MvSL based methods, Canonical Correlation Analysis (CCA)~\cite{hotelling1936relations,thompson2005canonical} learns two view-specific mapping functions such that the projected representations from two views are mostly correlated. Multi-view Canonical Correlation Analysis (MCCA)~\cite{nielsen2002multiset,rupnik2010multi} was later developed as an extension of CCA in multi-view scenario. Although the second-order pairwise correlations are maximized, the neglect of discriminant information may severely deteriorate the classification performance of CCA and MCCA. To deal with this problem, more advanced methods, such as the Generalized Multi-view Analysis (GMA) [5] and the Multi-view Modular Discriminant Analysis (MvMDA) [30], were proposed thereafter to render improved classification performance by taking into consideration either intra-view or inter-view discriminant information. Despite promising results obtained on some applications, these methods are only capable of discovering the intrinsic geometric structure of data lying on linear or near-linear manifolds~\cite{zhang2009patch,tenenbaum2000global}, and their performance cannot be guaranteed when the data is non-linearly embedded in high dimensional observation space or suffers from heavy outliers~\cite{davoudi2017dimensionality,belkin2003laplacian,koren2004robust}. 

To circumvent the linear limitation, a straightforward way is to extend these methods with the famed kernel trick. However, it is trivial to design a favorable kernel and it is inefficient to deal with the out-of-sample problem~\cite{arias2007connecting}. Moreover, to the best of our knowledge, majority of the current kernel extensions, such as Kernel Canonical Correlation Analysis~(KCCA)~\cite{akaho2006kernel,melzer2001nonlinear}, can only deal with two-view classification. On the other hand, several recent methods~(e.g., Multi-view Deep Network (MvDN)~\cite{kan2016multideep}, Deep Canonical Correlation Analysis (DCCA)~\cite{andrew2013deep,wang2015unsupervised}, Multimodal Deep Autoencoder~\cite{ngiam2011multimodal}) suggest using deep neural networks~(DNNs) to eliminate the nonlinear discrepancy between pairwise views. Same as other DNNs that have been widely used in different machine learning and computer vision tasks, it remains a question on the optimal selection of network topology and the network cannot be trained very well, theoretically and practically, when the training data is insufficient. 

This work also deals with nonlinear challenge. However, different from aforementioned work that is either incorporated into a kernel framework in a brute force manner or built upon an over-complicated DNN that is hard to train, our general idea is to enforce the latent representation to preserve the local geometry of intra-view and inter-view samples as much as possible, thus significantly improving its representation power. Albeit its simplicity, it is difficult to implement this idea in cross-view classification. This is because traditional local geometry preservation methods~(e.g., Locally Linear Embedding~(LLE)~\cite{roweis2000nonlinear} and Locality Preserving Projections~(LPP)~\cite{he2004locality}) require the manifolds are locally linear or smooth~\cite{silva2003global}. Unfortunately, this condition is not met in multi-view scenario, especially considering the fact that discontinuity could easily occur at the junction of pairwise views~\cite{van2009dimensionality}. 

To this end, we propose a novel cross-view classification algorithm with the \textit{Divide-and-Conquer} strategy \cite[chapter 4]{cormen2009introduction}. Instead of building a sophisticated ``unified" model, we partition the problem of cross-view classification into three subproblems and build one model to solve each subproblem. 
Specifically, the first model (i.e., low-dimensional embedding of paired samples or \textit{LE-paired}) aims to remove view discrepancy, whereas the second model (i.e., local discriminant embedding of intra-view samples or \textit{LDE-intra}) attempts to discover nonlinear embedding structure and to increase discriminability in intra-view samples~(i.e., samples from the same view). By contrast, the third model (i.e., local discriminant embedding of inter-view samples or \textit{LDE-inter}) attempts to discover nonlinear embedding structure and to increase discriminability in inter-view samples (i.e., samples from different views). We term the combined model that integrates \textit{LE-paired}, \textit{LDE-intra} and \textit{LDE-inter} Multi-view Hybrid Embedding~(MvHE).

To summarize, our main contributions are threefold:

1) Motivated by the \textit{Divide-and-Conquer} strategy, a novel method, named MvHE, is proposed for cross-view classification. Three subproblems associated with cross-view classification are pinpointed and three models are developed to solve each of them.

2) The part optimization and whole alignment framework~\cite{zhang2009patch} that generalizes majority of the prevalent single-view manifold learning algorithms has been extended to multi-view scenario, thus enabling us to precisely preserve the local geometry of inter-view samples.

3) Experiments on four benchmark datasets demonstrate that our method can effectively discover the intrinsic structure of multi-view data lying on nonlinear manifolds. Moreover, compared with previous MvSL based counterparts, our method is less sensitive to outliers.

The rest of this paper is organized as follows. Sect.~\ref{sec_related} introduces the related work. In Sect.~\ref{sec_proposed}, we describe MvHE and its optimization in detail. The kernel extension and complexity analysis of MvHE are also conducted. Experimental results on four benchmark datasets are presented in Sect.~\ref{sec_experiments}. Finally, Sect.~\ref{sec_conclusion} concludes this paper.

\section{Related Work and Preliminary Knowledge}
\label{sec_related}
In this section, the key notations used in this paper are summarized and the most relevant MvSL based cross-view classification methods are briefly reviewed. We also present the basic knowledge on part optimization and whole alignment modules, initiated in Discriminative Locality Alignment~(DLA)~\cite{zhang2009patch}, to make interested readers more familiar with our method that will be elaborated in Sect.~\ref{sec_proposed}.

\subsection{Multi-view Subspace Learning based Approaches}
Suppose we are given $n$-view data $\mathcal{X}=\{\bm{X}_1,\bm{X}_2,\ldots,\bm{X}_n\}$ $\left(n\geq2\right)$, where $\bm{X}_v\in\mathbb{R}^{d_v\times{m}}$ denotes the data matrix from the $v$-th view. Here, $d_v$ is the feature dimensionality and $m$ is the number of samples. We then suppose $\bm{x}_v^i\in\mathcal{X}$ is a sample from the $i$-th object under the $v$-th view and $\bm{x}_v^{ic}$ denotes that the sample $\bm{x}_v^i$ is from the $c$-th class, where $c\in\{1,2,\ldots,C\}$ and $C$ is the number of classes. Let $\{\bm{x}_1^k,\bm{x}_2^k,\ldots,\bm{x}_n^k\}$ refer to the paired samples from the $k$-th object.

MvSL based approaches aim to learn $n$ mapping functions $\{\bm{W}_v\}_{v=1}^n$, one for each view, to project data into a latent subspace $\mathcal{Y}$, where $\bm{W}_v\in\mathbb{R}^{d_v\times{d}}$ is the mapping function of the $v$-th view. For ease of presentation, let $\bm{\mu}_v^c$ denote the mean of all samples of the $c$-th class under the $v$-th view in $\mathcal{Y}$, $\bm{\mu}^c$ denote the mean of all samples of the $c$-th class over all views in $\mathcal{Y}$, $\bm{\mu}$ denote the mean of all samples over all views in $\mathcal{Y}$, $N_v^c$ denote the number of samples from the $v$-th view of the $c$-th class and $N^c$ denote the number of samples of the $c$-th class over all views. Also let $\mathrm{tr}\left(\cdot\right)$ denote the trace operator, $\textbf{I}_p$ denote a $p$-dimensional identity matrix and $\bm{e}$ indicate a column vector with all elements equal to one. 

We analyze different MvSL based cross-view classification methods below.




\subsubsection{CCA, KCCA and MCCA}
\label{subsec_cca}
Canonical Correlation Analysis~(CCA)~\cite{hotelling1936relations,thompson2005canonical,gong2013multi} aims to maximize the correlation between $\bm{W}_1^\mathrm{T}\bm{X}_1$ and $\bm{W}_2^\mathrm{T}\bm{X}_2$ with the following objective:

\begin{align}\label{cca}
&\max_{\bm{W}_1, \bm{W}_2}\bm{W}_1^\mathrm{T}\bm{X}_1\bm{X}_2^\mathrm{T}\bm{W}_2\nonumber\\
&\text{s.t.}~\bm{W}_1^\mathrm{T}\bm{X}_1\bm{X}_1^\mathrm{T}\bm{W}_1 = 1,~\bm{W}_2^\mathrm{T}\bm{X}_2\bm{X}_2^\mathrm{T}\bm{W}_2 = 1.
\end{align}


Kernel Canonical Correlation Analysis~(KCCA)~\cite{akaho2006kernel} extends CCA to a nonlinear model with the famed kernel trick. The objective of KCCA can be expressed using the Representer Theorem~\cite{vapnik2013nature}: 

\begin{align}\label{kcca}
&\max_{\bm{A}_1, \bm{A}_2}\bm{A}_1^\mathrm{T}\bm{K}_1\bm{K}_2\bm{A}_2\nonumber\\
&\text{s.t.}~\bm{A}_1^\mathrm{T}\bm{K}_1\bm{K}_1\bm{A}_1 = 1,~\bm{A}_2^\mathrm{T}\bm{K}_2\bm{K}_2\bm{A}_2 = 1,
\end{align}
where $\bm{K}_1$ and $\bm{K}_2$ are kernel matrices with respect to $\bm{X}_1$ and $\bm{X}_2$, $\bm{A}_1$ and $\bm{A}_2$ are atom matrices in their corresponding view. Similar to CCA, KCCA is limited to two-view data.

\label{subsec_mcca}
Multi-view Canonical Correlation Analysis (MCCA)~\cite{nielsen2002multiset,rupnik2010multi} generalizes CCA to multi-view scenario by maximizing the sum of pairwise correlations between any two views:
\begin{align}\label{mcca}
&\max_{\bm{W}_1, \bm{W}_2,\ldots,\bm{W}_n}\sum_{i<j}\bm{W}_i^\mathrm{T}\bm{X}_i\bm{X}_j^\mathrm{T}\bm{W}_j\nonumber\\
&\text{s.t.}~\bm{W}_i^\mathrm{T}\bm{X}_i\bm{X}_i^\mathrm{T}\bm{W}_i = 1, \quad i=1,2,\ldots,n.
\end{align}


As baseline methods, CCA, KCCA and MCCA suffer from the neglect of any discriminant information.

\subsubsection{PLS}
Partial Least Squares~(PLS)~\cite{rosipal2005overview,sharma2011bypassing} intends to map two-view data to a common subspace in which the convariance between two latent representations is maximized:
\begin{align}\label{pls}
&\max_{\bm{W}_1, \bm{W}_2}\bm{W}_1^\mathrm{T}\bm{X}_1\bm{X}_2^\mathrm{T}\bm{W}_2\nonumber\\
&\text{s.t.}~\bm{W}_1^\mathrm{T}\bm{W}_1 = 1,~\bm{W}_2^\mathrm{T}\bm{W}_2 = 1.
\end{align}
Same as CCA and its variants, PLS neglects class labels.

\subsubsection{GMA and MULDA}
Generalized Multi-view Analysis~(GMA)~\cite{sharma2012generalized} offers an advanced avenue to generalize CCA to supervised algorithm by taking into consideration intra-view discriminant information using the following objective:
\begin{align}
&\max_{\bm{W}_1, \bm{W}_2,\ldots,\bm{W}_n}\sum_{i=1}^n\mu_{i}\bm{W}_i^\mathrm{T}\bm{A}_i\bm{W}_i+\sum_{i<j}2\lambda_{ij}\bm{W}_i^\mathrm{T}\bm{X}_i\bm{X}_j^\mathrm{T}\bm{W}_j\nonumber\\
&\text{s.t.}\sum_{i}\gamma_i\bm{W}_i^\mathrm{T}\bm{B}_i\bm{W}_i=1,
\end{align}
where $\bm{A}_i$ and $\bm{B}_i$ are within-class and between-class scatter matrices of the $i$-th view respectively. 

Multi-view Uncorrelated Linear Discriminant Analysis (MULDA)~\cite{sun2016multiview} was later proposed to learn discriminant features with minimal redundancy by embedding uncorrelated LDA~(ULDA)~\cite{jin2001face} into CCA framework:

\begin{align}
&\max_{\bm{W}_1, \bm{W}_2,\ldots,\bm{W}_n}\sum_{i=1}^n\mu_{i}\bm{W}_{ir}^\mathrm{T}\bm{A}_i\bm{W}_{ir}+\sum_{i<j}2\lambda_{ij}\bm{W}_{ir}^\mathrm{T}\bm{X}_i\bm{X}_j^\mathrm{T}\bm{W}_{jr}\nonumber\\
&\text{s.t.}\sum_{i}\gamma_i\bm{W}_{ir}^\mathrm{T}\bm{B}_{i}\bm{W}_{ir}=1,\quad\bm{W}_{ir}^\mathrm{T}\bm{B}_{i}\bm{W}_{it}=0,\nonumber\\ 
&\quad \quad i=1,2,\ldots,n; \quad r,t=1,2,\ldots,d.
\end{align}

\subsubsection{MvDA}
Multi-view Discriminant Analysis~(MvDA)~\cite{Kan2012} is the multi-view version of Linear Discriminant Analysis~(LDA)~\cite{belhumeur1997eigenfaces}. It maximizes the ratio of the determinant of the between-class scatter matrix and that of the within-class scatter matrix:
\begin{align}
\left(\bm{W}_1^*,\bm{W}_2^*,\ldots,\bm{W}_n^*\right) = \mathop{\arg\max}_{\bm{W}_1,\bm{W}_2,\dots,\bm{W}_n}\frac{\mathrm{tr}\left(\bm{S}_B\right)}{\mathrm{tr}\left(\bm{S}_W\right)},
\end{align}
where the between-class scatter matrix $\bm{S}_B$ and the within-class scatter matrix $\bm{S}_W$ are given by:
\begin{align}\label{mvda_scatter}
&\bm{S}_B = \sum_{c=1}^CN^c\left(\bm{\mu}^c-\bm{\mu}\right)\left(\bm{\mu}^c-\bm{\mu}\right)^\mathrm{T},\nonumber\\
&\bm{S}_W = \sum_{v=1}^n\sum_{c=1}^C\sum_{i=1}^{N_{v}^c}(\bm{W}_v^\mathrm{T}\bm{x}_v^{ic}-\bm{\mu}^c)(\bm{W}_v^\mathrm{T}\bm{x}_v^{ic}-\bm{\mu}^c)^\mathrm{T}.
\end{align}

\subsubsection{MvMDA}
A similar method to MvDA is the recently proposed Multi-view Modular Discriminant Analysis (MvMDA)~\cite{cao2017generalized} that aims to separate various class centers across different views. The objective of MvMDA can be formulated as:
\begin{align}
\left(\bm{W}_1^*,\bm{W}_2^*,\ldots,\bm{W}_n^*\right)\!=\!\mathop{\arg\max}_{\bm{W}_1, \bm{W}_2,\ldots,\bm{W}_n}\frac{\mathrm{tr}\left(\bm{S}_B\right)}{\mathrm{tr}\left(\bm{S}_W\right)},
\end{align}
where the between-class scatter matrix $\bm{S}_B$ and the within-class scatter matrix $\bm{S}_W$ are given by:
\begin{align}\label{MvMDA_scatter}
&\bm{S}_B = \sum_{i=1}^n\sum_{j=1}^n\sum_{p=1}^C\sum_{q=1}^C\left(\bm{\mu}_i^p-\bm{\mu}_i^q\right)\left(\bm{\mu}_j^p-\bm{\mu}_j^q\right)^\mathrm{T},\nonumber\\
&\bm{S}_W = \sum_{v=1}^n\sum_{c=1}^C\sum_{i=1}^{N_v^c}\left(\bm{W}_v^\mathrm{T}\bm{x}_v^{ic}-\bm{\mu}_v^c\right)\left(\bm{W}_v^\mathrm{T}\bm{x}_v^{ic}-\bm{\mu}_v^c\right)^\mathrm{T}.
\end{align}

Different from GMA, MvDA and MvMDA incorporate inter-view discriminant information. However, all these methods are incapable of discovering the nonlinear manifolds embedded in multi-view data due to the global essence~\cite{zhang2009patch}.

\subsection{Part Optimization and Whole Alignment}
\label{global_and_local}


The \textit{LDE-intra} and \textit{LDE-inter} are expected to be able to precisely preserve local discriminant information in either intra-view or inter-view samples. A promising solution is the part optimization and whole alignment framework~\cite{zhang2009patch} that generalizes majority of the prevalent single-view manifold learning methods (e.g., LDA, LPP~\cite{he2004locality} and DLA~\cite{zhang2009patch}) by first constructing a patch for each sample and then optimizing over the sum of all patches. Despite its strong representation power and elegant flexibility, one should note that part optimization and whole alignment cannot be directly applied to multi-view scenario. Before elaborating our solutions in Sect.~\ref{sec_proposed}, we first introduce part optimization and whole alignment framework below for completeness.

\begin{figure*}[t]
    \centering
    \begin{minipage}{3cm}
		\includegraphics[width=3cm]{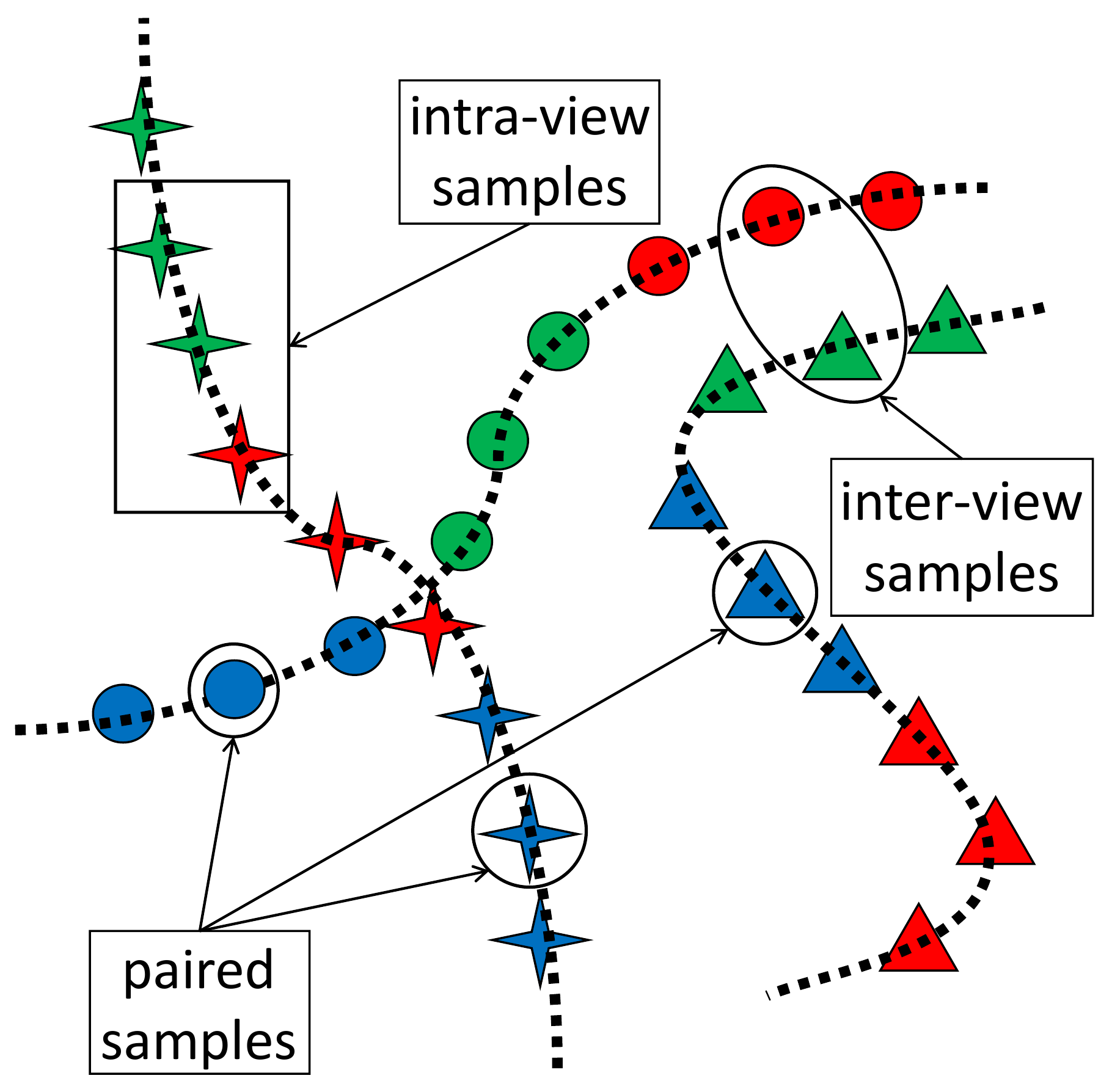}
		\subcaption{\footnotesize{original data}}
	\end{minipage}\hspace{0.3cm}
	\begin{minipage}{3cm}
		\includegraphics[width=3cm]{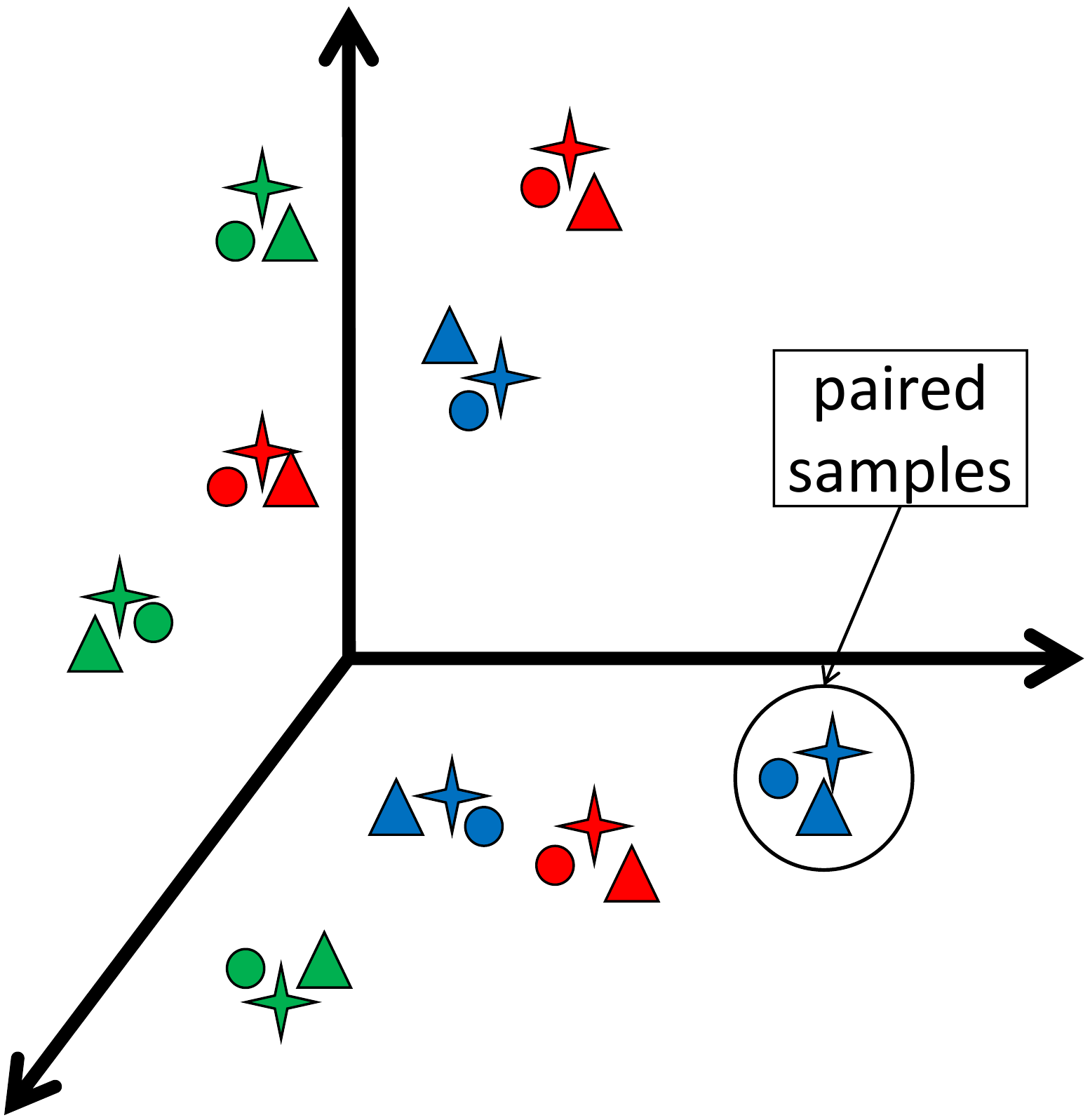}
		\subcaption{\footnotesize{\textit{LE-paired} embedding}}
	\end{minipage}\hspace{0.3cm}
	\begin{minipage}{3cm}
		\includegraphics[width=3cm]{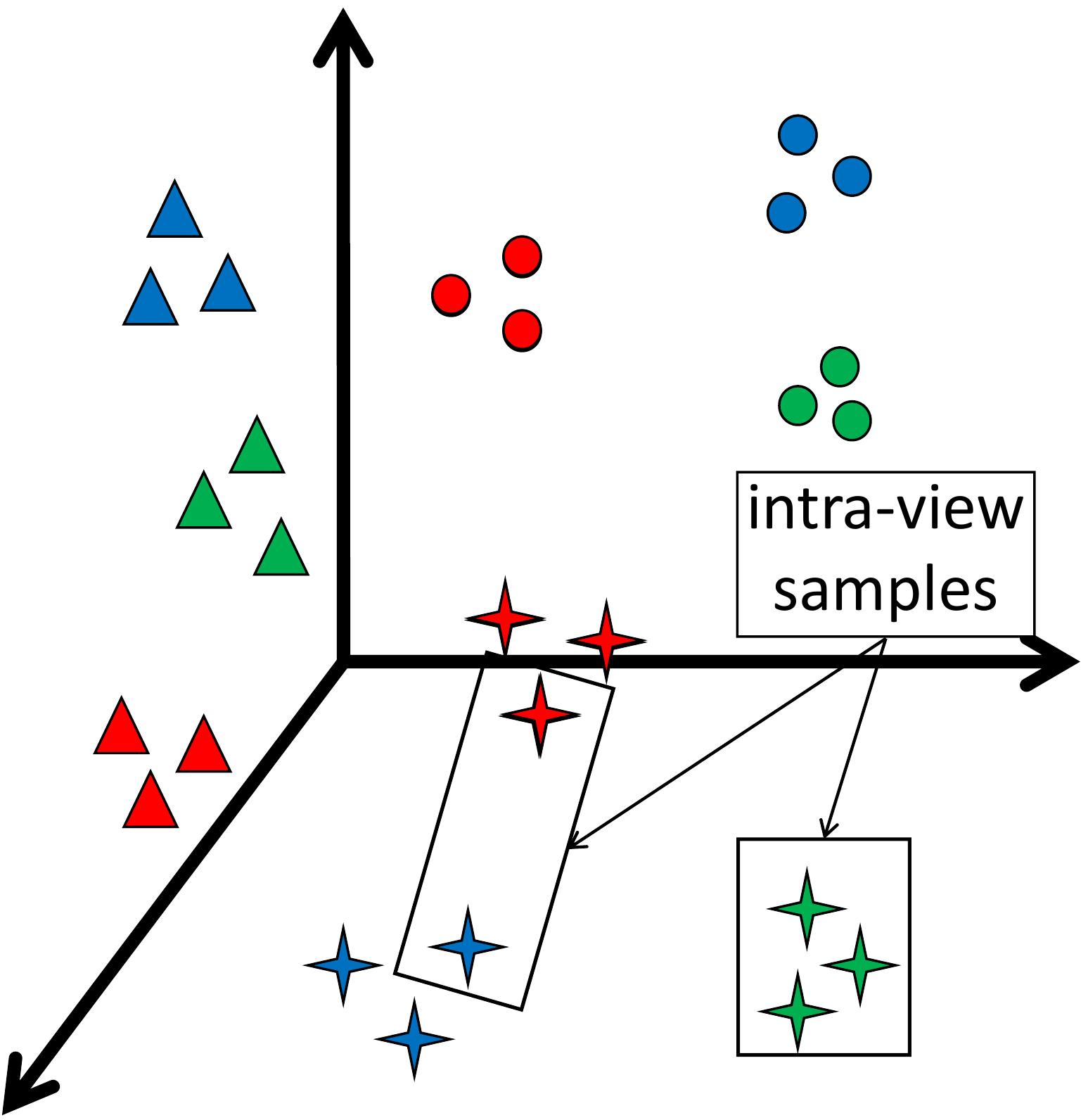}
		\subcaption{\footnotesize{\textit{LDE-intra} embedding}}
	\end{minipage}\hspace{0.3cm}
	\begin{minipage}{3cm}
		\includegraphics[width=3cm]{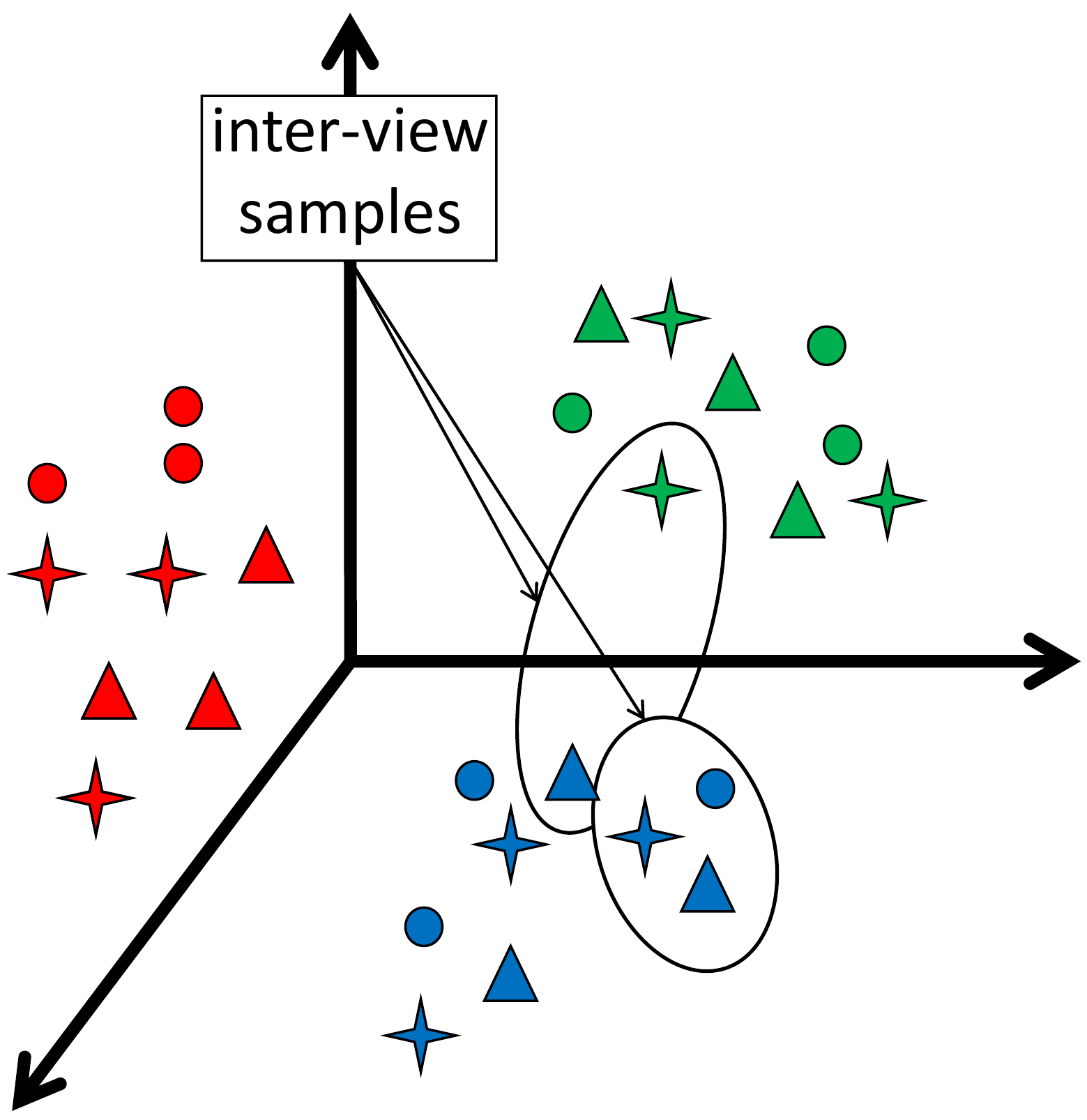}
		\subcaption{\footnotesize{\textit{LDE-inter} embedding}}
	\end{minipage}\hspace{0.3cm}
	\begin{minipage}{3cm}
		\includegraphics[width=3cm]{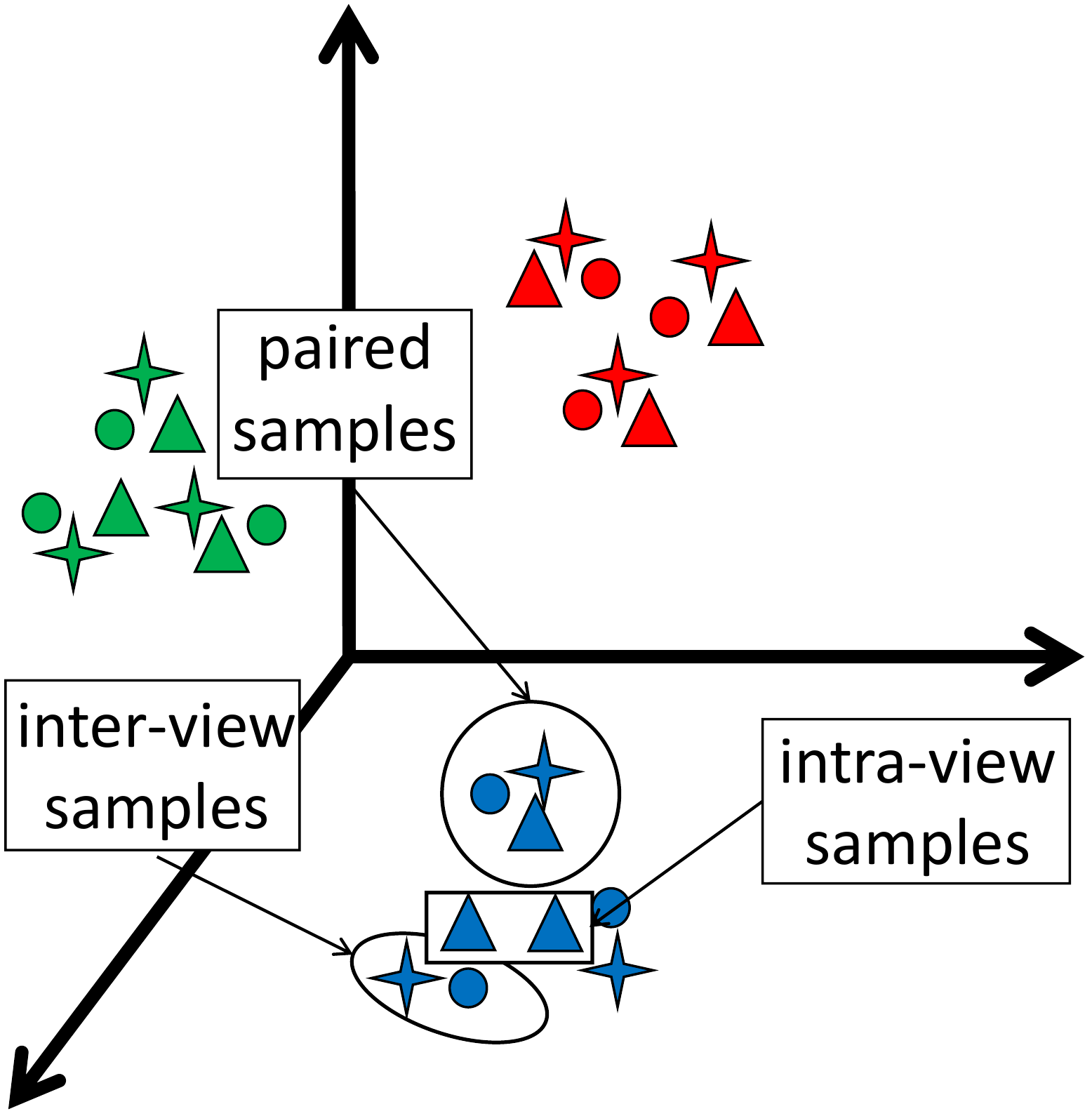}
		\subcaption{\footnotesize{MvHE embedding}}
	\end{minipage}
    \caption{An overview of MvHE. (a) shows the given data (i.e., $\mathcal{X}$) in the original high-dimensional observation space from three distinct views. Different markers indicate different views and different colors denote different classes. The paired samples refer to any two samples from different views but the same object. The inter-view samples denote any two samples from different views and different objects~(our definition), whereas the intra-view samples indicate any two samples from the same view. Without loss of generality, we project $\mathcal{X}$ into a three-dimensional subspace $\mathcal{Y}$. The \textit{LE-paired} embedding shown in (b) enforces convergence of paired samples to a small region in $\mathcal{Y}$. The \textit{LDE-intra} embedding shown in (c) unifies intra-view within-class samples and separates intra-view between-class samples. By contrast, the \textit{LDE-inter} embedding shown in (d) unifies inter-view within-class samples and separates inter-view between-class samples. (e) shows an ideal embedding of MvHE, where within-class samples are close to each other, between-class samples deviate from each other, and paired samples converge to a small region.\vspace{-0.5cm}}
    \label{fig:mvhe_illustration}
\end{figure*}

\textbf{Part optimization.} Assume that $\bm{X} = \left[\bm{x}_1, \bm{x}_2,\ldots, \bm{x}_m\right]\in\mathbb{R}^{D\times{m}}$ denotes a single-view dataset embedded in $D$-dimensional space, where $m$ is the number of samples. Considering $\bm{x}_i$ and its $K$ nearest neighbors, a patch $\bm{X}_i\!=\!\left[\bm{x}_i,{\bm{x}_i}_1,\ldots,{\bm{x}_i}_K\right]\!\in\!\mathbb{R}^{D\times{\left(K+1\right)}}$ with respect to $\bm{x}_i$ is obtained. Given a mapping (or dimensionality reduction) function $f$, for each patch $\bm{X}_i$, we project it to a latent representation $\bm{Y}_i$, where $\bm{Y}_i\!=\!\left[\bm{y}_i,{\bm{y}_i}_1,\ldots,{\bm{y}_i}_K\right]\!\in\!\mathbb{R}^{d\times{\left(K+1\right)}}$. Then, the part optimization is given by:


\begin{align}\label{part_optim}
\mathop{\arg\min}_{\bm{Y}_i}{\mathrm{tr}\left(\bm{Y}_i\bm{L}_i\bm{Y}_i^\mathrm{T}\right)},
\end{align}
where $\bm{L}_i$ is the objective function of the $i$-th patch, and it varies with different algorithms. For example, the $\bm{L}_i$ for LPP~\cite{he2004locality} can be expressed as:
\begin{align}
    &\bm{L}_i= \begin{pmatrix}
        \sum_{j=1}^{K+1}\bm{\theta}_i\!\left(j\right)&-\bm{\theta}_i^\mathrm{T}\\
        -\bm{\theta}_i&\textrm{diag}(\bm{\theta}_i)\\
\end{pmatrix},\\
&\bm{\theta}_i\!\left(j\right)\!=\!\textrm{exp}\left(-\norm{\bm{x}_i-{\bm{x}_i}_j}_2^2/t\right), 
\end{align}
where $\bm{\theta}_i\!\left(j\right)$ refers to the $j$-th element of $\bm{\theta}_i$ and $t$ is a tuning parameter of heat kernel.

Similarly, for DLA, $\bm{L}_i$ becomes:
\begin{align}
    &\bm{L}_i=\begin{pmatrix}
        \sum_{j=1}^{p_1+p_2}\bm{\theta}_i\!\left(j\right)&-\bm{\theta}_i^\mathrm{T}\\
        -\bm{\theta}_i&\textrm{diag}(\bm{\theta}_i)\\
\end{pmatrix},\\
&\bm{\theta}_i=\left[\overbrace{1,\ldots,1,}^{p_1}\overbrace{-\beta,\ldots,-\beta}^{p_2}\right]^\mathrm{T},
\end{align}
where $\beta$ is a scaling factor, $p_1$ is the number of within-class nearest neighbors of $\bm{x}_i$, $p_2$ is the number of between-class nearest neighbors of $\bm{x}_i$ and $p_1+p_2=K$.

\textbf{Whole alignment.} 
The local coordinate $\bm{Y}_i$ can be seen to come from a global coordinate $\bm{Y}$ using a selection matrix $\bm{S}_i$:
\begin{align}\label{whole_align}
\bm{Y}_i = \bm{Y}\bm{S}_i, \quad
\left(\bm{S}_i\right)_{pq}=\left\{
\begin{aligned}
1, & \quad \textrm{if}\quad p=F_i\left(q\right); \\
0, & \quad \textrm{others},
\end{aligned}
\right.
\end{align}
where $\bm{S}_i\!\in\!\mathop{R}^{m\times\left(K+1\right)}$ is the mentioned selection matrix, $\left(\bm{S}_i\right)_{pq}$ denotes its element in the $p$-th row and $q$-th column, and $F_i\!=\!\{i, i_1,...,i_K\}$ stands for the set of indices for the $i$-th patch. Then, Eq.~(\ref{part_optim}) can be rewritten as:
\begin{align}\label{part_optim2}
\mathop{\arg\min}_{\bm{Y}}{\mathrm{tr}\left(\bm{Y}\bm{S}_i\bm{L}_i\bm{S}_i^\mathrm{T}\bm{Y}^\mathrm{T}\right)}.
\end{align}

By summing over all patches, the whole alignment of $\bm{X}$ (also the overall manifold learning objective) can be expressed as:
\begin{align}\label{patch_align}
&\mathop{\arg\min}_{\bm{Y}}\sum_{i=1}^m{\mathrm{tr}\left(\bm{Y}\bm{S}_i\bm{L}_i\bm{S}_i^\mathrm{T}\bm{Y}^\mathrm{T}\right)}\nonumber\\
&=\mathop{\arg\min}_{\bm{Y}}{\mathrm{tr}\left(\bm{Y}\bm{L}\bm{Y}^\mathrm{T}\right).}
\end{align}


\section{Multi-view Hybrid Embedding~(MvHE) and Its Kernel Extension}
\label{sec_proposed}
In this section, we first detail the motivation and general idea of our proposed MvHE, and then give its optimization procedure. The kernel extension to MvHE and the computational complexity analysis are also conducted.


\subsection{Multi-view Hybrid Embedding}

The view discrepancy in multi-view data disrupts the local geometry preservation, posing a new challenge of handling view discrepancy, discriminability and nonlinearity simultaneously~\cite{kan2016multideep,cao2017generalized,ou2016multi}. Inspired by the \textit{Divide-and-Conquer} strategy, we partition the general problem of cross-view classification into three subproblems: 

\noindent Subproblem \uppercase\expandafter{\romannumeral1}: Remove view discrepancy;

\noindent Subproblem \uppercase\expandafter{\romannumeral2}: Increase discriminability and discover the intrinsic nonlinear embedding in intra-view samples;

\noindent Subproblem \uppercase\expandafter{\romannumeral3}: Increase discriminability and discover the intrinsic nonlinear embedding in inter-view samples.

Three models (i.e., \textit{LE-paired}, \textit{LDE-intra} and \textit{LDE-inter}), each for one subproblem, are developed and integrated in a joint manner. We term the combined model MvHE. An overview of MvHE is shown in Fig.~\ref{fig:mvhe_illustration}.


\subsubsection{\textit{LE-paired}}
Since paired samples are collected from the same object, it makes sense to assume that they share common characteristics~\cite{sharma2012generalized,cai2013regularized}, and extracting view-invariant feature representations could effectively remove view discrepancy~\cite{kan2016multideep}. Motivated by this idea, we require the paired samples converge to a small region~(or even a point) in the latent subspace $\mathcal{Y}$. Therefore, the objective of \textit{LE-paired} can be represented as:
\begin{align}\label{paired_1}
\mathcal{J}_1\!=\!\mathop{\arg\min}_{\bm{W}_1, \bm{W}_2,\ldots,\bm{W}_n}\sum_{i=1}^n\sum_{\substack{j=1\\j\neq{i}}}^n\sum_{k=1}^m
\norm{\bm{W}_i^\mathrm{T}\bm{x}_i^k-\bm{W}_j^\mathrm{T}\bm{x}_j^k}_2^2.
\end{align}

\subsubsection{\textit{LDE-intra}}
To improve intra-view discriminability, \textit{LDE-intra} attempts to unify intra-view within-class samples and separate intra-view between-class samples. A naive objective to implement this idea is given by:
\begin{align}\label{intra_1}
&\mathcal{J}_2\!=\!\mathop{\arg\min}_{\bm{W}_1, \bm{W}_2,\ldots,\bm{W}_n}\sum_{i=1}^n\sum_{k_1=1}^m\sum_{\substack{k_2=1\\k_2\neq{k_1}}}^m
\alpha\norm{\bm{W}_i^\mathrm{T}\bm{x}_i^{k_1}-\bm{W}_i^\mathrm{T}\bm{x}_i^{k_2}}_2^2, \nonumber\\
&\alpha=\left\{\begin{array}{ll}
1, &\quad \text{$\bm{x}_i^{k_1}$ and $\bm{x}_i^{k_2}$ are from the same class;}\\
-\beta,&\quad \textrm{otherwise;}
\end{array}
\right.
\end{align}
where $\beta\!>\!0$ is a scaling factor to unify different measurements of the within-class samples and the between-class samples.

One should note that Eq.~(\ref{intra_1}) is incapable of discovering the nonlinear structure embedded in high-dimensional observation space due to its global essence. Thus, to extend Eq.~(\ref{intra_1}) to its nonlinear formulation, inspired by the part optimization and whole alignment framework illustrated in Sect. II-B, for any given sample $\bm{x}_i^{k}$, we build an intra-view local patch containing $\bm{x}_i^k$, its $p_1$ within-class nearest samples, and $p_2$ between-class nearest samples~(see Fig.~\ref{fig:local_patch} for more details), thus formulating the part discriminator as:
\begin{align}\label{intra_part}
    \sum_{s\!=\!1}^{p_1}&\norm{\bm{W}_i^\mathrm{T}\bm{x}_i^k\!-\!\bm{W}_i^\mathrm{T}\left(\bm{x}_i^k\right)_s}_2^2\!-\!\beta\!\sum_{t\!=\!1}^{p_2}\norm{\bm{W}_i^\mathrm{T}\bm{x}_i^k\!-\!\bm{W}_i^T\left(\bm{x}_i^k\right)^t}_2^2,
\end{align}
where $\left(\bm{x}_i^k\right)_s$ is the $s$-th within-class nearest sample of $\bm{x}_i^k$ and $\left(\bm{x}_i^k\right)^t$ is the $t$-th between-class nearest sample of $\bm{x}_i^k$. It is worth mentioning that each sample is associated with such a local patch. By summing over all the part optimizations described in Eq.~(\ref{intra_part}), we obtain the whole alignment (also the overall objective) of \textit{LDE-intra} as:
\begin{align}\label{intra_2}
\begin{split}
\mathcal{J}_2^{'}\!=\!\mathop{\arg\min}_{\bm{W}_1, \bm{W}_2,\ldots,\bm{W}_n}\sum_{i\!=\!1}^n\!&\sum_{k\!=1\!}^m
\!\bigg(\!\sum_{s\!=\!1}^{p_1}\norm{\bm{W}_i^\mathrm{T}\bm{x}_i^k\!-\!\bm{W}_i^\mathrm{T}\left(\bm{x}_i^k\right)_s}_2^2\\
&\!-\!\beta\!\sum_{t\!=\!1}^{p_2}\norm{\bm{W}_i^\mathrm{T}\bm{x}_i^k\!-\!\bm{W}_i^T\left(\bm{x}_i^k\right)^t}_2^2\bigg).
\end{split}
\end{align}

The intra-view discriminant information and local geometry of observed data can be learned by minimizing Eq.~(\ref{intra_2}).

\subsubsection{\textit{LDE-inter}}
Similar to our operations on intra-view samples in Eq.~(18), a reliable way to improve inter-view discriminability is to unify inter-view within-class samples and separate inter-view between-class samples. Thus, a naive objective can be represented as:
\begin{align}\label{inter_1}
&\mathcal{J}_3\!=\!\mathop{\arg\min}_{\bm{W}_1, \bm{W}_2,\ldots,\bm{W}_n}\sum_{i=1}^n\sum_{k_1=1}^m\sum_{\substack{j=1\\j\neq{i}}}^n\sum_{\substack{k_2=1\\k_2\neq{k_1}}}^m
\alpha\norm{\bm{W}_i^\mathrm{T}\bm{x}_i^{k_1}-\bm{W}_j^\mathrm{T}\bm{x}_j^{k_2}}_2^2.\nonumber\\
&\alpha=\left\{\begin{array}{ll}
1, &\quad \text{$\bm{x}_i^{k_1}$ and $\bm{x}_j^{k_2}$ are from the same class;}\\
-\beta,&\quad \textrm{otherwise.}
\end{array}
\right.
\end{align}

\begin{figure}[t]
    \centering
    \includegraphics[height=4.5cm,width=8cm]{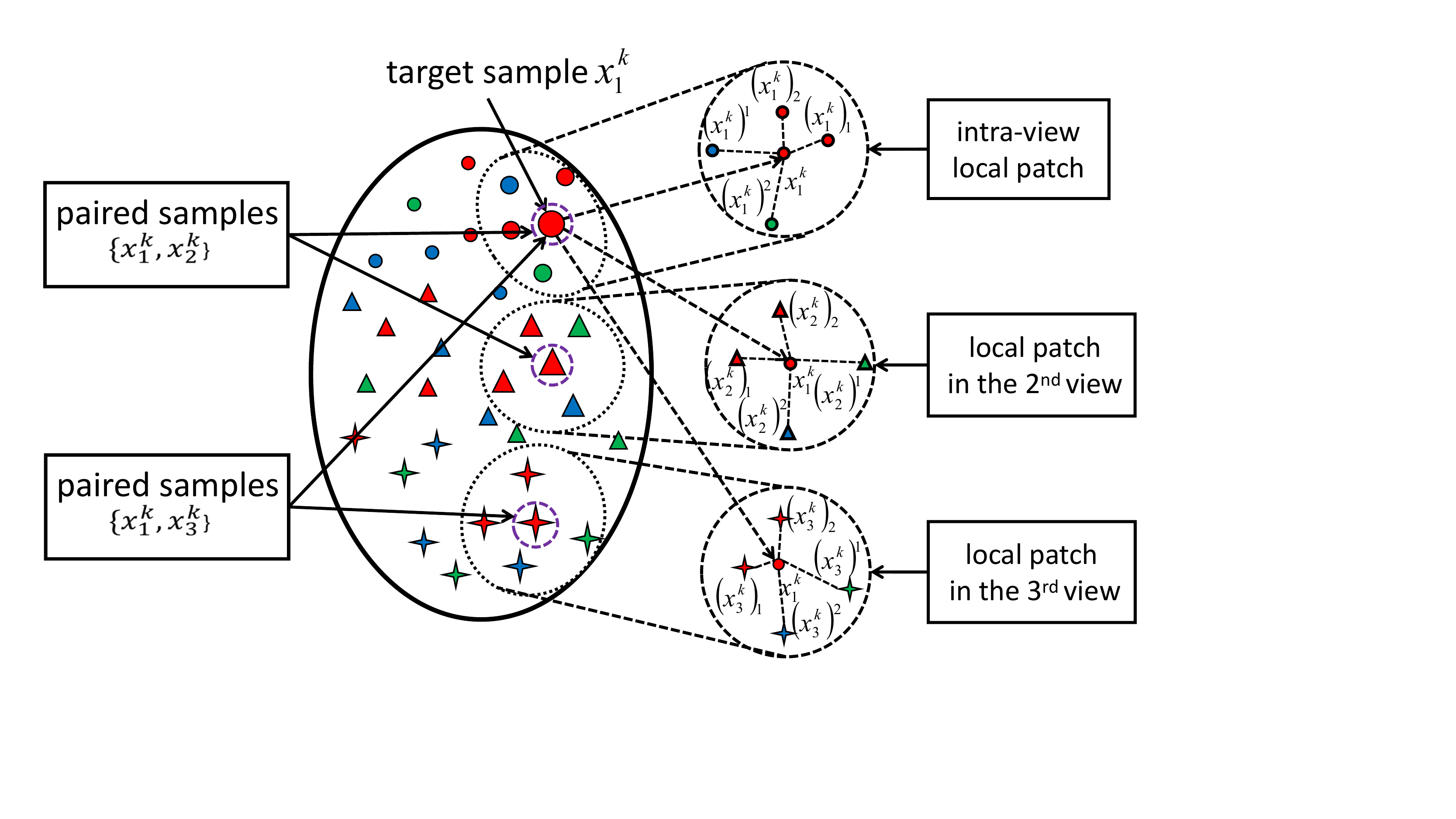}
    \caption{The local patches in \textit{LDE-intra} and \textit{LDE-inter} with respect to a target sample $\bm{x}_1^k$ (the red round with a purple dashed circle around it). Suppose we are given a dataset $\mathcal{X}$ from three views. Different markers indicate different views and different colors denote different classes. 
    \textit{LDE-intra} implements part optimization to $\bm{x}_1^k$ in its intra-view local patch (see the top dashed circle on the right), whereas \textit{LDE-inter} carries out part optimization to $\bm{x}_1^k$ in its different inter-view local patches. Specifically, to construct the local patch of $\bm{x}_1^k$ in the $2^{\text{nd}}$ view (see the middle dashed circle on the right), we select $p_1$ within-class nearest samples with respect to $\bm{x}_2^k$ (the paired sample of $\bm{x}_1^k$ in the $2^{\text{nd}}$ view, i.e., the red triangle with purple dashed circle around it) and $p_2$ between-class nearest samples also with respect to $\bm{x}_2^k$. Similarly, the local patch of $\bm{x}_1^k$ in the $3^{\text{rd}}$ view (see the bottom dashed circle on the right) is constructed using $p_1$ within-class nearest samples and $p_2$ between-class nearest samples with respect to $\bm{x}_3^k$ (the red four-pointed star with purple dashed circle around it).\vspace{-0.5cm}}
   \label{fig:local_patch}
\end{figure}

Similarly, Eq.~(\ref{inter_1}) fails to preserve the local geometry. Hence, we design \textit{LDE-inter} by generalizing the part optimization and whole alignment framework to multi-view scenario. In part optimization phase, each sample $\bm{x}_i^k$ is associated with $n\!-\!1$ inter-view local patches, one for each view, and each local patch includes $p_1$ nearest samples of a same class and $p_2$ nearest samples of different classes. Nevertheless, it is infeasible to directly calculate the similarity between $\bm{x}_i^k$ and heterogeneous samples due to the large view discrepancy. As an alternative, we make use of the paired samples of $\bm{x}_i^k$ in different views, and construct the local patch of $\bm{x}_i^k$ in the $j$-th ($j\neq i$) view using $\bm{x}_i^k$, $p_1$ within-class nearest samples with respect to $\bm{x}_j^k$, and $p_2$ between-class nearest samples also with respect to $\bm{x}_j^k$~(see Fig.~\ref{fig:local_patch} for more details). We thus formulate the part discriminator as:
\begin{align}\label{inter_part}
\begin{split}
    \sum_{\substack{j\!=\!1\\j\neq{i}}}^n\!\bigg(\!\sum_{s=1}^{p_1}&\norm{\bm{W}_i^\mathrm{T}\bm{x}_i^k\!-\!\bm{W}_j^\mathrm{T}\!\left(\bm{x}_j^k\!\right)_s}_2^2\\
    &\quad\quad\quad\!-\!\beta\!\sum_{t=1}^{p_2}\norm{\bm{W}_i^\mathrm{T}\bm{x}_i^k\!-\!\bm{W}_j^T\!\left(\bm{x}_j^k\!\right)^t}_2^2\!\bigg).
\end{split}
\end{align}

By summing over all the part optimization terms described in Eq.~(\ref{inter_part}), we obtain the whole alignment (also the overall objective) of \textit{LDE-inter} as:

\begin{align}\label{inter_2}
\begin{split}
\mathcal{J}_3^{'}\!=\!\mathop{\arg\min}_{\bm{W}_1, \bm{W}_2,\ldots,\bm{W}_n}\sum_{i=1}^n\sum_{k=1}^m&\sum_{\substack{j\!=\!1\\j\neq{i}}}^n
\!\bigg(\!\sum_{s=1}^{p_1}\norm{\bm{W}_i^\mathrm{T}\bm{x}_i^k\!-\!\bm{W}_j^\mathrm{T}\!\left(\bm{x}_j^k\!\right)_s}_2^2\\
&\!-\!\beta\!\sum_{t=1}^{p_2}\norm{\bm{W}_i^\mathrm{T}\bm{x}_i^k\!-\!\bm{W}_j^T\!\left(\bm{x}_j^k\!\right)^t}_2^2\!\bigg).
\end{split}
\end{align}

The local discriminant embedding of inter-view samples can be obtained by minimizing Eq.~(\ref{inter_2}).

\subsubsection{MvHE}
Combining Eqs.~(\ref{paired_1}), (\ref{intra_2}) and (\ref{inter_2}), we obtain the overall objective of MvHE:
\begin{align}\label{all_relation_2}
&\min_{\bm{W}_1,\bm{W}_2,\ldots,\bm{W}_n}\mathcal{J}_1\!+\!\lambda_1\mathcal{J}_2^{'}\!+\!\lambda_2\mathcal{J}_3^{'}\nonumber\\
&\text{s.t.} \quad \bm{W}_i^\mathrm{T}\bm{W}_i=\textbf{I},\quad i=1,2,\ldots,n,
\end{align}
where $\lambda_1$ and $\lambda_2$ are trade-off parameters and an orthogonal constraint is imposed to enforce an orthogonal subspace.

\subsection{Solution to MvHE}
\label{sec_optim}
We simplify~(\ref{all_relation_2}) for further analysis. Let $\bm{y}_i^k=\bm{W}_i^\mathrm{T}\bm{x}_i^k$, (\ref{all_relation_2}) becomes:
\begin{align}\label{all_relation_3}
&\min\!\sum_{i=1}^n\!\sum_{\substack{j\!=\!1\\j\neq{i}}}^n\!\sum_{k=1}^m\norm{\bm{y}_i^k\!-\!\bm{y}_j^k}_2^2\nonumber\\
&\!+\!\lambda_1\!\sum_{i=1}^n\!\sum_{k=1}^m\left(\sum_{s=1}^{p_1}\norm{\bm{y}_i^k\!-\!\left(\bm{y}_i^k\right)_s}_2^2\!-\!\beta\sum_{t=1}^{p_2}\norm{\bm{y}_i^k\!-\!\left(\bm{y}_i^k\right)^t}_2^2\right)\nonumber\\
&\!+\!\lambda_{2}\!\sum_{i=1}^n\!\sum_{k=1}^m\sum_{\substack{j\!=\!1\\j\neq{i}}}^n\left(\!\sum_{s=1}^{p_1}\norm{\bm{y}_i^k\!-\!\left(\bm{y}_j^k\right)_s}_2^2\!-\!\beta\!\sum_{t=1}^{p_2}\norm{\bm{y}_i^k\!-\!\left(\bm{y}_j^k\right)^t}_2^2\right),
\end{align}
where $\left(\bm{y}_i^k\right)_s$ denotes the $s$-th within-class nearest neighbor of $\bm{y}_i^k$ and $\left(\bm{y}_i^k\right)^t$ stands for the $t$-th between-class nearest neighbor of $\bm{y}_i^k$.


Denote
\begin{align}
&\bm{W} = \begin{pmatrix}
\bm{W}_1\\
\bm{W}_2\\
\vdots\\
\bm{W}_n\\
\end{pmatrix},\nonumber\\
&\bm{X}=\begin{pmatrix}
\bm{X}_1&&&\\
&\bm{X}_2&&\\
&&\ddots&\\
&&&\bm{X}_n\\
\end{pmatrix},\nonumber\\
&\bm{Y}=\begin{pmatrix}\bm{Y}_1&\bm{Y}_2&\ldots&\bm{Y}_n\end{pmatrix},\nonumber
\end{align}
where $\bm{W}_v$ ($1\leq v\leq n$) is the mapping function of the $v$-th view, $\bm{X}_v$ denotes data matrix from the $v$-th view, $\bm{Y}_v = \bm{W}_v^\mathrm{T}\bm{X}_v$ is the projection of $\bm{X}_v$ in latent subspace. Obviously, $\bm{Y}=\bm{W}^\mathrm{T}\bm{X}$.

This way, the matrix form of the objective of \textit{LE-paired} can be expressed as~(see supplementary material for more details):

\begin{align}\label{paired_sol}
\mathcal{J}_1&=\sum_{i=1}^n\sum_{\substack{j=1\\j\neq{i}}}^n\sum_{k=1}^m\norm{\bm{y}_i^k-\bm{y}_j^k}_2^2\nonumber\\
&=\mathrm{tr}\left(\bm{W}^\mathrm{T}\bm{X}\bm{J}\bm{X}^\mathrm{T}\bm{W}\right),
\end{align}
where $\bm{J}\!\in\!\mathbb{R}^{mn\times{mn}}$ is defined as below:
\begin{align}\label{defined_J}
&\bm{J}=\sum_{i=2}^n\left(\textbf{I}-\bm{J}_i\right)\left(\textbf{I}-\bm{J}_i\right)^\mathrm{T},\nonumber\\
&\bm{J}_i=\begin{bmatrix}
&\textbf{I}_{\left(\left(i-1\right)*{m}\right)}\\
\textbf{I}_{\left(\left(n-i+1\right)*{m}\right)}&\\
\end{bmatrix}.
\end{align}

On the other hand, the matrix form of objective of \textit{LDE-intra} is given by~(see supplementary material for more details):
\begin{align}\label{intra_sol}
\mathcal{J}_2^{'}&=\sum_{i=1}^n\sum_{k=1}^m\left(\sum_{s=1}^{p_1}\norm{\bm{y}_i^k-\left(\bm{y}_i^k\right)_s}_2^2-\beta\sum_{t=1}^{p_2}\norm{\bm{y}_i^k-\left(\bm{y}_i^k\right)^t}_2^2\right)\nonumber\\
&=\mathrm{tr}\left(\bm{W}^\mathrm{T}\bm{X}\bm{U}\bm{X}^\mathrm{T}\bm{W}\right),
\end{align}
where $\bm{U}\!\in\!\mathbb{R}^{mn\times{mn}}$ is defined as below: 
\begin{align}\label{defined_U}
&\bm{U}=\sum_{i=1}^n\sum_{k=1}^m\left(\bm{S}_i^k\bm{L}_i^k\left(\bm{S}_i^k\right)^\mathrm{T}\right),\nonumber\\
&\bm{L}_i^k = \begin{bmatrix}-\bm{e}_{p_1+p_2}^\mathrm{T}\\
\textbf{I}_{p_1+p_2}
\end{bmatrix}\textrm{diag}\left(\bm{\theta}_i^k\right)\begin{bmatrix}
-\bm{e}_{p_1+p_2}&\textbf{I}_{p_1+p_2}
\end{bmatrix},\nonumber\\
&\bm{\theta}_i^k=\left[\overbrace{1,\ldots,1,}^{p_1}\overbrace{-\beta,\ldots,-\beta}^{p_2}\right]^\mathrm{T},\nonumber\\
&\left(\bm{S}_i^k\right)_{pq}=\left\{
\begin{aligned}
1, & \quad \textrm{if}\quad p=F_i^k\left(q\right); \\
0, & \quad \textrm{others},
\end{aligned}
\right.\nonumber\\
&F_i^k=\{*_i^k, \left(*_i^k\right)_1,\dots,\left(*_i^k\right)_{p_1},\left(*_i^k\right)^1,\dots,\left(*_i^k\right)^{p_2}\},\nonumber\\
&\bm{Y}_i^k = \bm{Y}S_i^k,\nonumber\\
&\bm{Y}_i^k\!=\!\left[\bm{y}_i^k,\left(\bm{y}_i^k\right)_1,\dots,\left(\bm{y}_i^k\right)_{p_1},\left(\bm{y}_i^k\right)^1,\dots,\left(\bm{y}_i^k\right)^{p_2}\right].
\end{align}

In particular, $\bm{L}_i^k\!\in\!\mathbb{R}^{\left(p_1+p_2+1\right)\times{\left(p_1+p_2+1\right)}}$ encodes the objective function for the local patch with respect to $\bm{x}_i^k$.  $\bm{S}_i^k\!\in\!\mathbb{R}^{mn\times{\left(p_1+p_2+1\right)}}$ is a selection matrix, $F_i^k$ denotes the set of indices for the local patch of $\bm{x}_i^k$. $\bm{Y}_i^k$ is the representation matrix with respect to $\bm{x}_i^k$ and its local patch in the common subspace.

Similarly, the matrix form of the objective of \textit{LDE-inter} can be written as:
\begin{align}\label{inter_sol}
\mathcal{J}_3^{'}&=\sum_{i=1}^n\sum_{k=1}^m\sum_{\substack{j=1\\j\neq{i}}}^n\left(\sum_{s=1}^{p_1}\norm{\bm{y}_i^k-\left(\bm{y}_j^k\right)_s}_2^2-\beta\sum_{t=1}^{p_2}\norm{\bm{y}_i^k-\left(\bm{y}_j^k\right)^t}_2^2\right)\nonumber\\
&=\mathrm{tr}\left(\bm{W}^\mathrm{T}\bm{X}\bm{V}\bm{X}^\mathrm{T}\bm{W}\right),
\end{align}
where $\bm{V}\!\in\!\mathbb{R}^{mn\times{mn}}$ is defined as below:
\begin{align}\label{defined_V}
&\bm{V}=\sum_{i=1}^n\sum_{k=1}^m\sum_{\substack{j=1\\j\neq{i}}}^n\left(\bm{S}_j^k\bm{L}_j^k\left(\bm{S}_j^k\right)^\mathrm{T}\right),\nonumber\\
&\bm{L}_j^k = \begin{bmatrix}-\bm{e}_{p_1+p_2}^\mathrm{T}\\
\textbf{I}_{p_1+p_2}
\end{bmatrix}\textrm{diag}\left(\bm{\theta}_j^k\right)\begin{bmatrix}
-\bm{e}_{p_1+p_2}&\textbf{I}_{p_1+p_2}
\end{bmatrix},\nonumber\\
&\bm{\theta}_j^k=\left[\overbrace{1,\ldots,1,}^{p_1}\overbrace{-\beta,\ldots,-\beta}^{p_2}\right]^\mathrm{T},\nonumber\\
&\left(\bm{S}_j^k\right)_{pq}=\left\{
\begin{aligned}
1, & \quad \textrm{if}\quad p=F_j^k\left(q\right); \\
0, & \quad \textrm{others},
\end{aligned}
\right.\nonumber\\
&F_j^k=\{*_i^k, \left(*_j^k\right)_1,\dots,\left(*_j^k\right)_{p_1},\left(*_j^k\right)^1,\dots,\left(*_j^k\right)^{p_2}\},\nonumber\\
&\bm{Y}_j^k = \bm{Y}S_j^k,\nonumber\\
&\bm{Y}_j^k\!=\!\left[\bm{y}_i^k,\left(\bm{y}_j^k\right)_1,\dots,\left(\bm{y}_j^k\right)_{p_1},\left(\bm{y}_j^k\right)^1,\dots,\left(\bm{y}_j^k\right)^{p_2}\right].
\end{align}

Combining Eqs.~(\ref{paired_sol}), (\ref{intra_sol}) and (\ref{inter_sol}), the objective of MvHE can be reformulated as:
\begin{align}\label{mllda_sol}
&\min_{\bm{W}}\mathrm{tr}\left(\bm{W}^\mathrm{T}\bm{X}\left(\bm{J}+\lambda_{1}\bm{U}+\lambda_{2}\bm{V}\right)\bm{X}^\mathrm{T}\bm{W}\right)\nonumber\\
&\text{s.t.} \quad \bm{W}^\mathrm{T}\bm{W}=n\textbf{I}.
\end{align}


Eq.~(\ref{mllda_sol}) is a standard eigenvalue decomposition problem. The optimal solution of $\bm{W}$ consists of the eigenvectors corresponding to the $d$ smallest eigenvalues of $\bm{X}\left(\bm{J}+\lambda_{1}\bm{U}+\lambda_{2}\bm{V}\right)\bm{X}^\mathrm{T}$.

\subsection{Kernel Multi-view Hybrid Embedding~(KMvHE)}
KMvHE performs MvHE in the Reproducing Kernel Hilbert Space (RKHS). Given $\bm{X}_v=\left[\bm{x}_v^1,\bm{x}_v^2,\ldots,\bm{x}_v^m\right]$, there exists a nonlinear mapping $\phi$ such that $\kappa\left(\bm{x}_v^i,\bm{x}_v^j\right)=\langle\!\phi\!\left(\bm{x}_v^i\right),\!\phi\!\left(\bm{x}_v^j\right)\rangle$, where $\kappa\left(\bm{x},\bm{z}\right)$ is the kernel function. For $\bm{X}_v$, its kernel matrix $\bm{K}_v = \langle\!\phi\!\left(\bm{X}_v\right),\!\phi\!\left(\bm{X}_v\right)\rangle$ can be obtained with $\bm{K}_v\left(i,j\right)=\kappa\left(\bm{x}_v^i,\bm{x}_v^j\right)$. We denote the mapping function $\bm{W}_v$ in RKHS as $\phi\!\left(\bm{W}_v\right)$. Assume that atoms of $\bm{W}_v$ lie in the space spanned by $\bm{X}_v$~\cite{scholkopf2001generalized}, we have:
\begin{align}\label{repre_theory}
\phi\!\left(\bm{W}_v\right)\!=\!\phi\!\left(\bm{X}_v\right)\bm{A}_v,
\end{align}
where $\bm{A}_v$ is an atom matrix of the $v$-th view and $\phi\!\left(\bm{X}_v\right)=\left[\phi\!\left(\bm{x}_v^1\right),\phi\!\left(\bm{x}_v^2\right),\ldots,\phi\!\left(\bm{x}_v^m\right)\right]$. Then, $\bm{W}$ can be rewritten as:
\begin{align}\label{repre_W}
\phi\!\left(\bm{W}\right)&=\begin{pmatrix}
\phi\!\left(\bm{W}_1\right)\\
\phi\!\left(\bm{W}_2\right)\\
\vdots\\
\phi\!\left(\bm{W}_v\right)
\end{pmatrix}
=\begin{pmatrix}
\phi\!\left(\bm{X}_1\right)\bm{A}_1\\
\phi\!\left(\bm{X}_2\right)\bm{A}_2\\
\vdots\\
\phi\!\left(\bm{X}_n\right)\bm{A}_n
\end{pmatrix}\nonumber\\
&=\begin{pmatrix}
\phi\!\left(\bm{X}_1\right)&&&\\
&\phi\!\left(\bm{X}_2\right)&&\\
&&\ddots&\\
&&&\phi\!\left(\bm{X}_n\right)\\
\end{pmatrix}
\begin{pmatrix}
\bm{A}_1\\
\bm{A}_2\\
\vdots\\
\bm{A}_n\\
\end{pmatrix}\nonumber\\
&=\phi\!\left(\bm{X}\right)\bm{A}.
\end{align}

Given that
\begin{align}\label{kernelize}
&\mathrm{tr}\left(\phi^\mathrm{T}\left(\bm{W}\right)\phi\!\left(\bm{X}\right)\left(\bm{J}\!+\!\lambda_{1}\bm{U}\!+\!\lambda_{2}\bm{V}\right)\phi^\mathrm{T}\left(\bm{X}\right)\phi\!\left(\bm{W}\right)\right)\nonumber\\
=&\mathrm{tr}\left(\bm{A}^\mathrm{T}\phi^\mathrm{T}\left(\bm{X}\right)\phi\!\left(\bm{X}\right)\left(\bm{J}\!+\!\lambda_{1}\bm{U}\!+\!\lambda_{2}\bm{V}\right)\phi^\mathrm{T}\left(\bm{X}\right)\phi\!\left(\bm{X}\right)\bm{A}\right)\nonumber\\
=&\mathrm{tr}\left(\bm{A}^\mathrm{T}\bm{K}\left(\bm{J}\!+\!\lambda_{1}\bm{U}\!+\!\lambda_{2}\bm{V}\right)\bm{K}\bm{A}\right),
\end{align}
and
\begin{align}
&\phi^\mathrm{T}\left(\bm{W}\right)\phi\!\left(\bm{W}\right)\nonumber\\
=&\bm{A}^\mathrm{T}\phi^\mathrm{T}\left(\bm{X}\right)\phi\!\left(\bm{X}\right)\bm{A}\nonumber\\
=&\bm{A}^\mathrm{T}\bm{K}\bm{A},
\end{align}
where $\bm{K}=\phi^\mathrm{T}\left(\bm{X}\right)\phi\!\left(\bm{X}\right)$ is the kernel matrix of all samples from all views. Then, the objective of KMvHE in analogue to Eq.~(31) can then be expressed as:
\begin{align}\label{ker_solution}
&\min_{\bm{A}}\mathrm{tr}\left(\bm{A}^\mathrm{T}\bm{K}\left(\bm{J}+\lambda_{1}\bm{U}+\lambda_{2}\bm{V}\right)\bm{K}\bm{A}\right)\nonumber\\
&\text{s.t.} \quad \bm{A}^\mathrm{T}\bm{K}\bm{A}=n\textbf{I}.
\end{align}

Obviously, Eq.~(\ref{ker_solution}) can be optimized the same as MvHE.

\subsection{Computational Complexity Analysis}
\label{sec_complexity}
For convenience, suppose that $d_v\!=\!D\!$~$\left(\!v\!=\!1,2,\ldots,n\!\right)$, we first analyze the computational complexity of MvHE. Since~(\ref{mllda_sol}) is a standard eigenvalue decomposition problem, its computational complexity is $\mathcal{O}\!\left(N^3\right)$, where $N\!=\!nD$ is the order of square matrix $\bm{X}\!\left(\!\bm{J}\!+\!\lambda_1\bm{U}\!+\!\lambda_2\bm{V}\!\right)\!\bm{X}^\mathrm{T}$. On the other hand, the computational complexity of $\bm{J}$ in Eq.~(\ref{paired_sol}), $U$ in Eq.~(\ref{intra_sol}), and $V$ in Eq.~(\ref{inter_sol}) can be easily derived. 
For example, the computational complexity of the selection matrix $\bm{S}_i^k$ in Eq. ~(\ref{defined_U}) is $\mathcal{O}\!\left(\!mD+m\log\left(\!m\!\right)\!+\!mnP\!\right)$, and the computational complexity of $\bm{S}_i^k\bm{L}_i^k\left(\bm{S}_i^k\right)^\mathrm{T}$ is $\mathcal{O}\!\left(\!mnP\!\left(\!P\!+\!mn\!\right)\!\right)$, where $P\!=\!p_1\!+\!p_2$. Therefore, the computational complexity of $\bm{U}$ is given by $\mathcal{O}\!\left(\!m^2n^2P\!\left(\!P\!+\!mn\!\right)\!\right)$. Similarly, the computational complexity of $\bm{V}$ and $\bm{J}$ are $\mathcal{O}\!\left(\!m^2n^3P\!\left(\!P\!+\!mn\!\right)\!\right)$ and $\mathcal{O}\!\left(m^3n^4\right)$. To summarize, the overall computational complexity of MvHE is approximately $\mathcal{O}\!\left(\!m^3n^4+\!m^2n^3P\!\left(\!P\!+\!mn\!\right)\!+\!m^2n^3D\!+\!mn^3D^2\!+\!n^3D^3\right)$.

By contrast, the computational complexity of MvDA and MvMDA are approximately $\mathcal{O}\!\left(\!n^2D^2\!\left(\!C^2\!+\!mn\!+\!nD\!\right)\!\right)$ and $\mathcal{O}\!\left(\!mn^2\!\left(\!C^2m+Dm+D^2\!\right)\!+\!n^3D^3\!\right)$ respectively. In practice, we have $D\!<\!m$, $C\!<\!m$, $P\!<\!m$ and $n\!\ll\!m$, thus the computational complexity of MvHE, MvDA and MvMDA can be further simplified with $\mathcal{O}\!\left(\!m^3n^4P\!\right)$, $\mathcal{O}\!\left(\!n^2D^2C^2\!+\!mn^3D^2\!\right)$ and $\mathcal{O}\!\left(\!m^2n^2C^2\!+\!m^2n^2D\!\right)$. Since $m^3n^2P\!>\!m^2C^2\!>\!D^2C^2$, our method does not exhibit any advantage in computational complexity. Nevertheless, the extra computational cost is still acceptable considering the overwhelming performance gain that will be demonstrated in the next section. On the other hand, the experimental results in Sect.~\ref{sec:face_pose} also corroborate our computational complexity analysis.

\subsection{Relation and Difference to DLA}
\label{label_relation}
We specify the relation and difference of MvHE to DLA~\cite{zhang2009patch}. Both MvHE and DLA are designed to deal with nonlinearity by preserving the local geometry of data. However, MvHE differs from DLA in the following aspects. First, as a well-known single-view algorithm, DLA is not designed for multi-view data, hence it does not reduce the view discrepancy for the problem of cross-view classification. By contrast, MvHE is an algorithm specifically for multi-view data and it removes the view discrepancy by performing \textit{LE-paired}. Second, in DLA, only the discriminant information and the nonlinearity in intra-view samples are taken into consideration. By contrast, MvHE incorporates discriminant information and nonlinear embedding structure in both intra-view samples and inter-view samples via performing \textit{LDE-intra} and \textit{LDE-inter} respectively. One should note that, the inter-view discriminancy and nonlinearity are especially important for cross-view classification since the task aims to distinguish inter-view samples.



\section{Experiments}
\label{sec_experiments}
In this section, we evaluate the performance of our methods on four benchmark datasets: the Heterogeneous Face Biometrics (HFB) dataset, the CUHK Face Sketch FERET (CUFSF) database\footnote{http://mmlab.ie.cuhk.edu.hk/archive/cufsf/}, the CMU Pose, Illumination, and Expression (PIE) database (CMU PIE) and the Columbia University Image Library (COIL-100)\footnote{http://www.cs.columbia.edu/CAVE/software/softlib/coil-100.php}. We first specify the experimental setting in Sect.~\ref{sec:experimental_settings}. Following this, we demonstrate that our methods can achieve the state-of-the-art performance on two-view datasets~(i.e., HFB and CUFSF) in Sect.~\ref{sec:face_sensors}. We then verify that the overwhelming superiority of our methods also holds on multi-view data, including CMU PIE in Sect.~\ref{sec:face_pose} and COIL-100 in Sect.~\ref{sec:object_pose}. Finally, we conduct a sensitivity analysis to different parameters in our methods in Sect.~\ref{sec:parameter_sensitivity}.

\subsection{Experimental Setting}
\label{sec:experimental_settings}
We compare our methods with eight baselines and five state-of-the-art methods. The first two baseline methods, PCA~\cite{turk1991face}, LDA~\cite{belhumeur1997eigenfaces}, are classical classification methods for single-view data. The other four baseline methods are CCA~\cite{hotelling1936relations}, KCCA~\cite{akaho2006kernel}, MCCA~\cite{rupnik2010multi} and PLS~\cite{sharma2011bypassing}, i.e., the most popular unsupervised methods for multi-view data that aim to maximize the~(pairwise) correlations or convariance between multiple latent representations. Due to the close relationship between our methods and DLA~\cite{zhang2009patch} as mentioned in Sect.~\ref{label_relation}. We also list DLA~\cite{zhang2009patch} as a final baseline method. Moreover, we also extend DLA to multi-view scenario~(term it MvDLA). Note that, to decrease discrepancy among views, we have to set a large $p_1$~(the number of within-class nearest neighbors). This straightforward extension, although seemingly reliable, leads to some issues, we will discuss them in the experimental part in detail. Moreover, The state-of-the-art methods selected for comparison include GMA~\cite{sharma2012generalized}, MULDA~\cite{sun2016multiview}, MvDA~\cite{Kan2012}, MvDA-VC~\cite{kan2016multi} and MvMDA~\cite{cao2017generalized}. GMA and MULDA are supervised methods that jointly consider view discrepancy and intra-view discriminability, whereas MvDA, MvDA-VC and MvMDA combine intra-view discriminability and inter-view discriminability in a unified framework.

For a fair comparison, we repeat experiments 10 times independently by randomly dividing the given data into training set and test set at a certain ratio, and report the average results. Furthermore, the hyper-parameters of all methods are determined by 5-fold cross validation. To reduce dimensionality, PCA is first applied for all methods, and the PCA dimensions are empirically set to achieve the best classification accuracy via traversing possible dimensions as conducted in~\cite{kan2016multi}. In the test phase, we first project new samples into a subspace with the learned mapping functions and then categorize their latent representations with a 1-NN classifier. For KCCA and KMvHE, the radial basis function~(RBF) kernel is selected and the kernel size is tuned with cross validation. All the experiments mentioned in this paper are conducted on MATLAB 2013a, with CPU i5-6500 and 16.0GB memory size.

\begin{table}[htb]
	\centering
	\caption{The average classification accuracy $(\%)$ on HFB dataset and CUFSF dataset. The best two results are marked with {\color{red}red} and {\color{blue}blue} respectively.}
	\label{tab:HFB_CUFSF}
	\begin{tabular}{c|cc|cc}
	\hline
    Dataset&\multicolumn{2}{c|}{HFB}&\multicolumn{2}{c}{CUFSF}\\
	\hline
	Gallery-Probe&NIR-VIS&VIS-NIR&Photo-Sketch&Sketch-Photo\\
	\hline
	PCA&18.6$\pm$6.7&19.1$\pm$4.7&26.4$\pm$2.2&15.7$\pm$1.2\\
	LDA&8.1$\pm$2.7&13.5$\pm$3.3&0.6$\pm$0.3&0.7$\pm$0.3\\
	DLA&5.2$\pm$2.1&7.6$\pm$3.1&53.3$\pm$2.1&60.7$\pm$2.9\\
	CCA&22.5$\pm$4.7&22.4$\pm$5.0&47.7$\pm$2.5&48.7$\pm$2.1\\
	KCCA&31.6$\pm$6.3&27.5$\pm$4.1&49.1$\pm$2.3&52.3$\pm$1.6\\
	PLS&23.3$\pm$5.3&20.3$\pm$4.4&59.4$\pm$1.5&31.9$\pm$1.7\\
	MvDLA&35.0$\pm$5.3&33.5$\pm$3.9&53.3$\pm$2.1&60.7$\pm$2.9\\
	GMA&25.1$\pm$5.6&23.3$\pm$3.3&N/A&N/A\\
	MULDA&24.5$\pm$4.2&23.8$\pm$2.5&N/A&N/A\\
	MvDA&32.9$\pm$4.4&28.0$\pm$6.0&59.8$\pm$2.1&60.0$\pm$1.5\\
	MvMDA&9.6$\pm$3.1&16.7$\pm$4.8&N/A&N/A\\
	MvDA-VC&36.3$\pm$5.2&37.0$\pm$4.4&66.1$\pm$1.4&63.1$\pm$2.1\\
	MvHE&{\color{blue}42.1}$\pm$4.4&{\color{blue}42.0}$\pm$6.2&{\color{blue}69.4}$\pm$2.2&{\color{blue}63.4}$\pm$2.2\\
	KMvHE&{\color{red}42.7}$\pm$6.0&{\color{red}42.3}$\pm$6.5&{\color{red}72.3}$\pm$1.9&{\color{red}70.0}$\pm$1.8\\
	\hline
	\end{tabular}
	\vspace{-0.3cm}
\end{table}

\subsection{The Efficacy of (K)MvHE on Two-view Data}
\label{sec:face_sensors}
We first compare (K)MvHE with all the other competing methods on HFB and CUFSF to demonstrate the superiority of our methods on two-view data. HFB~\cite{li2009hfb,yi2007face} contains 100 individuals, and each has 4 visual light images and 4 near-infrared images. This dataset is used to evaluate visual (VIS) light image versus near-infrared (NIR) image heterogeneous classification. CUFSF contains 1194 subjects, with 1 photo and 1 sketch per subject. This dataset is used to evaluate photo versus sketch face classification. For HFB, 60 individuals are selected as training data and the remaining individuals are used for testing. For CUFSF, 700 subjects are selected as training data and the remaining subjects are used for testing. The results are listed in Table~\ref{tab:HFB_CUFSF}. Note that the performance of GMA and MvMDA is omitted on CUFSF, because there is only one sample in each class under each view. 

As can be seen, PCA, LDA and DLA fail to provide reasonable classification results as expected. Note that the performance of DLA is the same as MvDLA on the CUFSF dataset, due to only one within-class nearest neighbor for any one sample on this dataset. CCA and PLS perform the worst among MvSL based methods due to the naive utilization of view discrepancy. KCCA, GMA and MULDA improve the performance of CCA and PLS by incorporating the kernel trick or taking into consideration intra-view discriminant information respectively. On the other hand, MvDA and MvDA-VC achieve a large performance gain over GMA by further considering inter-view discriminant information. Although MvMDA is a supervised method, its performance is very poor. One possible reason is that there is a large class-center discrepancy between views on HFB, and the within-class scatter matrix $\bm{S}_W$ (defined in Eq.~(\ref{MvMDA_scatter})) in MvMDA does not consider inter-view variations. Although MvDLA achieves comparable performance to MvDA, there is still a performance gap between MvDLA and our methods. This result confirms our concern that a large $p_1$ in MvDLA makes it ineffective to preserve local geometry of data. Our methods, MvHE and KMvHE, perform the best on these two databases by properly utilizing local geometric discriminant information in both intra-view and inter-view samples.

\begin{figure}[htb]
	\centering
	\includegraphics[height=3cm,width=8cm]{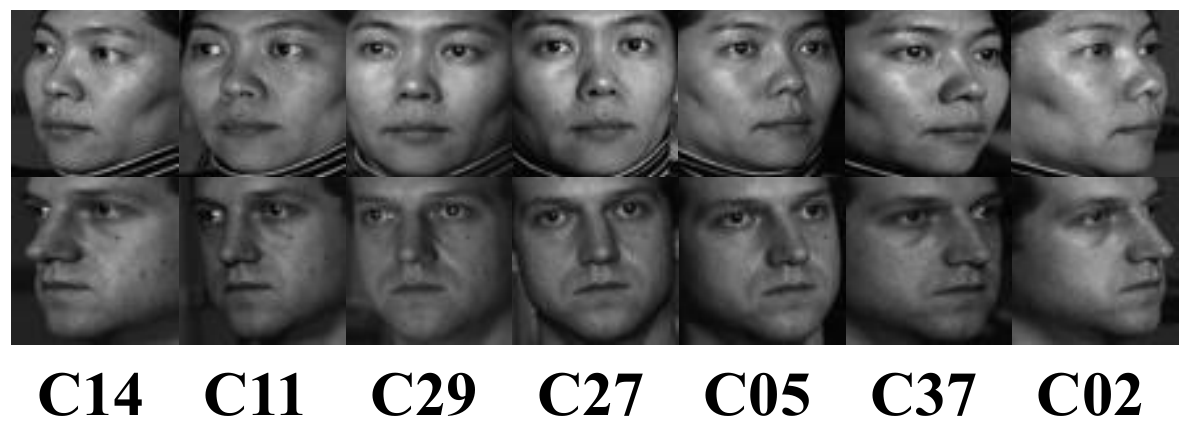}
	\caption{Exemplar subjects from CMU PIE dataset. C14, C11, C29, C27, C05, C37 and C02 are selected to construct multi-view data.\vspace{-0.5cm}}
	\label{fig:CMU_PIE}
\end{figure}

\begin{table*}[htb]
    \centering
	\caption{The average classification accuracy $(\%)$ of our MvHE (the last row) and its degraded baseline variants in terms of mean accuracy (mACC) on CMU PIE dataset. The best two results are marked with {\color{red}red} and {\color{blue}blue} respectively.}
	\label{tab:layers_compared}
	\begin{tabular}{c|c|c||c|c|c|c|c|c|c|c}
	\hline
	\textit{LE-paired}&\textit{LDE-intra}&\textit{LDE-inter}&Case 1&Case 2&Case 3&Case 4&Case 5&Case 6&Case 7&Case 8\\
	\hline
	\checkmark&&&30.1$\pm$6.4&17.0$\pm$4.6&11.7$\pm$3.1&54.2$\pm$6.2&34.8$\pm$3.4&27.3$\pm$5.8&33.7$\pm$6.0&19.0$\pm$2.1\\
	&\checkmark&&15.8$\pm$2.3&11.8$\pm$4.0&10.6$\pm$2.4&23.1$\pm$1.7&14.0$\pm$2.8&10.9$\pm$1.2&21.6$\pm$1.9&20.6$\pm$1.4\\
    &&\checkmark&66.1$\pm$8.9&56.5$\pm$6.9&43.5$\pm$6.5&62.2$\pm$4.7&49.0$\pm$7.0&40.2$\pm$5.9&47.2$\pm$4.4&36.3$\pm$2.4\\
    \checkmark&\checkmark&&\textcolor{blue}{78.4}$\pm$6.7&\textcolor{blue}{69.4}$\pm$5.7&\textcolor{blue}{56.1}$\pm$4.8&79.4$\pm$3.2&68.4$\pm$3.4&\textcolor{blue}{57.7}$\pm$3.8&\textcolor{blue}{78.4}$\pm$2.1&70.9$\pm$2.4\\
    \checkmark&&\checkmark&75.5$\pm$8.9&65.4$\pm$7.0&47.0$\pm$6.6&\textcolor{blue}{80.3}$\pm$4.5&\textcolor{blue}{69.8}$\pm$3.6&56.4$\pm$5.3&78.4$\pm$2.5&\textcolor{blue}{71.0}$\pm$2.8\\
    &\checkmark&\checkmark&68.5$\pm$7.2&58.7$\pm$5.5&45.0$\pm$5.6&63.2$\pm$5.7&49.5$\pm$5.7&40.5$\pm$6.5&48.2$\pm$4.2&37.4$\pm$2.8\\
	\checkmark&\checkmark&\checkmark&{\color{red}78.6}$\pm$7.1&{\color{red}72.1}$\pm$5.4&{\color{red}57.1}$\pm$5.1&{\color{red}80.9}$\pm$4.6&{\color{red}70.2}$\pm$3.6&{\color{red}59.6}$\pm$4.7&{\color{red}78.6}$\pm$3.0&{\color{red}71.5}$\pm$3.0\\
    \hline
	\end{tabular}
	\vspace{-0.2cm}
\end{table*}

\begin{table*}[htb]
    \centering
	\caption{The average classification accuracy $(\%)$ of our methods and their multi-view counterparts in terms of mean accuracy (mACC) on CMU PIE dataset. The best two results are marked with {\color{red}red} and {\color{blue}blue} respectively.}
	\label{tab:cmu_pie_all_cases}
	\begin{tabular}{c|c|c|c|c|c|c|c|c}
	\hline
	Methods&Case 1&Case 2&Case 3&Case 4&Case 5&Case 6&Case 7&Case 8\\
	\hline
	PCA&42.9$\pm$10.6&33.6$\pm$7.1&23.5$\pm$5.8&40.2$\pm$5.5&28.3$\pm$5.0&21.5$\pm$4.5&39.2$\pm$4.6&30.3$\pm$4.2\\
	LDA&19.5$\pm$3.0&12.0$\pm$4.0&10.5$\pm$5.5&21.2$\pm$2.5&11.8$\pm$3.1&9.7$\pm$2.6&20.9$\pm$2.4&18.1$\pm$2.2\\
	DLA&14.0$\pm$6.4&16.7$\pm$4.0&16.9$\pm$3.8&19.6$\pm$3.7&16.7$\pm$2.8&20.2$\pm$1.9&20.1$\pm$2.3&19.6$\pm$3.7\\
	MCCA&54.2$\pm$7.4&44.5$\pm$9.0&28.2$\pm$4.6&57.4$\pm$3.7&39.2$\pm$5.3&32.6$\pm$4.2&48.4$\pm$3.4&39.4$\pm$1.8\\
	MvDLA&66.4$\pm$7.8&58.0$\pm$3.5&45.9$\pm$4.1&68.4$\pm$4.5&56.2$\pm$3.6&44.2$\pm$3.7&56.4$\pm$2.6&42.6$\pm$1.3\\
	GMA&58.3$\pm$8.2&53.0$\pm$9.0&34.0$\pm$6.4&60.3$\pm$6.9&43.2$\pm$4.0&37.0$\pm$3.8&55.0$\pm$3.7&47.4$\pm$4.4\\
	MULDA&58.1$\pm$8.2&52.4$\pm$7.3&36.1$\pm$6.2&63.4$\pm$5.1&47.8$\pm$4.8&39.4$\pm$5.5&58.4$\pm$3.4&51.3$\pm$3.9\\
	MvDA&64.9$\pm$7.2&60.1$\pm$5.6&42.3$\pm$4.8&70.8$\pm$5.5&58.1$\pm$5.2&51.0$\pm$3.4&66.8$\pm$4.2&63.0$\pm$4.4\\
	MvMDA&68.2$\pm$9.0&52.3$\pm$6.4&48.1$\pm$5.2&69.2$\pm$6.1&54.6$\pm$3.4&48.7$\pm$4.0&60.3$\pm$2.7&49.0$\pm$2.2\\
	MvDA-VC&70.4$\pm$10.9&62.4$\pm$4.9&48.4$\pm$4.5&78.0$\pm$4.0&65.4$\pm$3.4&53.2$\pm$2.6&76.2$\pm$3.7&69.1$\pm$2.1\\
	MvHE&{\color{blue}78.6}$\pm$7.1&{\color{blue}72.1}$\pm$5.4&{\color{blue}57.1}$\pm$5.1&{\color{blue}80.9}$\pm$4.6&{\color{blue}70.2}$\pm$3.6&{\color{blue}59.6}$\pm$4.7&{\color{blue}78.6}$\pm$3.0&{\color{blue}71.5}$\pm$3.0\\
	KMvHE&{\color{red}80.9}$\pm$7.9&{\color{red}75.7}$\pm$5.0&{\color{red}59.0}$\pm$3.8&{\color{red}83.1}$\pm$4.5&{\color{red}72.5}$\pm$3.4&{\color{red}60.9}$\pm$4.4&{\color{red}78.9}$\pm$2.0&{\color{red}73.1}$\pm$2.6\\
	\hline
	\end{tabular}
	\vspace{-0.2cm}
\end{table*}

\begin{figure*}[htb]
	\centering
	\begin{minipage}{2.8cm}
		\includegraphics[width=2.8cm]{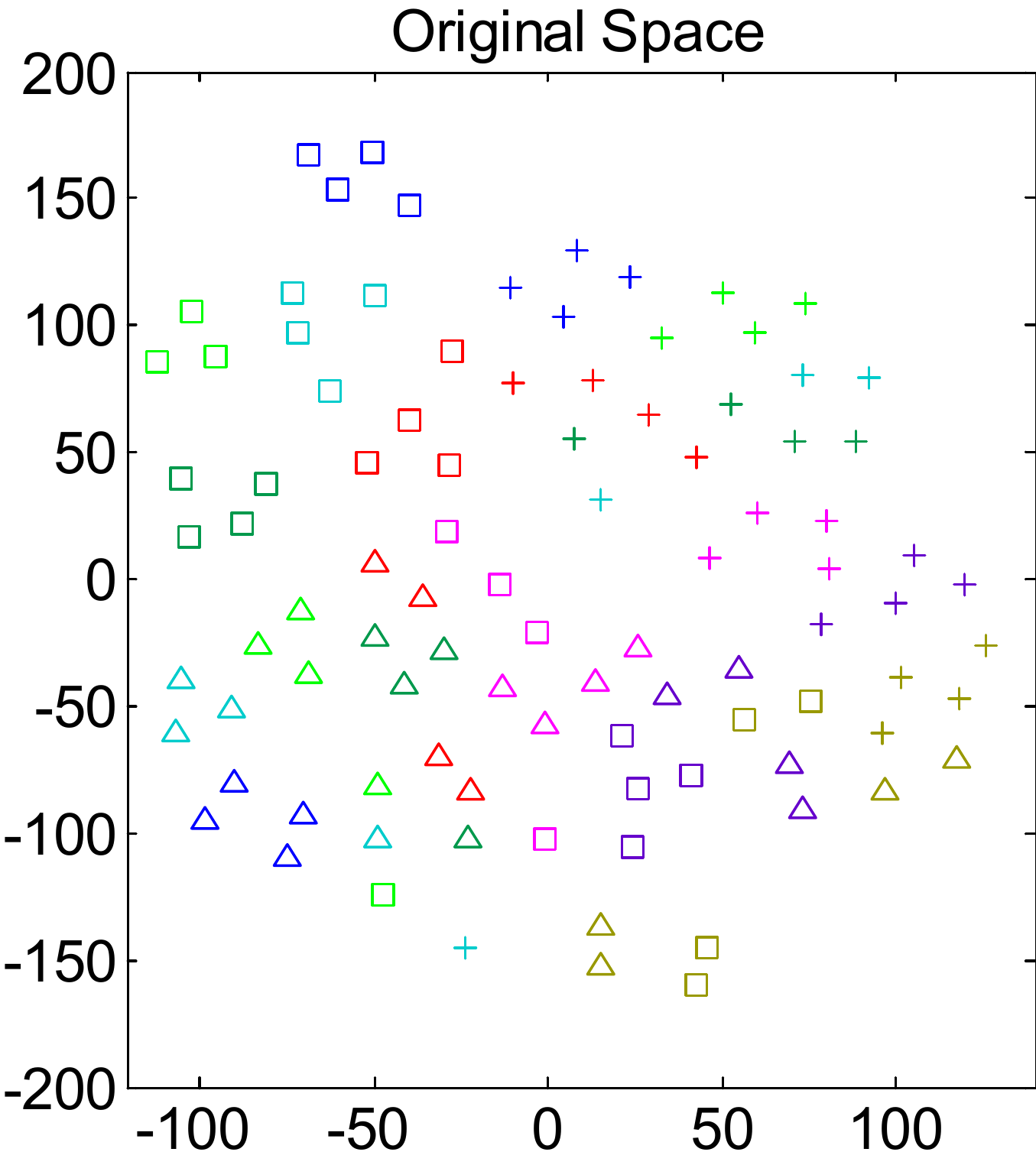}
	\end{minipage}
	\begin{minipage}{3cm}
		\includegraphics[width=3cm]{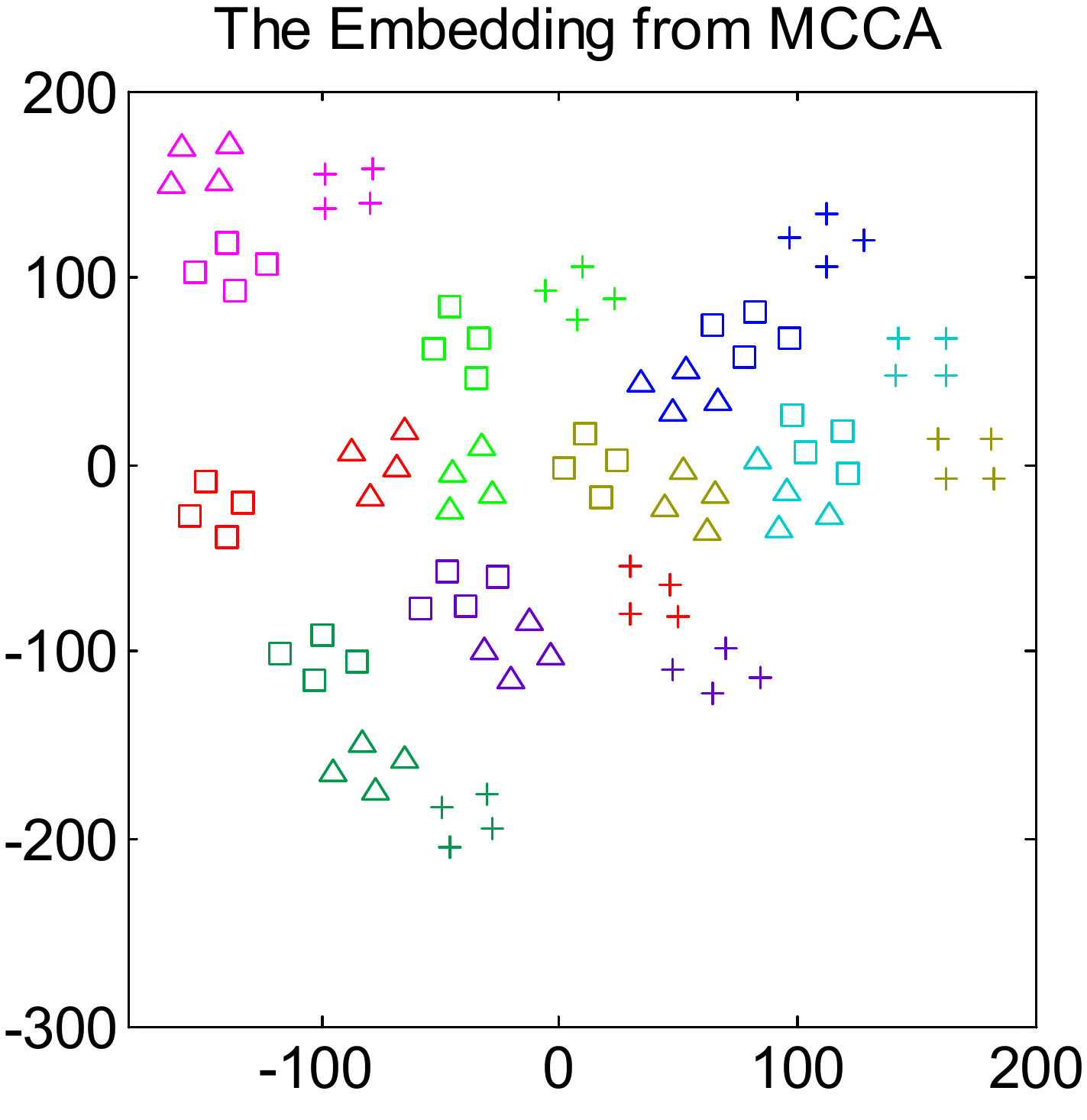}
	\end{minipage}
	\begin{minipage}{3cm}
		\includegraphics[width=3cm]{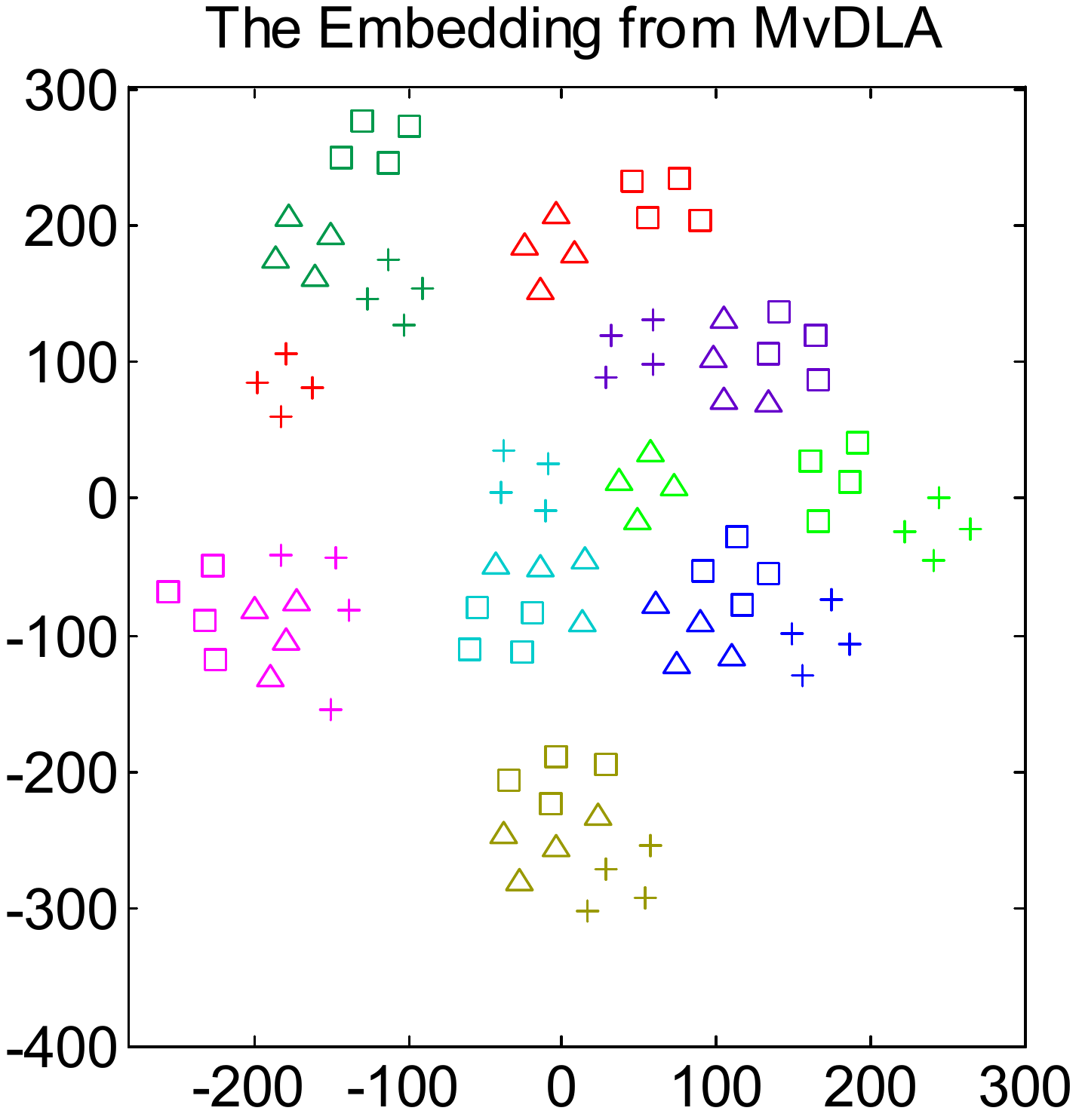}
	\end{minipage}
	\begin{minipage}{3cm}
		\includegraphics[width=3cm]{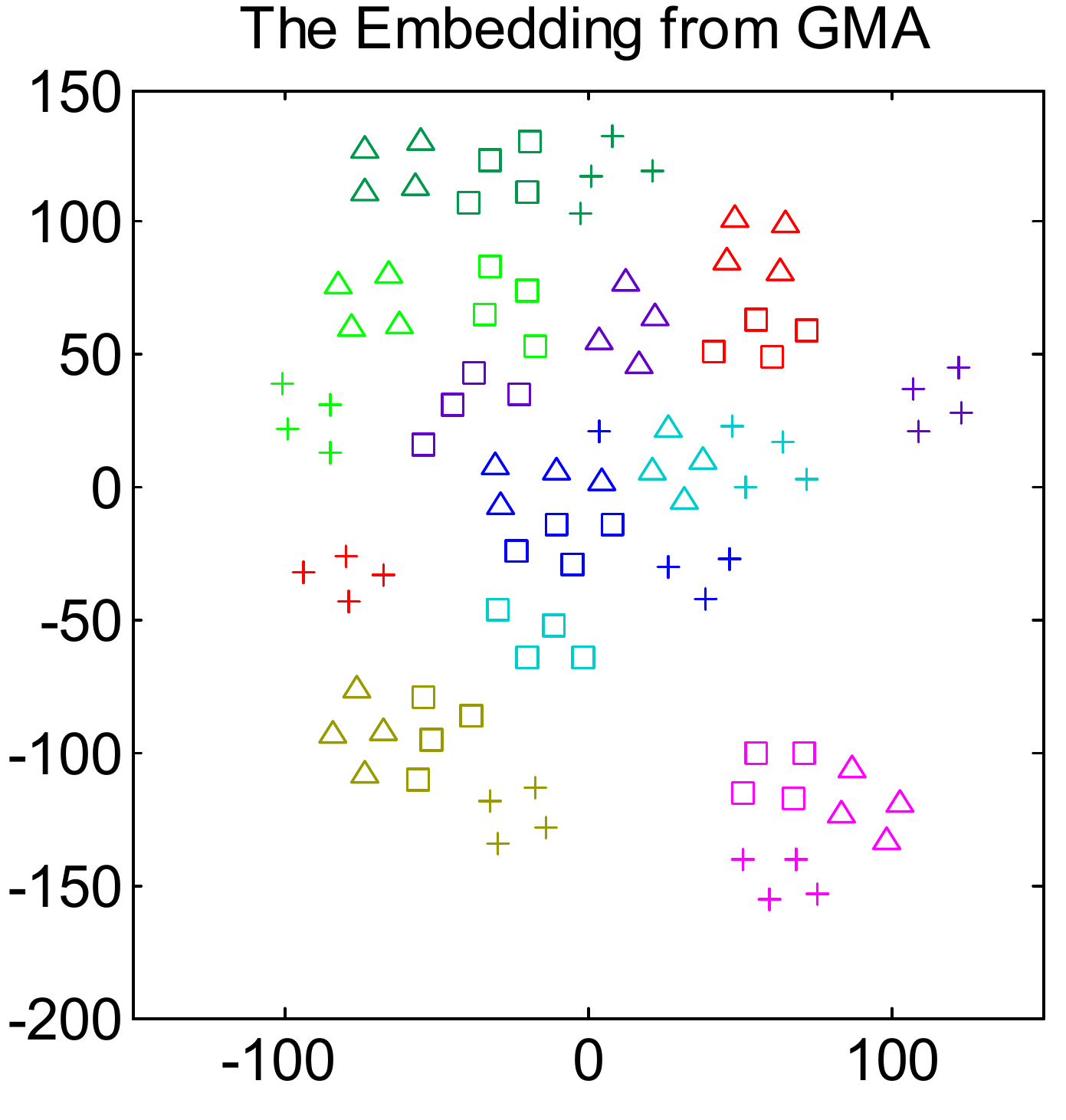}
	\end{minipage}
	\begin{minipage}{3cm}
		\includegraphics[width=3cm]{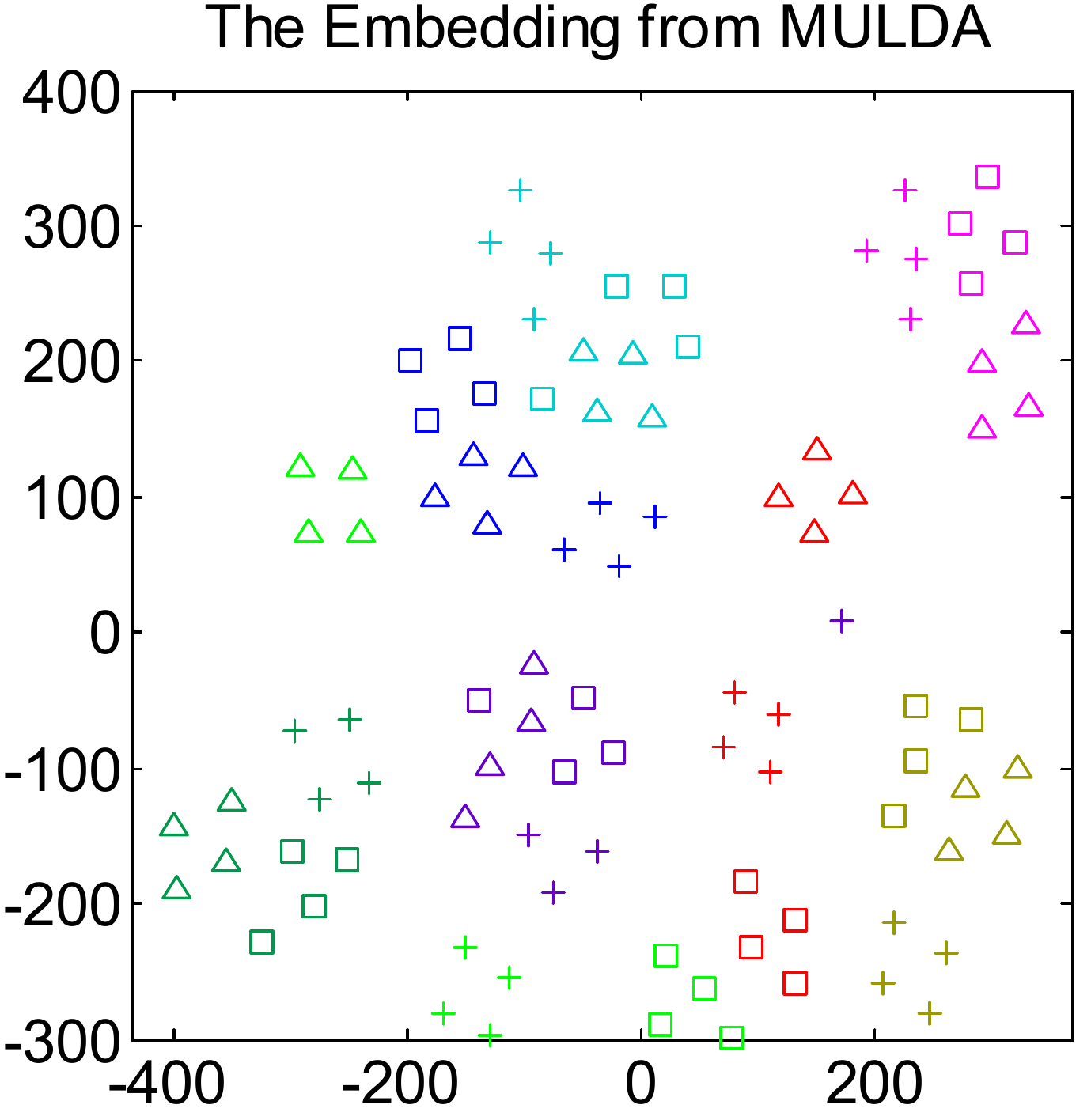}
	\end{minipage}\\
	\hspace{0.2cm}
	\begin{minipage}{3cm}
		\includegraphics[width=3cm]{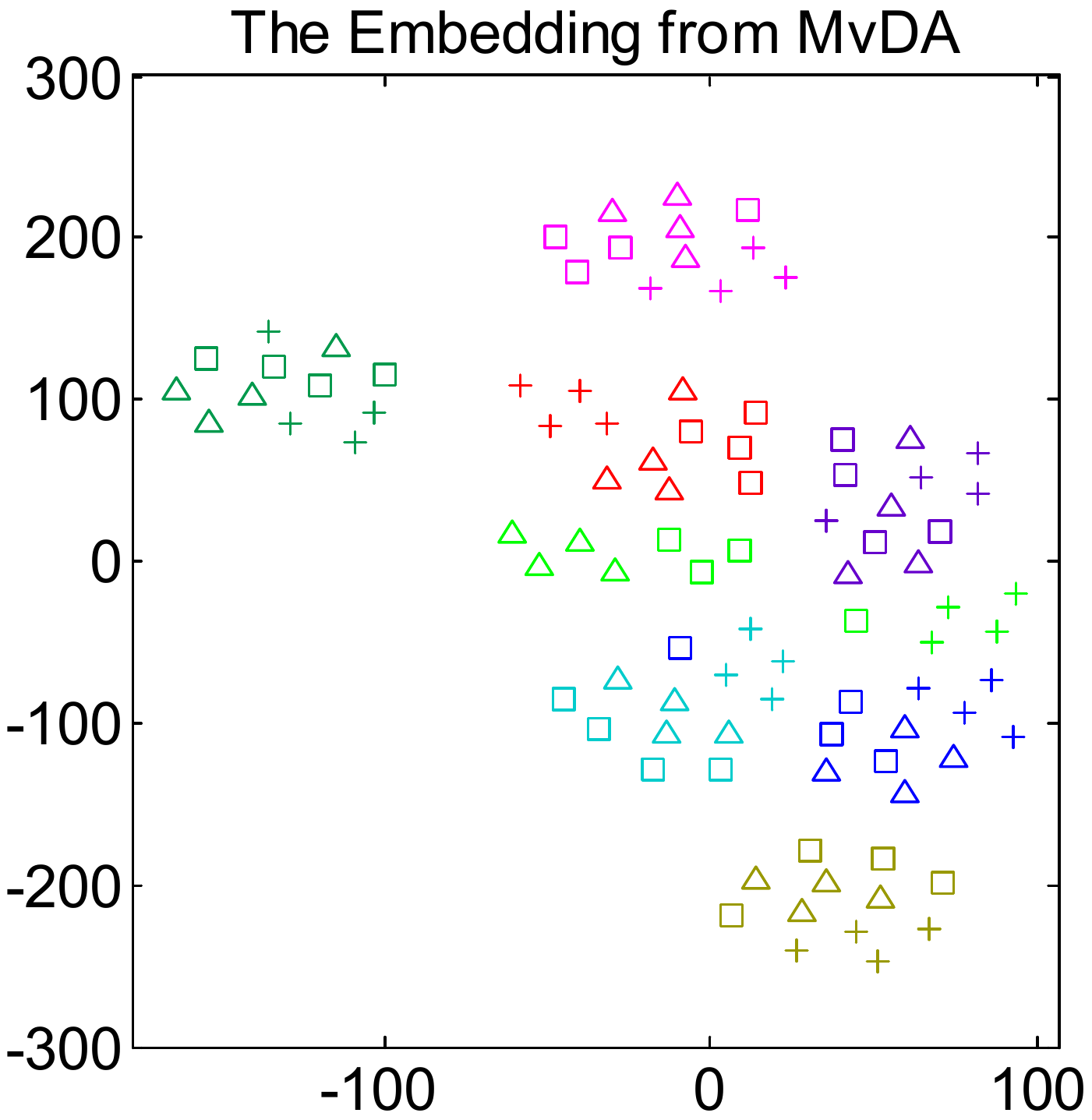}
	\end{minipage}
	\begin{minipage}{3cm}
		\includegraphics[width=3cm]{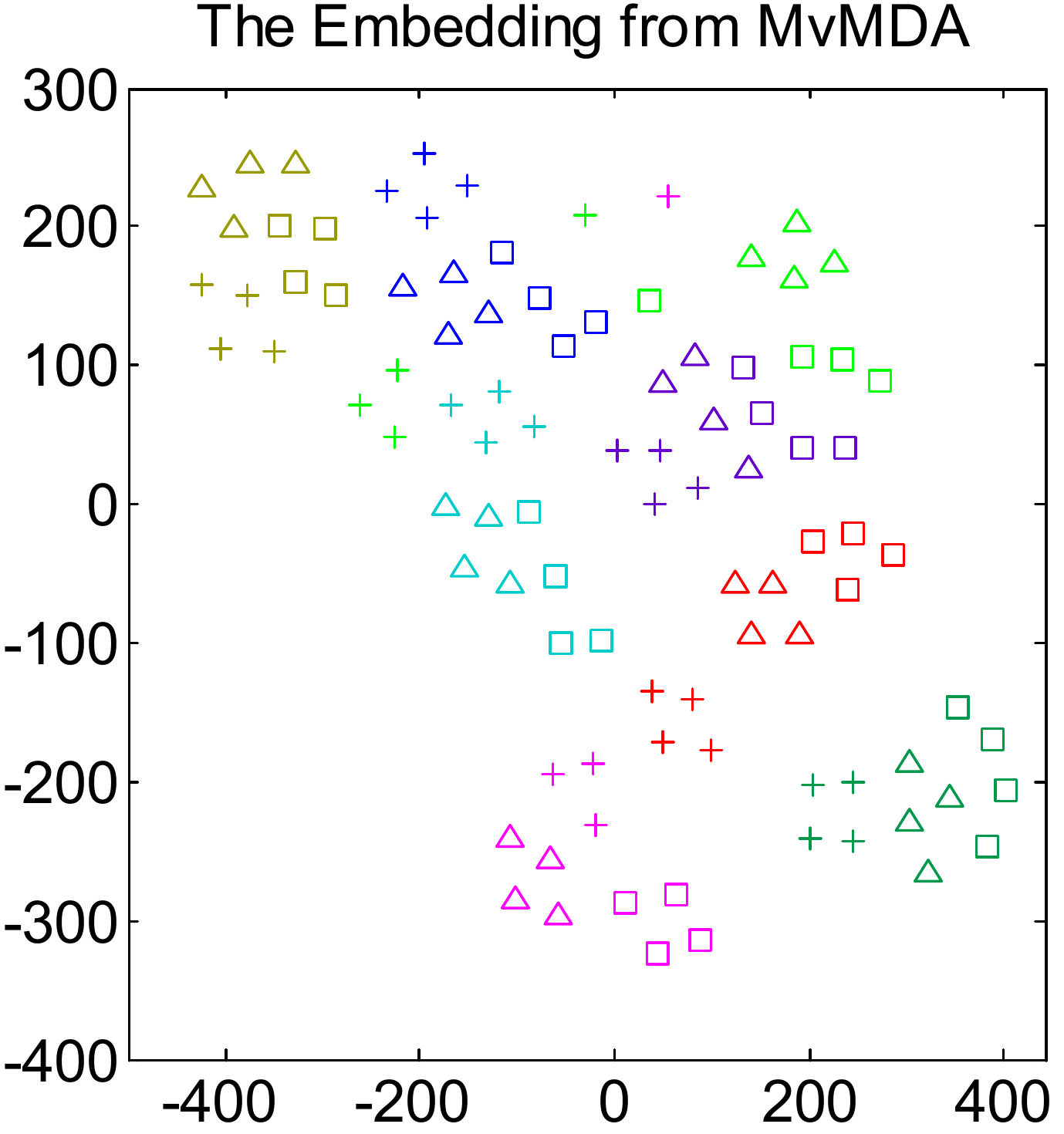}
	\end{minipage}
	\begin{minipage}{3cm}
		\includegraphics[width=3cm]{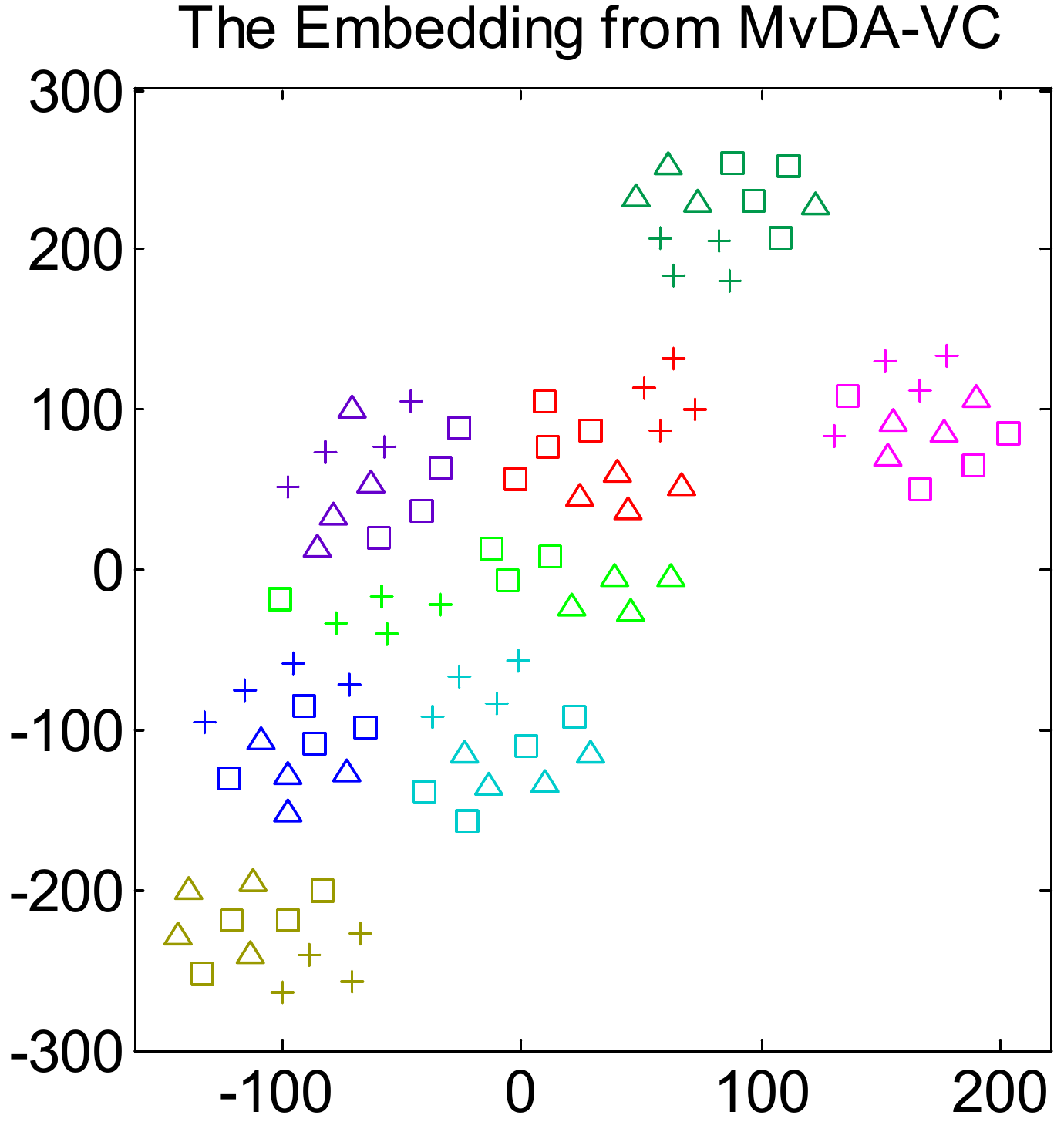}
	\end{minipage}
	\begin{minipage}{3cm}
		\includegraphics[width=3cm]{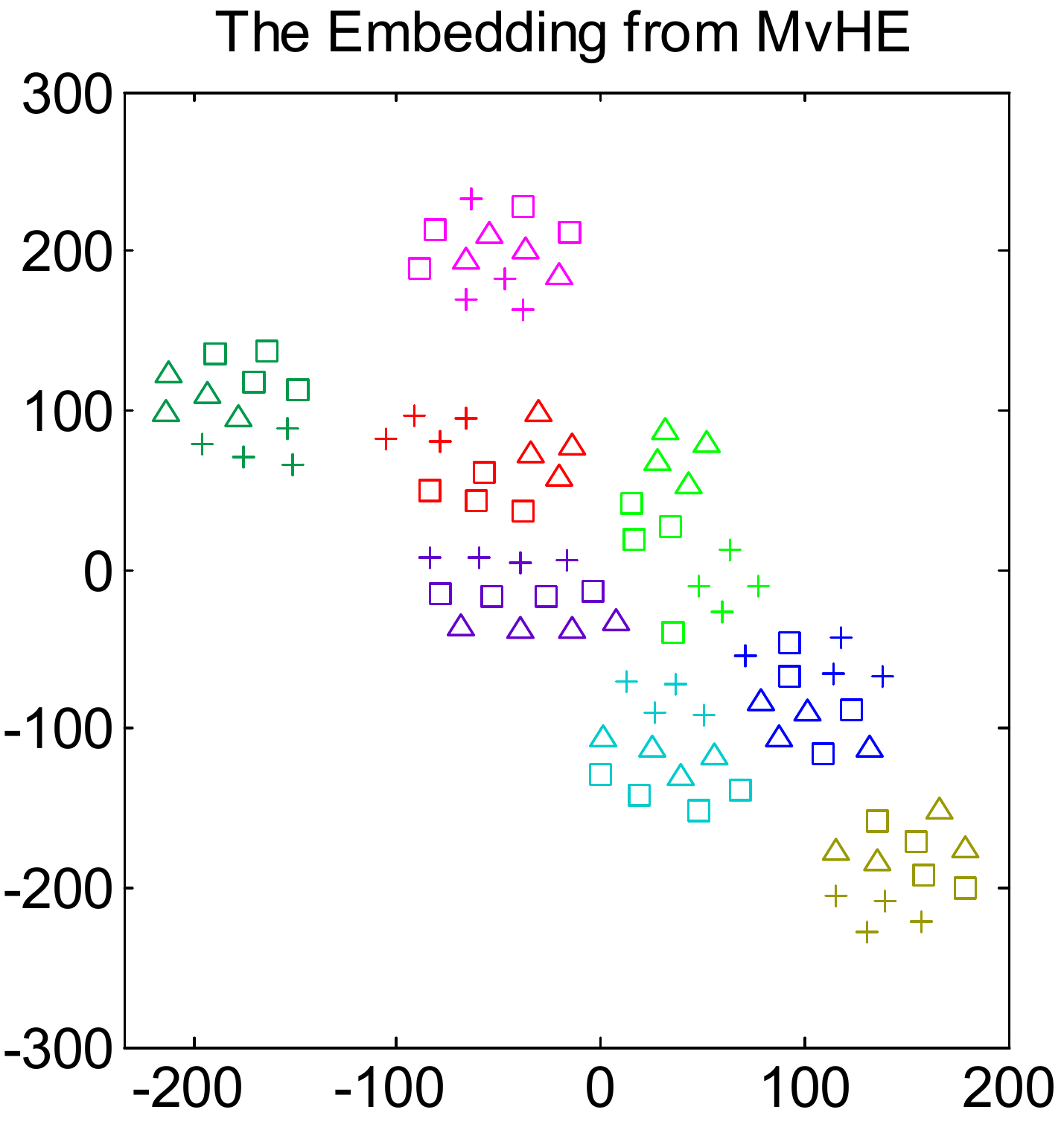}
	\end{minipage}
	\begin{minipage}{3.6cm}
		\includegraphics[width=3.4cm]{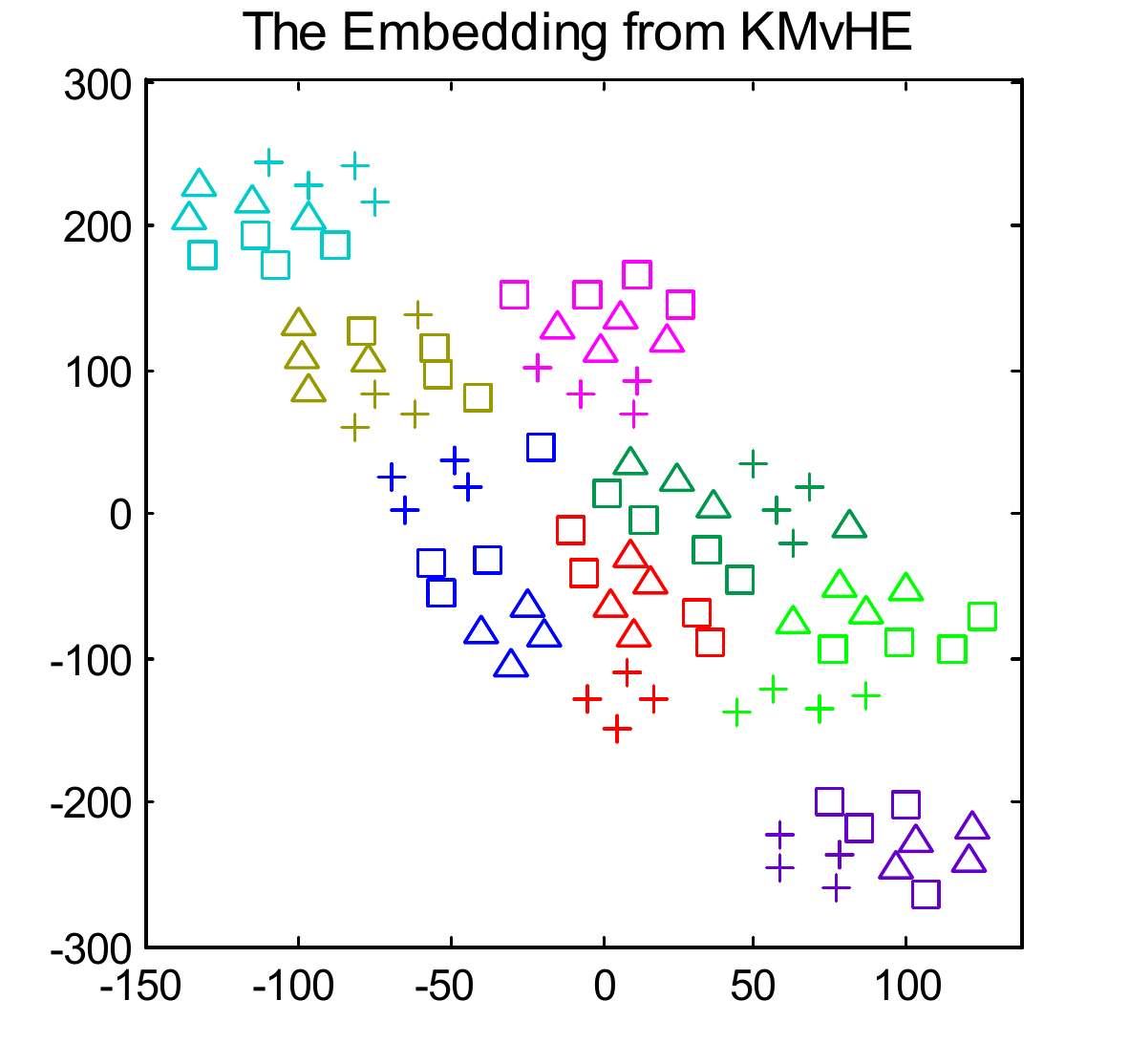}
	\end{minipage}
	\caption{Illustration of 2D embedding in case 5 of CMU PIE dataset. The embedding from MCCA, MvDLA, GMA, MULDA, MvDA, MvMDA, MvDA-VC, the proposed method MvHE and its kernel extension KMvHE. Different colors denote different classes, and different markers stand for different views. To make the figure clear, only the first 8 classes are shown.\vspace{-0.3cm}}
	\label{fig:visualize}
\end{figure*}


\subsection{The Efficacy of (K)MvHE on Multi-view Data}
\label{sec:face_pose}
In this section, we evaluate the performance of (K)MvHE on CMU PIE to demonstrate the superiority of our methods on multi-view face data. CMU PIE~\cite{sim2002cmu} contains 41,368 images of 68 people with 13 different poses, 43 diverse illumination conditions, and 4 various expressions. In our experiments, seven poses (i.e., C14, C11, C29, C27, C05, C37 and C02) are selected to construct multi-view data~(see Fig.~\ref{fig:CMU_PIE} for exemplar subjects). Each person at a given pose has 4 normalized grayscale images of size $64\times64$. This database is used to evaluate face classification across different poses. Similar to~\cite{ding2014low,ding2015deep}, we conduct experiments in 8 cases~(i.e., subsets), where case 1: $\{\textrm{C}27, \textrm{C}29\}$, case 2: $\{\textrm{C}27, \textrm{C}11\}$, case 3: $\{\textrm{C}27, \textrm{C}14\}$, case 4: $\{\textrm{C}05, \textrm{C}27, \textrm{C}29\}$, case 5: $\{\textrm{C}37, \textrm{C}27, \textrm{C}11\}$, case 6: $\{\textrm{C}02, \textrm{C}27, \textrm{C}14\}$, case 7: $\{\textrm{C}37, \textrm{C}05, \textrm{C}27, \textrm{C}29, \textrm{C}11\}$ and case 8: $\{\textrm{C}02, \textrm{C}37, \textrm{C}05, \textrm{C}27, \textrm{C}29, \textrm{C}11, \textrm{C}14\}$. During training phase, 45 people are randomly selected for each case and the remaining people are used for testing. Taking case 5 as an example, there are 3 poses in the case and 4 images are selected for each person. Hence, $3\times45\times4=540$ samples are selected for training, and the remaining $3\times23\times4=276$ samples are used for testing.
For testing in each case, the images from one view are used as gallery while the images from another view are used as probe, and all pairwise results are averaged as the mean accuracy~(mACC). For example, samples in case 5 contain 3 views, thus leading to $3\!\times\!2\!=\!6$ classification accuracy, we average all 6 results as the mACC.

\begin{figure*}[h]
	\centering
	\begin{minipage}{5cm}
		\includegraphics[width=5cm]{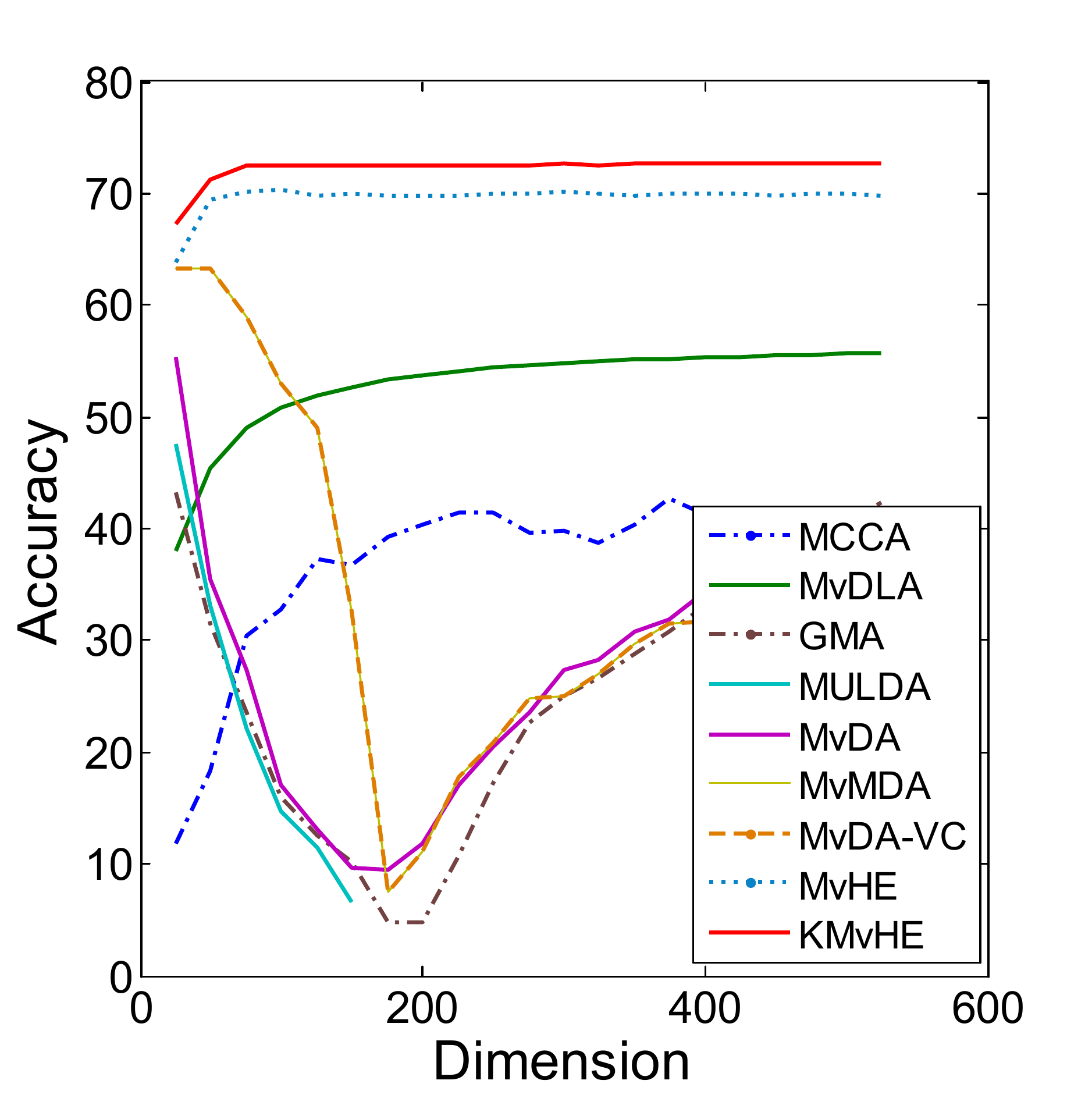}
		\subcaption{}
	\end{minipage}%
	\begin{minipage}{5cm}
		\includegraphics[width=5cm]{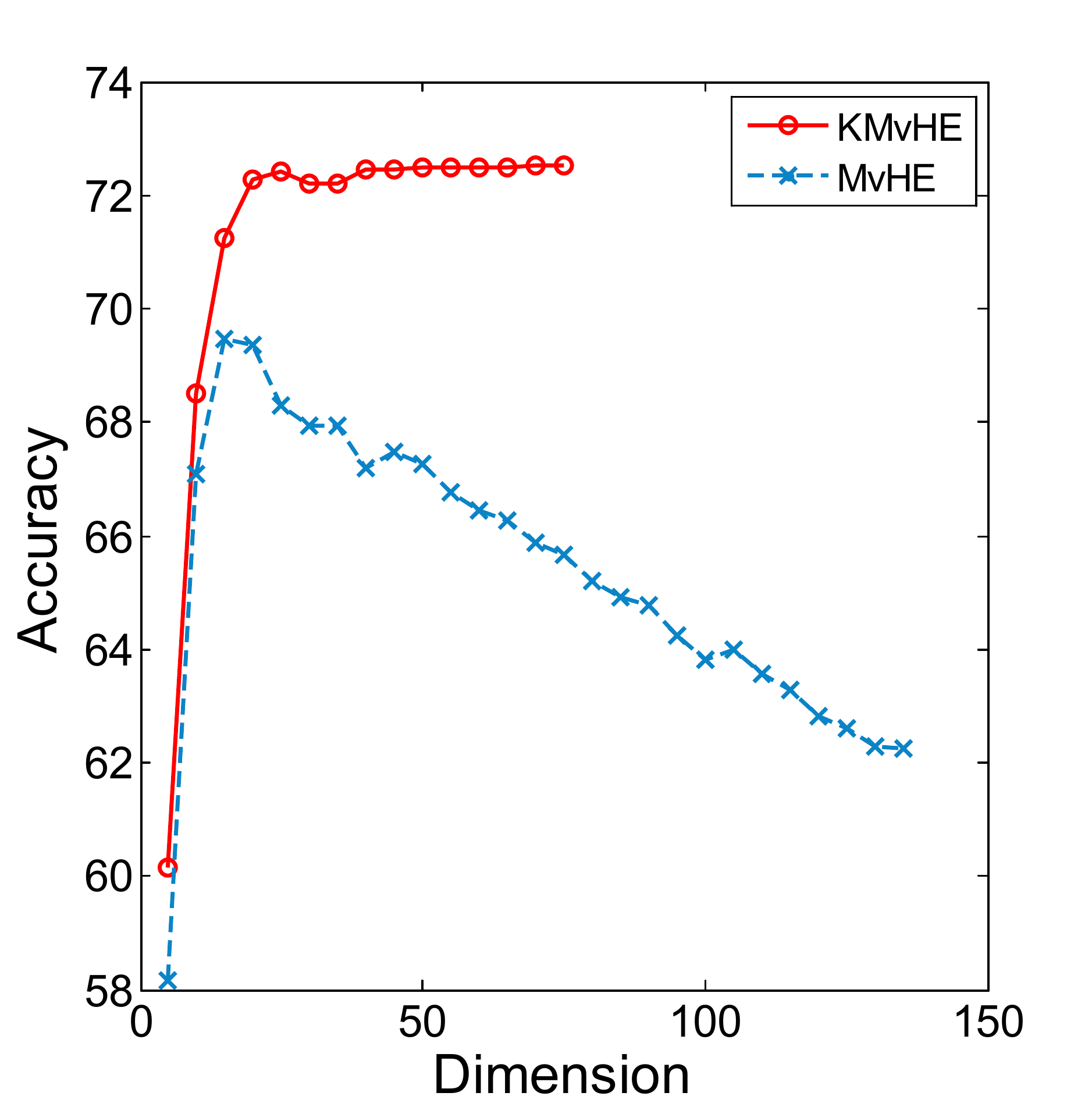}
		\subcaption{}
	\end{minipage}%
	\begin{minipage}{5cm}
		\includegraphics[width=5cm]{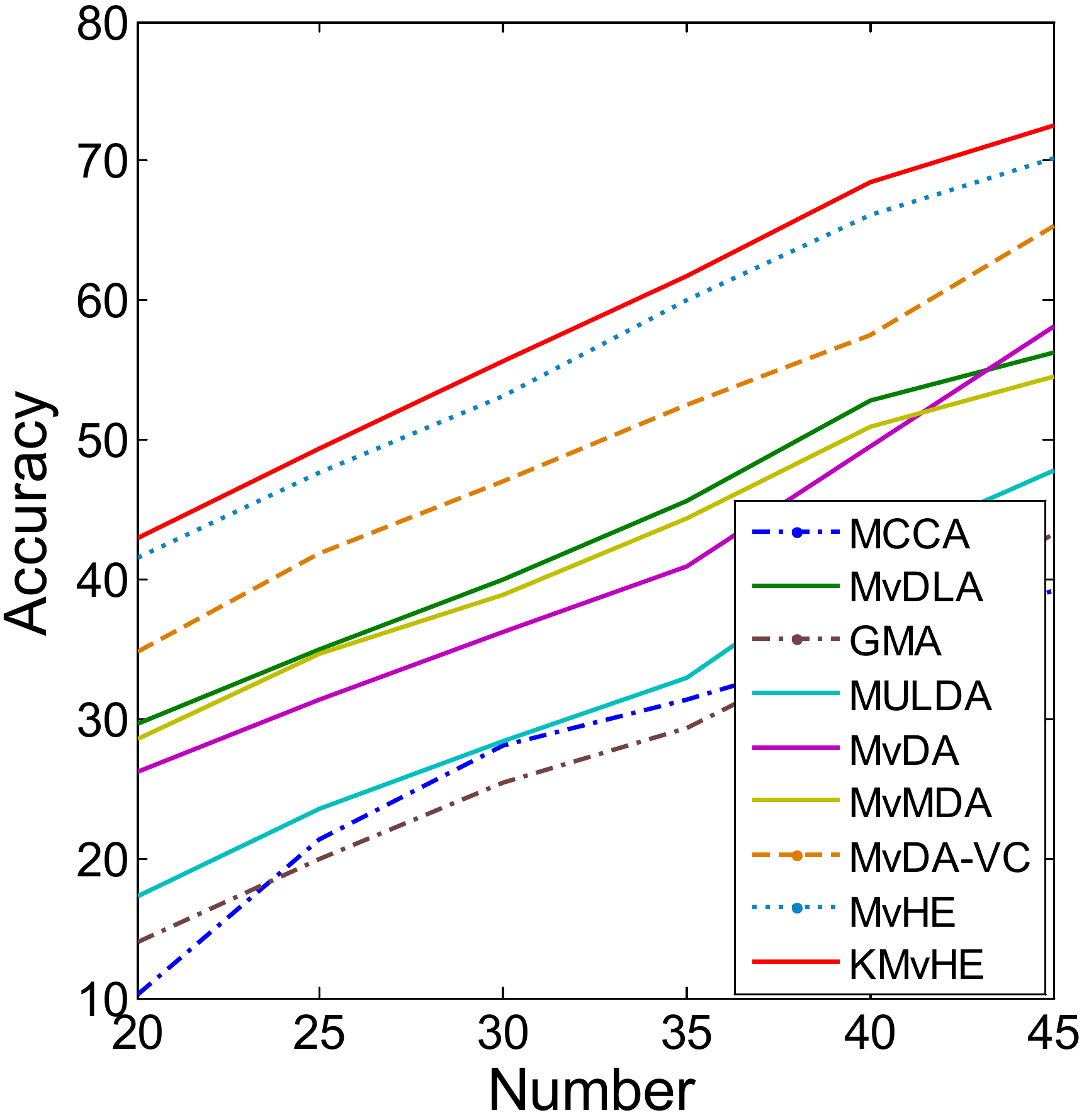}
		\subcaption{}
	\end{minipage}%
	\caption{Experimental results in case 5 of CMU PIE dataset: (a) shows the mACC of all competing methods with different PCA dimensions; (b) shows the mACC of MvHE and KMVHE with different latent subspace dimensions~(after PCA projection); (c) shows the mACC for all competing methods with respect to different training size (i.e., number of people). \vspace{-0.5cm}}
	\label{fig:cmu_pie_figure}
\end{figure*}

\subsubsection{Component-wise contributions of (K)MvHE} \label{sec:component_wise}
In the first experiment, we compare MvHE with its different baseline variants to demonstrate the component-wise contributions to the performance gain. In each case, 45 people are selected randomly as training data and the remaining samples from another 23 people are used for testing. Table~\ref{tab:layers_compared} summarized the performance differences among MvHE and its six baseline variants. Note that an orthogonal constraint with respect to mapping functions is imposed on all models. As can be seen, models that only consider one attribute~(i.e., \textit{LE-paired}, \textit{LDE-intra} and \textit{LDE-inter}) rank the lowest. Moreover, it is interesting to find that \textit{LDE-inter} consistently outperforms \textit{LE-paired} and \textit{LDE-intra}. This is because the view discrepancy can be partially mitigated by increasing inter-view discriminability~\cite{kan2016multi}. Moreover, either the combination of \textit{LE-paired} and \textit{LDE-intra} or the combination of \textit{LE-paired} and \textit{LDE-inter} can obtain remarkable improvements due to effectively handling view discrepancy. Our MvHE performs the best as expected. The results indicate that the elimination of view discrepancy, the preservation of local geometry and the increase of discriminability are essentially important for the task of cross-view classification. Therefore, the \textit{Divide-and-Conquer} strategy is a promising and reliable avenue towards enhancing cross-view classification performance.

\subsubsection{Superiority of (K)MvHE}
We then evaluate the performance of all competing algorithms in 8 cases. The performance of CCA and KCCA is omitted, because they can only handle two-view data. The results are summarized in Table~\ref{tab:cmu_pie_all_cases}.

As can be seen, MCCA performs poorly due to the neglect of discriminant information. MULDA has higher accuracy than GMA in most of cases by further mitigating redundant features. Furthermore, MvDLA, MvDA, MvMDA and MvDA-VC perform better than MCCA, GMA and MULDA due to the exploitation of inter-view supervised information. Our MvHE and KMvHE outperform MvDLA, MvDA, MvMDA and MvDA-VC with a large performance gain due to the preservation of intra-view and inter-view local geometric structures. The results validate the efficacy and superiority of our methods on multi-view data. Fig.~\ref{fig:visualize} displays the embeddings generated by nine algorithms in case 5. It is obvious that the embeddings corroborate classification accuracy. 

\subsubsection{Robustness Analysis of (K)MvHE}
\label{sec:face_pose_robsut}
In order to evaluate the robustness of different methods, we randomly permute the the class labels of $5\%$, $10\%$, $15\%$ and $20\%$ paired samples in case 5 and report the classification results in Table~\ref{tab:outlier_test_pie}. The values in parentheses indicate the relative performance loss $(\%)$ with respect to the scenario in the absence of outliers. As can be seen, GMA, MULDA, MvDA, MvMDA and MvDA-VC are much sensitive to outliers than MvHE and KMvHE. Take the $20\%$ permutation scenario as an example, our KMvHE only suffers from a relative $3.9\%$ performance drop from its original $72.5\%$ accuracy, whereas the accuracy of GMA decreases to $30.0\%$ with the relative performance drop nearly $31\%$. One probable reason for the superior robustness of our methods is the adopted local optimization~\cite{davoudi2017dimensionality,belkin2003laplacian}. One should note that the performance of MvDLA degrades dramatically with the increase of outliers. This, agian, validates that MvDLA cannot preserve local geometry of data with large $p_1$. We also report the training time in Table~\ref{tab:training_times}. As we have analyzed in Sect.~\ref{sec_complexity}, the computational time for all the competing methods is approximate in the same range and our methods do not exhibit any advantage in time complexity due to the increment of model complexity caused by the preservation of local geometry. However, considering the overwhelming performance gain, the extra computational cost is still acceptable. Moreover, the computational complexity of MvDLA is even higher than MvHE. This is because large $p_1$ significantly increases the cost of selection matrix calculation.

\begin{table}[htb]
    \centering
	\caption{The average classification accuracy $(\%)$ in terms of mean accuracy (mACC) in case 5 of CMU PIE dataset with outliers. The best two results are marked with {\color{red}red} and {\color{blue}blue} respectively. The values in parentheses indicate the relative performance loss $(\%)$ with respect to the scenario in the absence of outliers. ``PR" denotes permutation ratio.}
	\label{tab:outlier_test_pie}
	\scalebox{0.9}[0.9]{
	\begin{tabular}{c|c|c|c|c|c}
	\hline
	PR&0\%&5\%&10\%&15\%&20\%\\
	\hline
	MvDLA&56.2~(0.0)&49.4~(12.1)&41.1~(26.9)&37.0~(34.2)&32.5~(42.2)\\
	GMA&43.2~(0.0)&37.2~(13.9)&34.0~(21.3)&31.8~(26.4)&30.0~(30.6)\\
	MULDA&47.8~(0.0)&41.9~(12.3)&39.9~(16.5)&39.0~(18.4)&37.9~(20.7)\\
	MvDA&58.1~(0.0)&44.5~(23.4)&39.4~(32.2)&36.4~(37.4)&32.8~(43.6)\\
	MvMDA&54.6~(0.0)&52.2~(4.4)&49.6~(9.2)&46.4~(15.0)&42.1~(22.9)\\
	MvDA-VC&65.4~(0.0)&56.8~(13.1)&52.0~(20.5)&51.2~(21.7)&47.3~(27.7)\\
	MvHE&{\color{blue}70.2}~(0.0)&{\color{blue}68.2}~(2.9)&\textcolor{blue}{67.0}~(4.6) &{\color{blue}66.1}~(5.8)&{\color{blue}62.3}~(11.3)\\
	KMvHE&{\color{red}72.5}~(0.0)&{\color{red}72.2}~(0.4)&\textcolor{red}{71.3}~(1.7)&{\color{red}70.9}~(2.2)&{\color{red}69.7}~(3.9)\\
	\hline
	\end{tabular}
	}
	\vspace{-0.3cm}
\end{table}
	
\begin{table}[htb]
    \centering
	\caption{The training time (seconds), averaged over 10 iterations, of nine MvSL based methods on CMU PIE dataset (case 5). \textbf{Bold} denotes the best results.}
	\label{tab:training_times}
	\begin{tabular}{c|c|c|c|c}
	\hline
	Methods&Case 1&Case 4&Case 7&Case 8\\
	\hline
	MCCA&14&27&64&117\\
	MvDLA&24&63&256&564\\
	GMA&13&25&54&96\\
	MULDA&14&25&56&100\\
	MvDA&\textbf{12}&\textbf{22}&{52}&\textbf{92}\\
	MvMDA&14&26&57&101\\
	MvDA-VC&{12}&{23}&\textbf{51}&{92}\\
    MvHE&21&48&158&327\\
	KMvHE&21&46&170&345\\
    \hline
	\end{tabular}
	\vspace{-0.3cm}
\end{table}	



Finally, to further demonstrate the performance of our methods with respect to different PCA dimensions, projection subspace dimensions and the size of training set. We vary the PCA dimensions from 25 to 525 at an interval of 25, the latent subspace dimensions from 5 to its maximum value at an interval of 5 and the number of people in the training set from 20 to 45 also at an interval of 5. The experimental results in case 5 are shown in Fig.~\ref{fig:cmu_pie_figure}. Note that the classification accuracy of MvDA (and MULDA) at PCA dimension 425 (and 150) or above is omitted in Fig.~\ref{fig:cmu_pie_figure}(a). This is because their singular value decomposition does not converge when PCA dimension exceeds 425 (150) for MvDA (MULDA). As can be seen, our MvHE and KMvHE are more robust to PCA dimensions compared with GMA, MvDA and MvDA-VC. Moreover, with the increase of training data size, our methods are consistently better than their MvSL counterparts. If we further compare our MvHE with KMvHE, it is interesting to find that MvHE is more sensitive to subspace dimensions, whereas the performance of KMvHE remains stable when subspace dimension exceeds 20. 

\begin{figure}[htb]
	\centering
	\includegraphics[height=3.5cm,width=8cm]{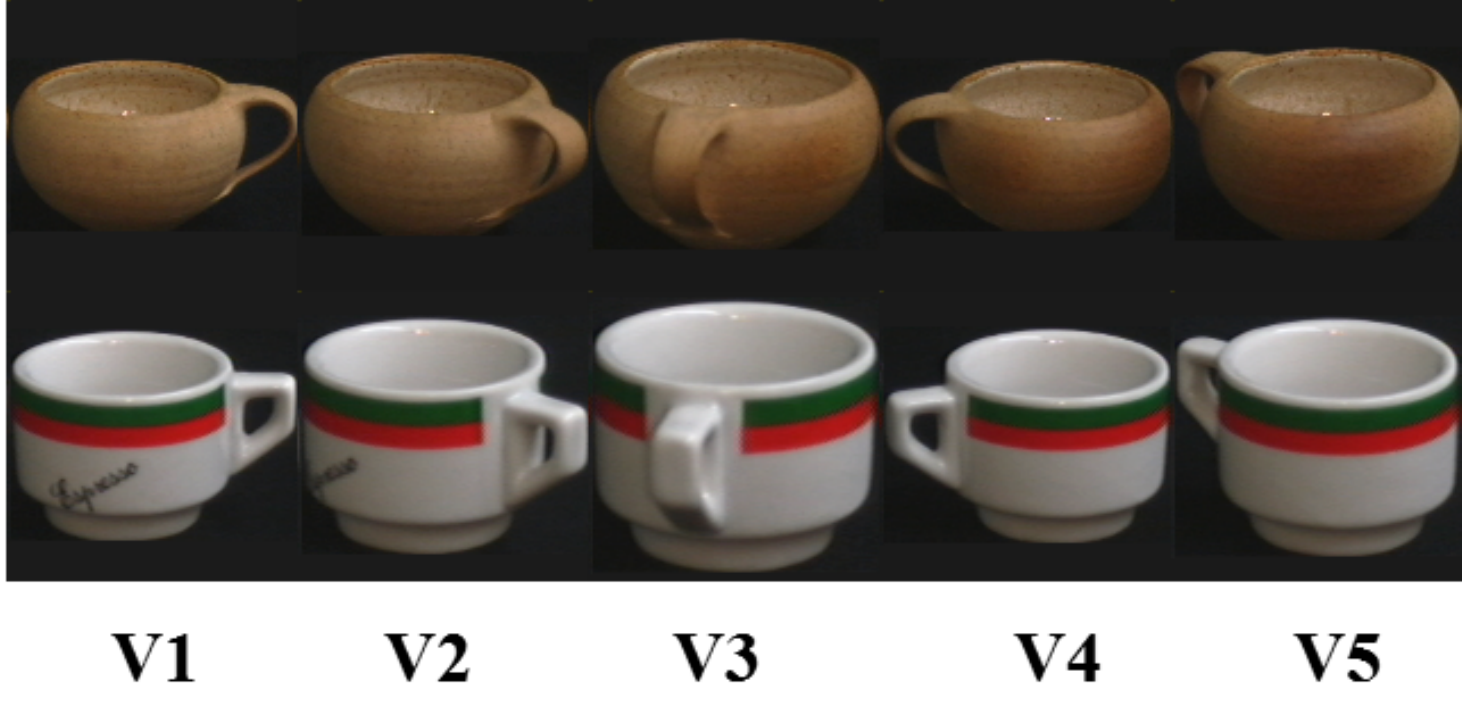}
	\caption{Exemplar objects of two classes from the COIL-100 dataset. V1, V2, V3, V4 and V5 are selected to construct
multi-view data.\vspace{-0.5cm}}
	\label{fig:coil-100}
\end{figure}


\subsection{The Superiority of (K)MvHE on COIL-100 Dataset}
\label{sec:object_pose}
We also demonstrate the efficacy of (K)MvHE on natural images. The COIL-100 dataset contains 100 objects. Each object has 72 images, and each is captured with 5-degree rotation. In our experiments, five poses (i.e., V1: [$0^\circ$, $15^\circ$], V2: [$40^\circ$, $55^\circ$], V3: [$105^\circ$, $120^\circ$], V4: [$160^\circ$, $175^\circ$] and V5: [$215^\circ$, $230^\circ$]) are selected to construct multi-view data (see Fig.~\ref{fig:coil-100} for exemplar objects). Each image is converted to grayscale and downsampled to the size of $48\times48$. The COIL-100 dataset is used to evaluate object classification across poses. Similar to CMU PIE, experiments are conducted in 4 cases, where case 1: $\{\textrm{V}1, \textrm{V}2\}$, case 2: $\{\textrm{V}1, \textrm{V}2, \textrm{V}3\}$, case 3: $\{\textrm{V}1, \textrm{V}2, \textrm{V}3, \textrm{V}4\}$ and case 4: $\{\textrm{V}1, \textrm{V}2, \textrm{V}3, \textrm{V}4, \textrm{V}5\}$. Testing in each case is also conducted in pairwise manner and all pairwise results are averaged as the mean accuracy~(mACC).

\begin{figure*}[h]
	\centering
	\begin{minipage}{5cm}
		\includegraphics[width=5cm]{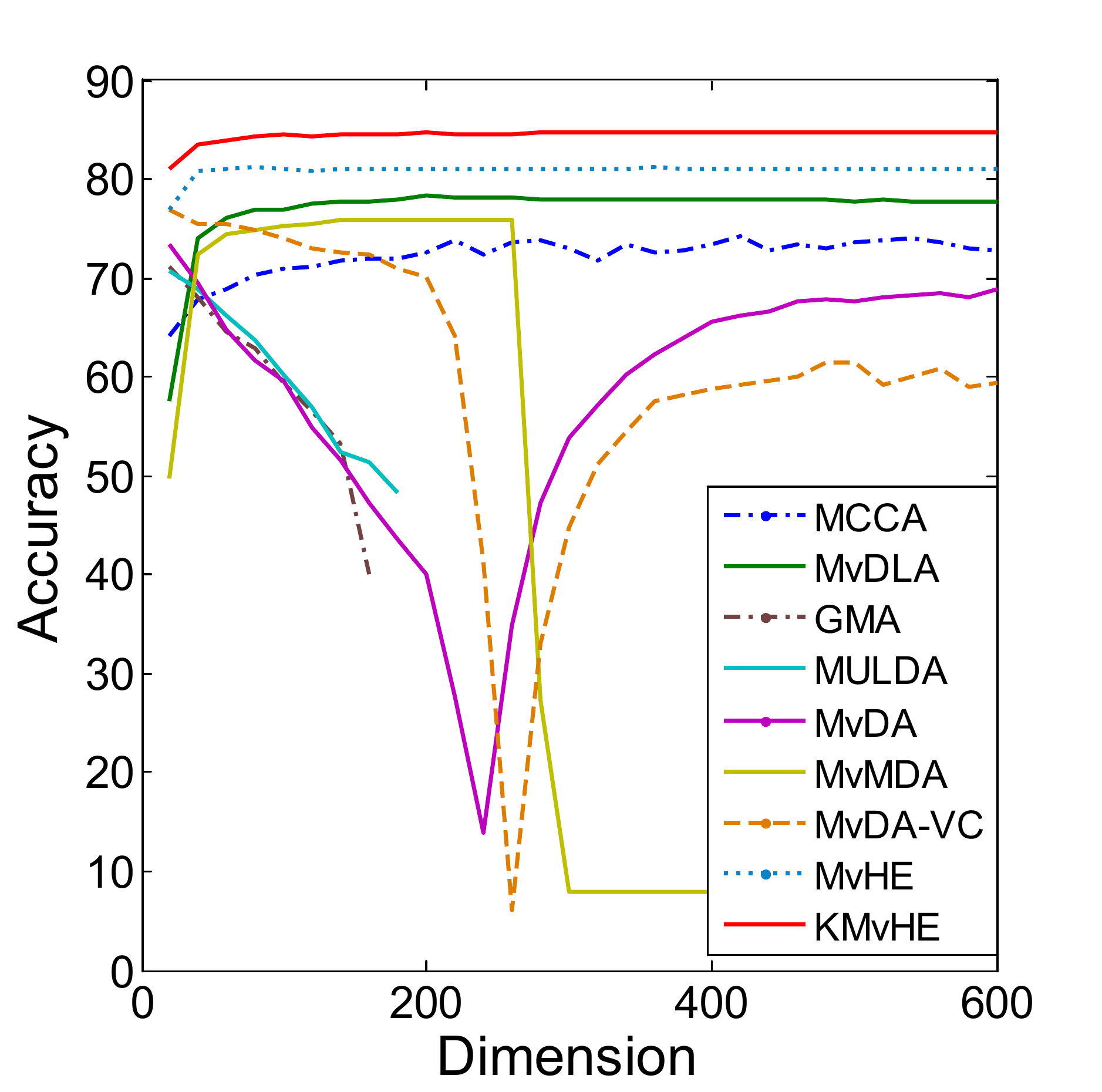}
		\subcaption{}
	\end{minipage}%
	\begin{minipage}{5cm}
		\includegraphics[width=5cm]{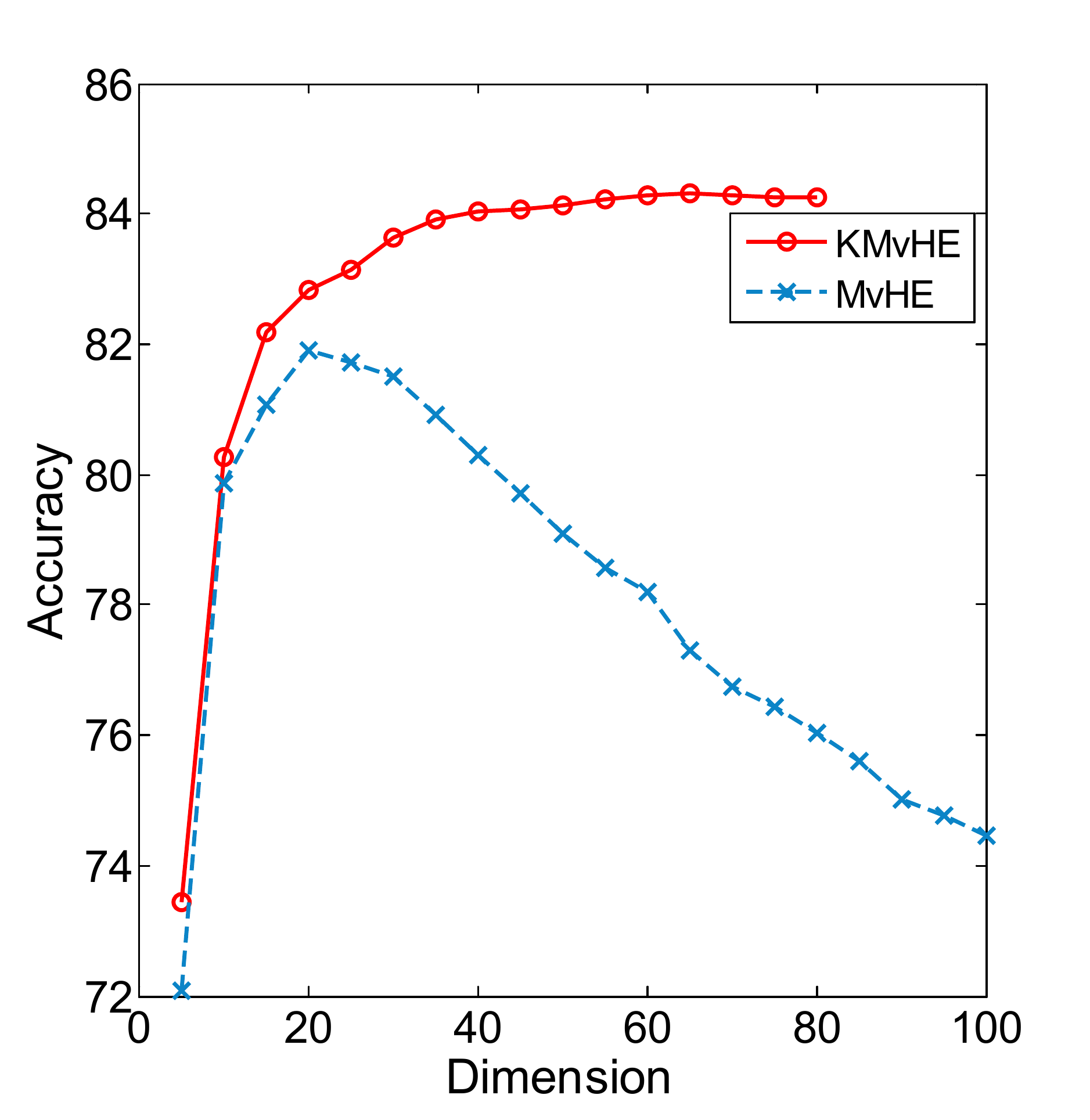}
		\subcaption{}
	\end{minipage}%
	\begin{minipage}{5cm}
		\includegraphics[width=5cm]{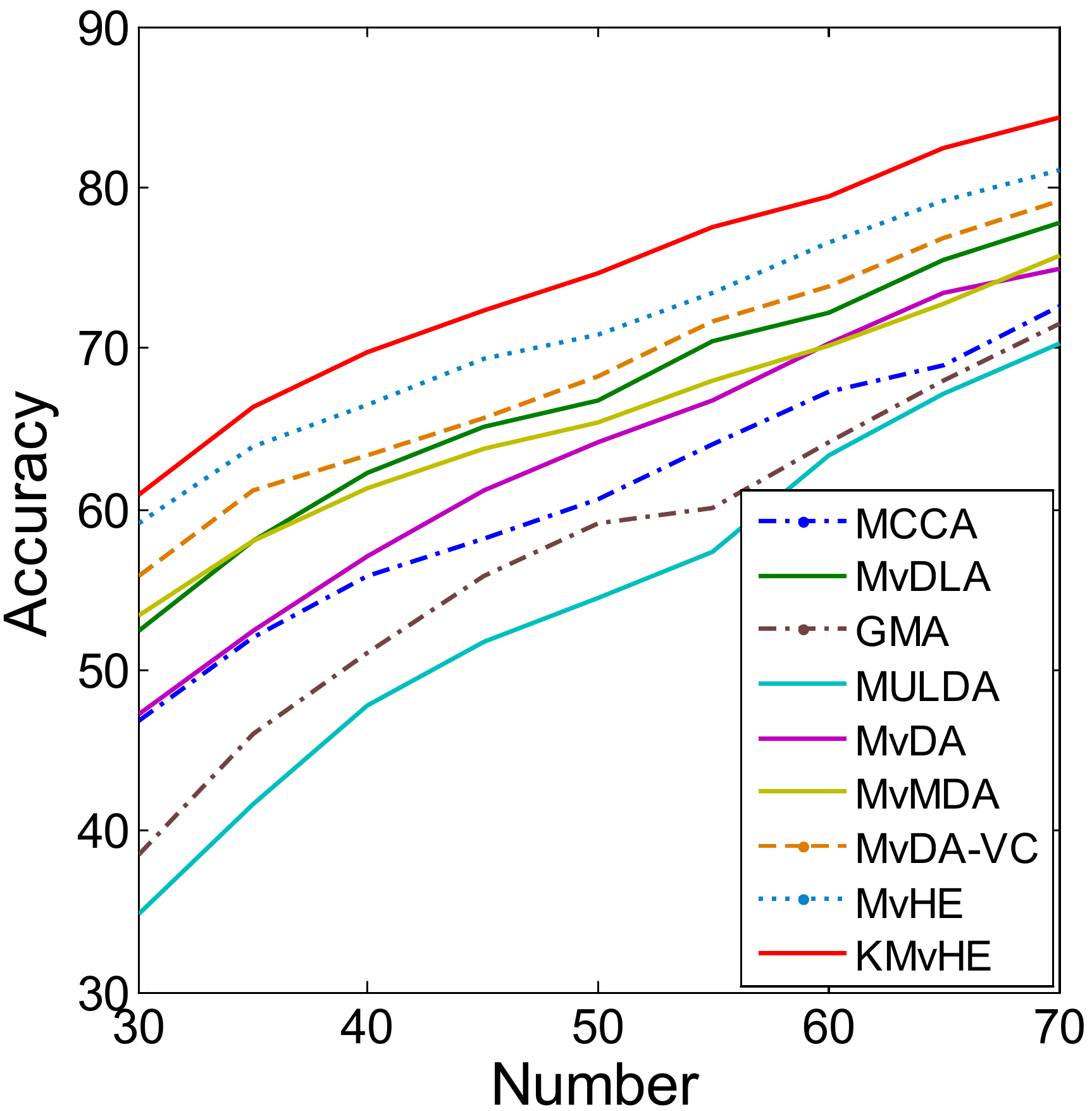}
		\subcaption{}
	\end{minipage}%
	\caption{Experimental results in case 2 of COIL-100 dataset: (a) shows the mACC of all competing methods with different PCA dimensions; (b) shows the mACC of MvHE and KMVHE with different latent subspace dimensions~(after PCA projection); (c) shows the mACC for all competing methods with respect to different training size (i.e., number of objects).\vspace{-0.5cm}}
	\label{fig:coil_figure}
\end{figure*}

We repeat the similar experimental setting in Sect.~\ref{sec:face_pose}. Table~\ref{tab:case_coil} shows the classification accuracy in 4 cases when 70 objects are selected for training and the remaining is used for testing. It is obvious that our MvHE and KMvHE achieve the best two results in all cases. To evaluate the robustness of different methods, we randomly permute the class labels of $10\%$, $20\%$, $30\%$ and $40\%$ paired samples in case 2 and report the classification results in Table~\ref{tab:outlier_test_coil}. Again, among all involved competitors, our methods are the most robust ones to different degrees of outliers. Finally, to demonstrate the performance of our methods with respect to different PCA dimensions, projection subspace dimensions and the size of training set. We vary the PCA dimensions from 20 to 600 at an interval of 20, the latent subspace dimensions from 5 to its maximum value at an interval of 5 and the number of people in the training set from 30 to 70 also at an interval of 5. The experimental results in case 2 are shown in Fig.~\ref{fig:coil_figure}. The general trends shown in each subfigure of Fig.~\ref{fig:coil_figure} are almost the same as shown in Fig.~\ref{fig:cmu_pie_figure}.


The above results further validate the superiority of our methods on classification accuracy and robustness. Moreover, the results also validate that the superiority of our methods are not limited to facial images.

\begin{table}[htb]
    \centering
	\caption{The average classification accuracy $(\%)$ of our methods and their multi-view counterparts in terms of mean accuracy (mACC) on COIL-100 dataset. The best two results are marked with {\color{red}red} and {\color{blue}blue} respectively.}
	\label{tab:case_coil}
	\begin{tabular}{c|c|c|c|c}
	\hline
	Methods&case 1&case 2&case 3&case 4\\
	\hline
	PCA&69.1$\pm$5.5&60.9$\pm$4.3&60.5$\pm$4.1&61.4$\pm$4.4\\
	LDA&44.8$\pm$5.3&41.9$\pm$4.6&43.1$\pm$4.7&43.8$\pm$4.9\\
	DLA&18.5$\pm$4.0&24.4$\pm$3.1&23.5$\pm$2.8&25.0$\pm$2.6\\
	MCCA&78.0$\pm$7.9&72.6$\pm$4.8&72.7$\pm$4.0&73.0$\pm$3.7\\
	MvDLA&86.4$\pm$3.1&77.8$\pm$1.9&74.0$\pm$2.3&71.7$\pm$3.9\\
	GMA&81.4$\pm$2.8&71.5$\pm$2.1&72.0$\pm$2.3&73.0$\pm$3.0\\
	MULDA&82.8$\pm$4.4&70.3$\pm$2.9&70.8$\pm$3.9&70.8$\pm$3.7\\
	MvDA&85.0$\pm$5.4&75.0$\pm$2.1&77.1$\pm$2.1&77.9$\pm$2.1\\
	MvMDA&87.1$\pm$3.3&75.8$\pm$2.7&75.4$\pm$3.3&75.4$\pm$3.3\\
	MvDA-VC&87.2$\pm$3.2&79.1$\pm$2.8&80.8$\pm$2.4&81.8$\pm$3.3\\
	MvHE&\textcolor{blue}{90.2}$\pm$3.0&\textcolor{blue}{81.1}$\pm$1.5&\textcolor{blue}{82.9}$\pm$2.4&\textcolor{blue}{84.1}$\pm$2.1\\
	KMvHE&\textcolor{red}{91.6}$\pm$2.6&\textcolor{red}{84.3}$\pm$1.7&\textcolor{red}{84.9}$\pm$2.5&\textcolor{red}{85.5}$\pm$2.4\\
	\hline
	\end{tabular}
	\vspace{-0.2cm}
\end{table}

\begin{table}[htb]
    \centering
	\caption{The average classification accuracy $(\%)$ in terms of mean accuracy (mACC) in case 2 of COIL-100 dataset with outliers. The best two results are marked with {\color{red}red} and {\color{blue}blue} respectively. The values in parentheses indicate the relative performance loss $(\%)$ with respect to the scenario in the absence of outliers. ``PR" denotes permutation ratio.}
	\label{tab:outlier_test_coil}
	\scalebox{0.9}[0.9]{
	\begin{tabular}{c|c|c|c|c|c}
	\hline
	PR&0\%&10\%&20\%&30\%&40\%\\
	\hline
	MvDLA&77.8~(0.0)&67.2~(13.6)&57.3~(26.4)&49.6~(36.9)&39.3~(49.5)\\
	GMA&71.5~(0.0)&66.2~(7.1)&63.6~(11.0)&61.7~(13.7)&59.1~(17.3)\\
	MULDA&70.3~(0.0)&61.5~(12.5)&61.3~(12.8)&56.9~(19.1)&55.5~(21.1)\\
	MvDA&75.0~(0.0)&66.8~(10.9)&62.3~(16.9)&58.8~(21.6)&56.4~(24.8)\\
	MvMDA&75.8~(0.0)&76.0~(-0.3)&74.6~(1.6)&71.5~(5.7)&67.7~(10.7)\\
	MvDA-VC&79.1~(0.0)&74.8~(5.4)&73.6~(7.0)&72.6~(8.2)&71.2~(10.0)\\
	MvHE&{\color{blue}81.1}~(0.0)&{\color{blue}80.5}~(0.7)&{\color{blue}79.9}~(1.5)&{\color{blue}79.0}~(2.6)&{\color{blue}78.3}~(3.5)\\
	KMvHE&{\color{red}84.3}~(0.0)&{\color{red}83.4}~(1.1)&{\color{red}82.6}~(2.0)&{\color{red}81.7}~(3.1)&{\color{red}81.5}~(3.3)\\
    \hline
	\end{tabular}
	}
	\vspace{-0.5cm}
\end{table}

\begin{figure*}[htb]
	\centering
	\begin{minipage}{3cm}
		\includegraphics[width=3.3cm]{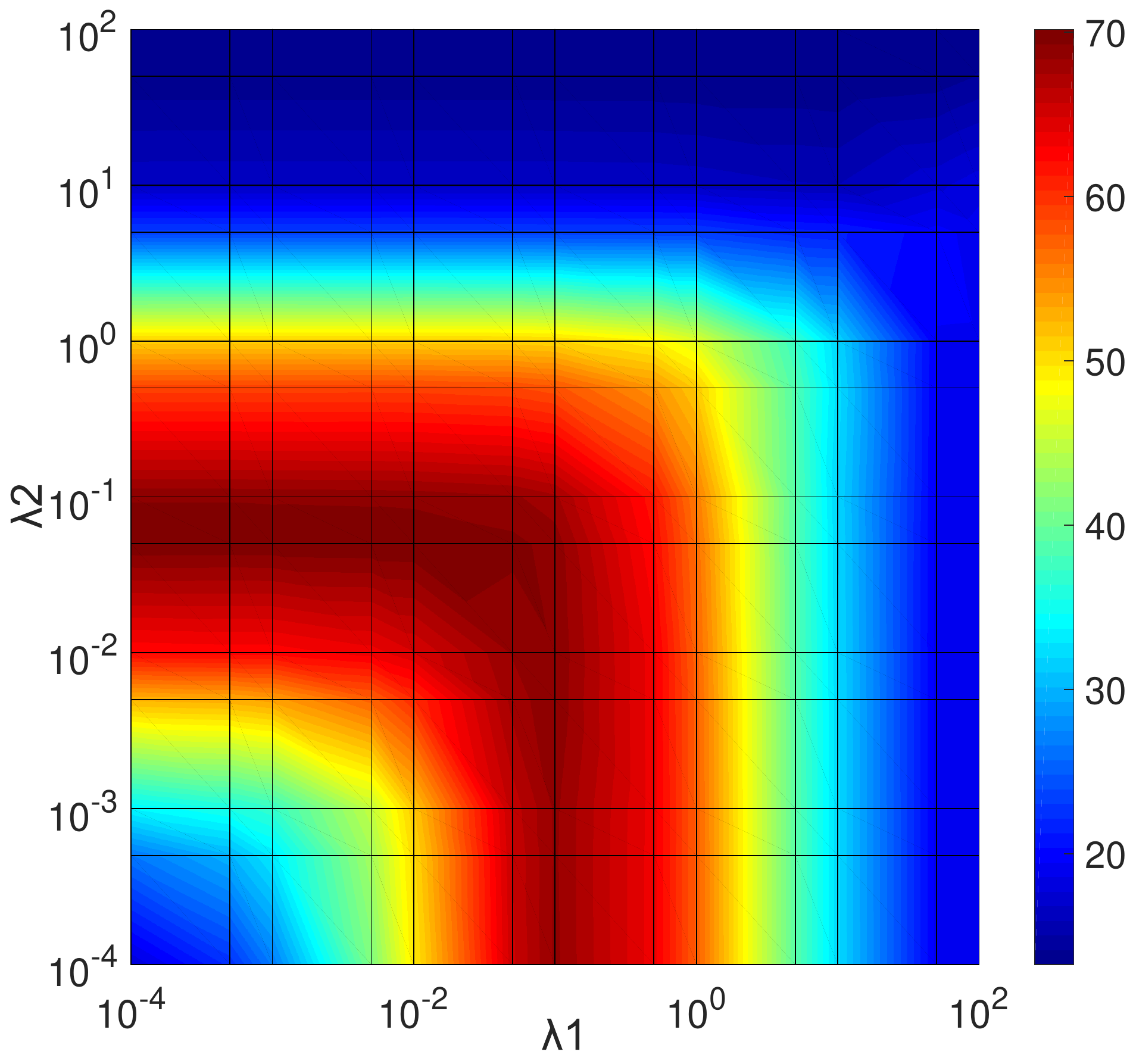}
		\subcaption{}
	\end{minipage}%
	\hspace{0.2cm}
	\begin{minipage}{3cm}
		\includegraphics[width=3.3cm]{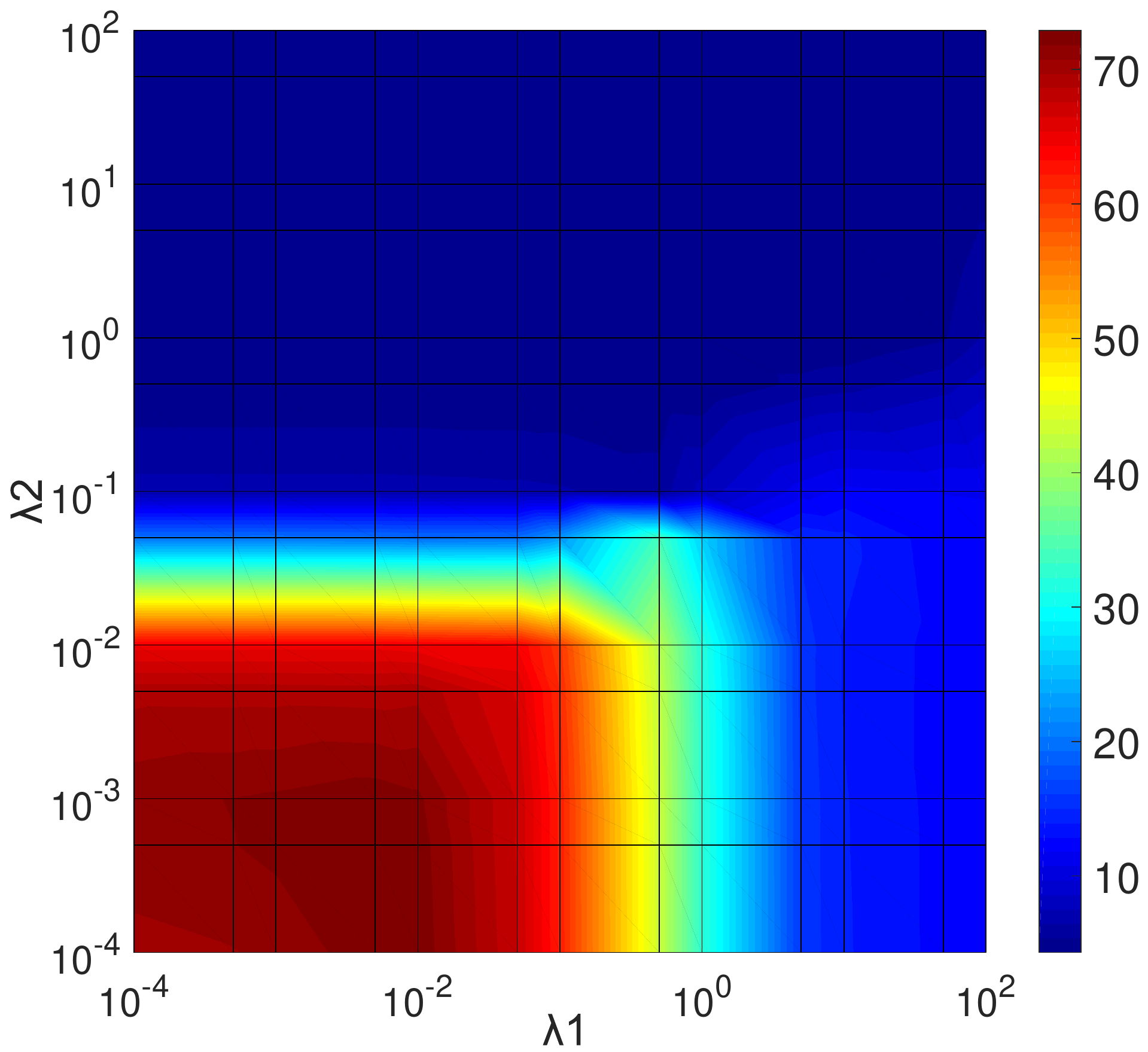}
		\subcaption{}
	\end{minipage}%
	\hspace{0.2cm}
	\begin{minipage}{3cm}
		\includegraphics[width=3.3cm]{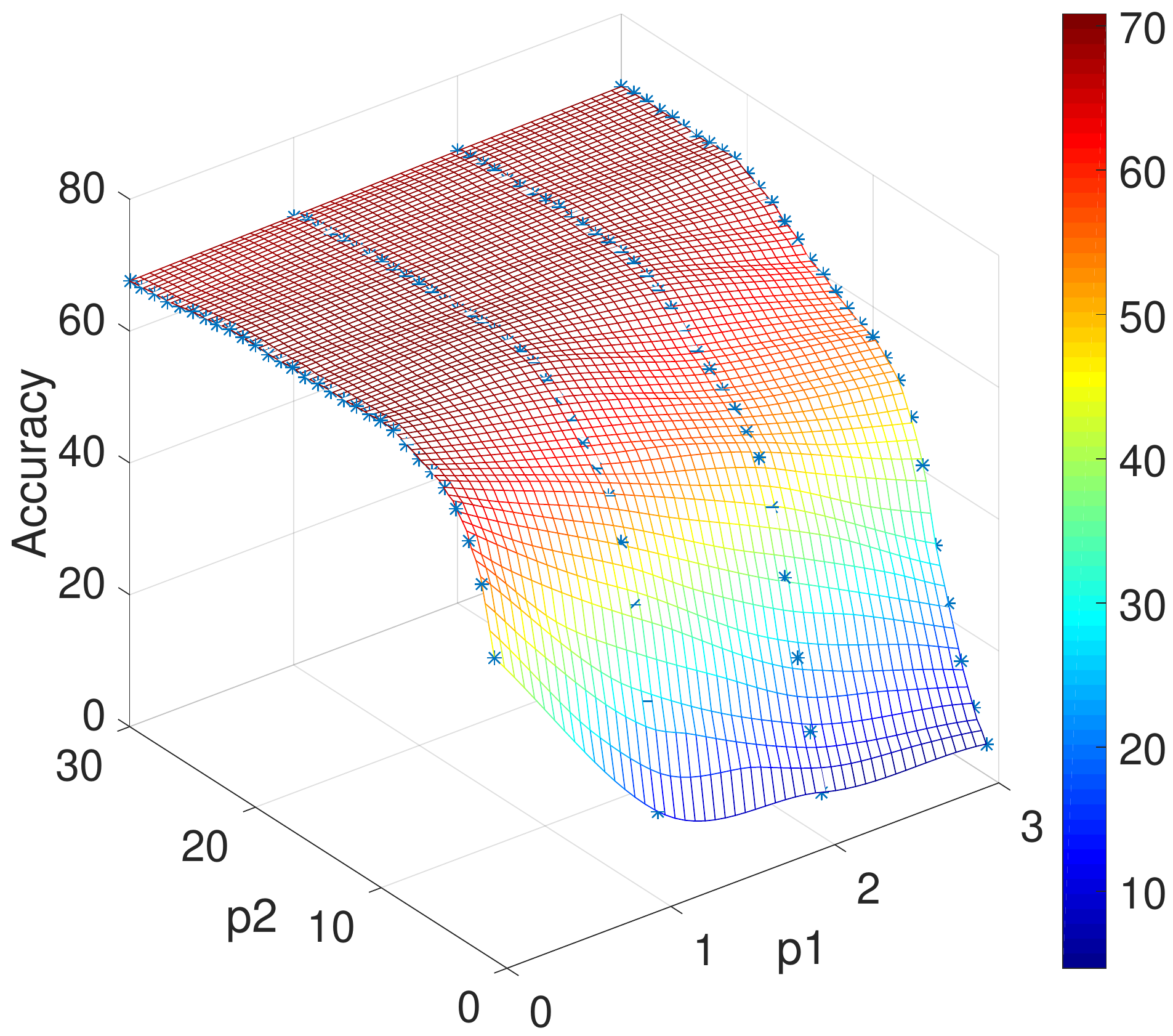}
		\subcaption{}
	\end{minipage}
	\hspace{0.2cm}
	\begin{minipage}{3cm}
		\includegraphics[width=3.3cm]{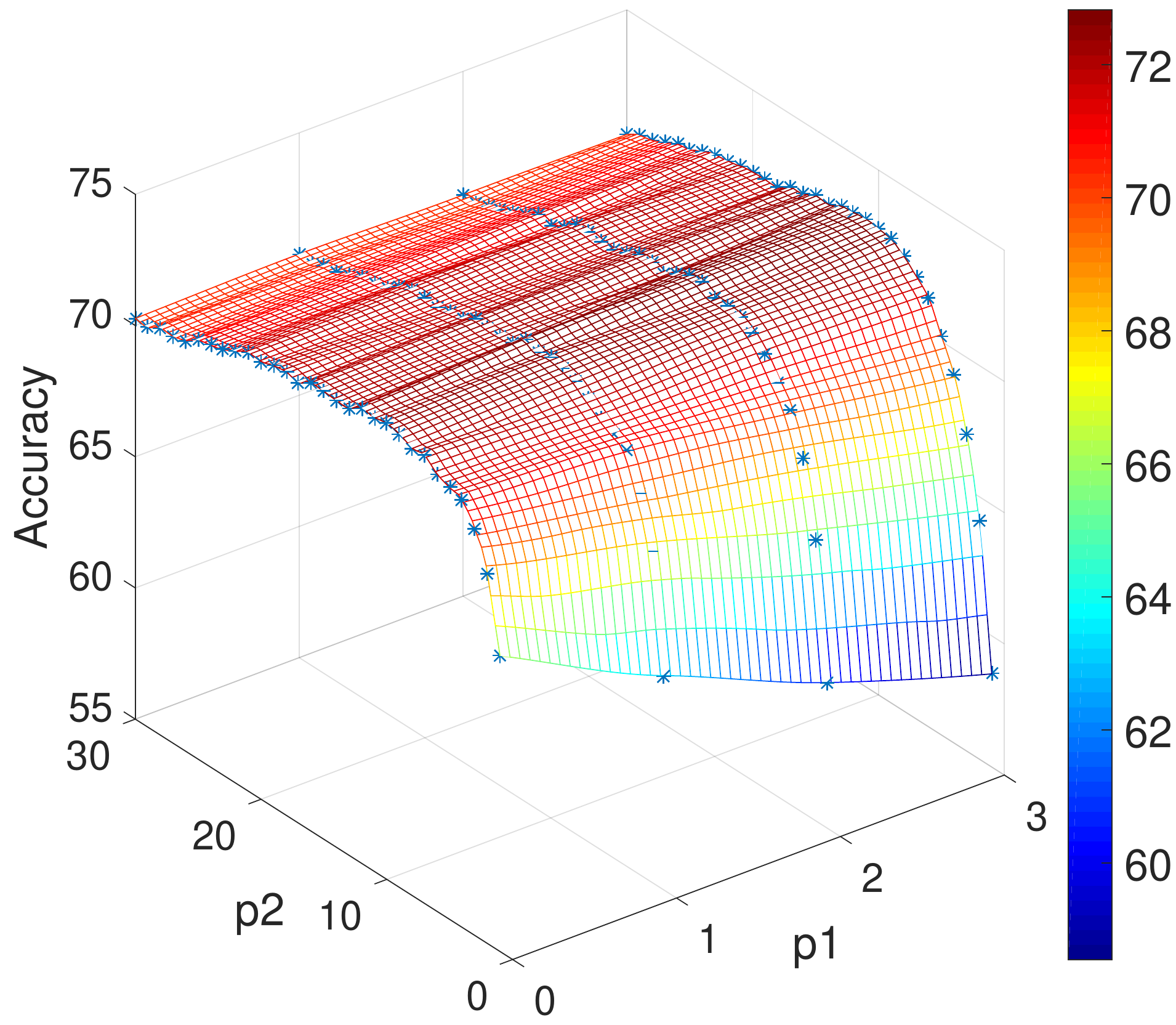}
		\subcaption{}
	\end{minipage}
	\hspace{0.2cm}
	\begin{minipage}{3cm}
		\includegraphics[width=3cm]{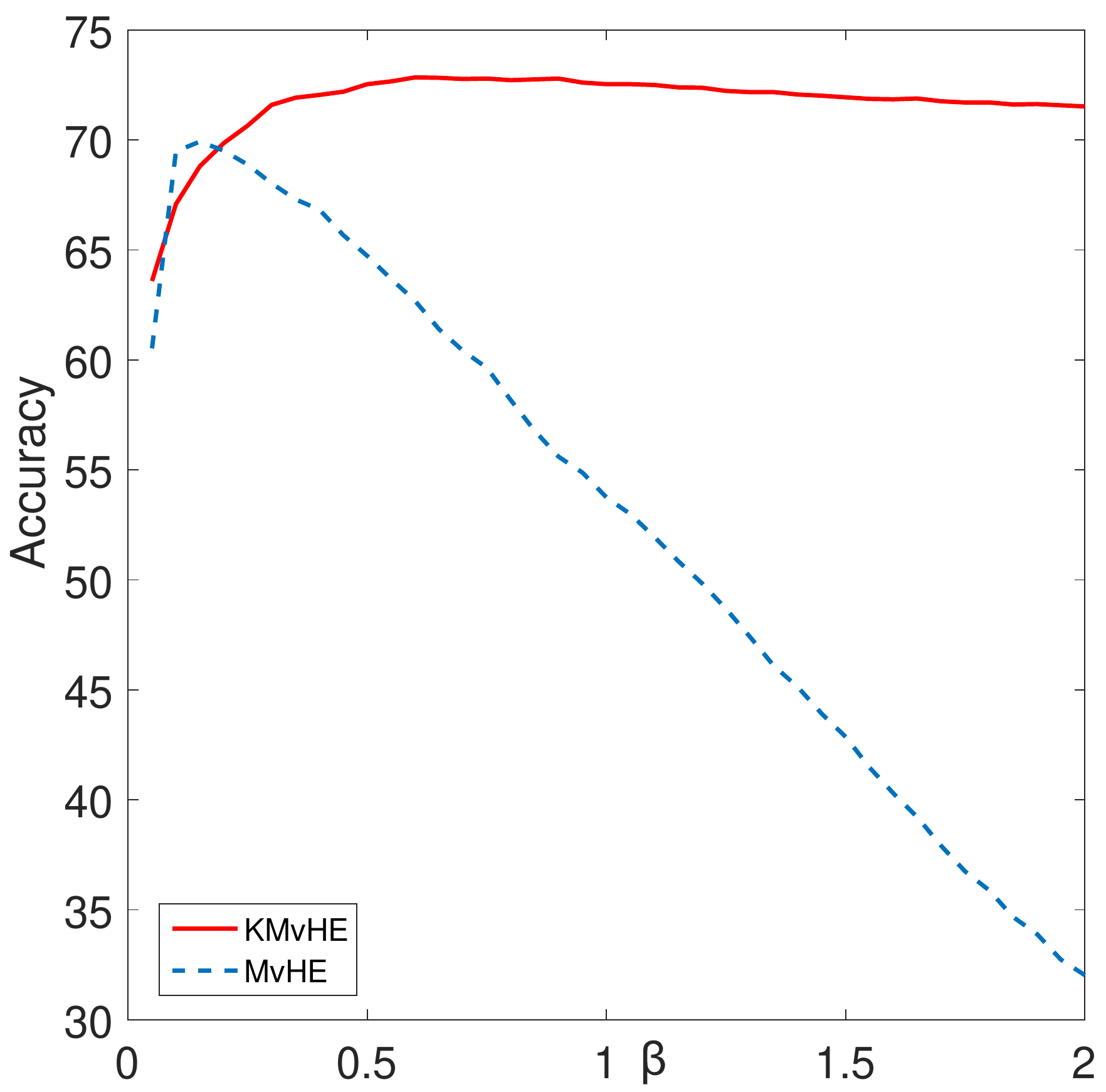}
		\subcaption{}
	\end{minipage}
	\caption{Classification accuracy of MvHE and KMvHE with different parameter setting in terms of mean accuracy (mACC) in case 5 of CMU PIE dataset: (a) and (b) show mACC with different $\lambda_{1}$ and $\lambda_{2}$; (c) and (d) show mACC with different $p_1$ and $p_2$; whereas (e) shows mACC with different $\beta$.\vspace{-0.5cm}}
	\label{fig:parameter_1_2}
\end{figure*}

\subsection{Parameter Sensitivity}
\label{sec:parameter_sensitivity}
We finally evalute the classification accuracy of our methods in case 5 of CMU PIE with different parameter values. To this end, we divide hyperparameters into three groups based on their correlations: (1) $\{\lambda_{1},\lambda_{2}\}$, (2) $\{p_1,p_2\}$, (3) $\beta$. When we test the effect of one group of parameters, other groups are set to default values, i.e., $\{\lambda_{1},\lambda_{2}\}=\{10^{-2}, 5\times{10^{-2}}\}$, $\{p_1,p_2\}=\{1, 15\}$, $\beta=0.1$ for MvHE and $\{\lambda_{1},\lambda_{2}\}=\{5\times{10^{-3}}, {10^{-3}}\}$, $\{p_1,p_2\}=\{3, 15\}$, $\beta=1.1$ for KMvHE.


From the results shown in Fig.~\ref{fig:parameter_1_2}, we can find that the performance of our methods is robust to their parameters in a reasonable range. Specifically, Figs.~\ref{fig:parameter_1_2}(a) and \ref{fig:parameter_1_2}(b) indicate that MvHE can achieve a high classification accuracy when $\lambda_{1}\in\left[\!10^{-4},~10^{-1}\!\right]$ and $\lambda_{2}\in\left[\!10^{-2},~10^{-1}\!\right]$ or $\lambda_{1}\in\left[\!10^{-2},~10^{-1}\!\right]$ and $\lambda_{2}\in\left[\!10^{-4},~10^{-1}\!\right]$, whereas KMvHE achieves a satisfactory accuracy when $\lambda_{1}\in\left[\!10^{-4},~5\!\times\!10^{-2}\!\right]$ and $\lambda_{2}\in\left[\!10^{-4},~10^{-2}\!\right]$. However, when the value of $\lambda_{1}$ and $\lambda_{2}$ are out of those ranges, the performance of our methods deteriorates dramatically. Figs.~\ref{fig:parameter_1_2}(c) and~\ref{fig:parameter_1_2}(d) indicate that our methods are robust to the variation of $p_1$, although they both favor a relatively large value of $p_2$. We suggest using cross validation to determine the value of $p_2$ in different applications. Finally, from Fig.~\ref{fig:parameter_1_2}(e), we can see that appropriate $\beta$ is significant for MvHE. If $\beta$ exceeds its optimal value, the classification accuracy decreases monotonically with the increase of $\beta$. By contrast, the performance of our KMvHE does not suffer from the variation of $\beta$. This is a nice property that can benefit real applications. 


\section{Conclusion}
\label{sec_conclusion}
In this paper, inspired by the \textit{Divide-and-Conquer} strategy, we present a novel Multi-view Hybrid Embedding~(MvHE) for cross-view classification. Extensive experiments are conducted on four real-world datasets. Compared with other state-of-the-art approaches, our method can effectively discover the nonlinear structure embedded in high-dimensional multi-view data, thus significantly improving classification accuracy. Moreover, our method also shows superior robustness against all other competitors in terms of different degrees of outliers and the selection of parameters. 

In future work, we will investigate methods to speed up MvHE. According to current results, the extra computational cost mainly comes from the calculation of the selection matrix (i.e., $S_i^k$ and $S_j^k$ in Eqs.~(\ref{defined_U}) and (\ref{defined_V})) in DLA~\cite{zhang2009patch}. Hence, one feasible solution to reduce the computational complexity is to speed up DLA. Some initial work has been started.




%

\appendices
\section{Matrix Form of \textit{LE-paired}} 
\label{appendixA}
We derive the matrix form of \textit{LE-paired} as follows.

Denote
\begin{align}
\bm{Y}_i\!=\!\left[\bm{y}_i^1,\bm{y}_i^2,\ldots,\bm{y}_i^m\right],
\end{align}
we have:
\begin{align}\label{paired_sol_1}
\mathcal{J}_1&=\sum_{i=1}^n\sum_{\substack{j=1\\j\neq{i}}}^n\sum_{k=1}^m\norm{\bm{y}_i^k-\bm{y}_j^k}_2^2=\sum_{i=1}^n\sum_{\substack{j=1\\j\neq{i}}}^n\norm{\bm{Y}_i-\bm{Y}_j}_F^2\nonumber\\
&=\sum_{i=1}^n\norm{\bm{Y}_i\!-\!\bm{Y}_1}_F^2\!+\!\norm{\bm{Y}_i\!-\!\bm{Y}_2}_F^2\!+\!\dots\!+\!\norm{\bm{Y}_i\!-\!\bm{Y}_n}_F^2\nonumber\\
&=\sum_{i=2}^n\bigg(\norm{\bm{Y}_1\!-\!\bm{Y}_i}_F^2\!+\!\dots\!+\!\norm{\bm{Y}_{n+1-i}\!-\!\bm{Y}_n}_F^2\!+\!\dots\!\nonumber\\
&\quad\quad\quad\quad+\!\norm{\bm{Y}_{n+2-i}\!-\!\bm{Y}_1}_F^2\!+\!\dots\!+\!\norm{\bm{Y}_n\!-\!\bm{Y}_{i-1}}_F^2\bigg)\nonumber\\
&=\sum_{i=2}^n\norm{\left[\bm{Y}_1,\bm{Y}_2,...,\bm{Y}_n\right]\!-\!\left[\bm{Y}_i,\dots,\bm{Y}_n,\bm{Y}_1,\dots,\bm{Y}_{i-1}\right]}_F^2.
\end{align}

Let $\bm{Y}^{\left(i\right)}\!=\!\left[\bm{Y}_i,\dots,\bm{Y}_n,\bm{Y}_1,...\bm{Y}_{i-1}\right]$, we have $\bm{Y}^{\left(i\right)}=\bm{Y}\bm{J}_i$, and
\begin{align}
\bm{J}_i=\begin{bmatrix}
&\textbf{I}_{\left(\left(i-1\right)*{m}\right)}\\
\textbf{I}_{\left(\left(n-i+1\right)*{m}\right)}&\\
\end{bmatrix}.\nonumber
\end{align}

Therefore, Eq.~(\ref{paired_sol_1}) can be rewritten as:
\begin{align}\label{paired_sol_2}
\mathcal{J}_1=&\norm{\bm{Y}\!-\!\bm{Y}^{\left(2\right)}}_F^2\!+\!\norm{\bm{Y}\!-\!\bm{Y}^{\left(3\right)}}_F^2\!+\!\ldots\!+\!\norm{\bm{Y}\!-\!\bm{Y}^{\left(n\right)}}_F^2\nonumber\\
=&\sum_{i=2}^n\norm{\bm{Y}\!-\!\bm{Y}\bm{J}_i}_F^2\!=\!\sum_{i=2}^n\mathrm{tr}\left(\bm{Y}\left(\textbf{I}\!-\!\bm{J}_i\right)\left(\textbf{I}\!-\!\bm{J}_i\right)^\mathrm{T}\bm{Y}^\mathrm{T}\right)\nonumber\\
=&\mathrm{tr}\left(\bm{W}^\mathrm{T}\bm{X}\bm{J}\bm{X}^\mathrm{T}\bm{W}\right),
\end{align}
where 
\begin{align}
   \bm{J}=\sum_{i=2}^n\left(\textbf{I}-\bm{J}_i\right)\left(\textbf{I}-\bm{J}_i\right)^\mathrm{T}. 
\end{align} 

\section{Matrix Form of \textit{LDE-intra}} 
\label{appendixB}
We derive the matrix form of \textit{LDE-intra} as follows.
\begin{align}\label{intra_sol_1}
\mathcal{J}_2^{'}=\sum_{i=1}^n\sum_{k=1}^m\left(\sum_{s=1}^{p_1}\norm{\bm{y}_i^k-\left(\bm{y}_i^k\right)_s}_2^2-\beta\sum_{t=1}^{p_2}\norm{\bm{y}_i^k-\left(\bm{y}_i^k\right)^t}_2^2\right).
\end{align}

To make the derivation easier to follow, we first consider the term inside the bracket. We define $\bm{\theta}_i^k$ as:
\begin{align}
    &\bm{\theta}_i^k=\left[\overbrace{1,\ldots,1,}^{p_1}\overbrace{-\beta,\ldots,-\beta}^{p_2}\right]^\mathrm{T},
\end{align}
then we have:
\begin{align}\label{intra_local_patch_sol}
&\sum_{s=1}^{p_1}\norm{\bm{y}_i^k\!-\!\left(\bm{y}_i^k\right)_s}_2^2\!-\!\beta\sum_{t=1}^{p_2}\norm{\bm{y}_i^k\!-\!\left(\bm{y}_i^k\right)^t}_2^2\nonumber\\
=&\sum_{s=1}^{p_1}\norm{\bm{y}_i^k\!-\!\left(\bm{y}_i^k\right)_s}_2^2\bm{\theta}_i^k\left(s\right)\!+\!\sum_{t=1}^{p_2}\norm{\bm{y}_i^k\!-\!\left(\bm{y}_i^k\right)^t}_2^2\bm{\theta}_i^k\left(t\!+\!p_1\right),\nonumber\\
\end{align}
where $\bm{\theta}_i^k\left(s\right)$ and $\bm{\theta}_i^k\left(t+p_1\right)$ refer to the $s$-th and the $(t+p_1)$-th elements of $\bm{\theta}_i^k$ respectively.

Denote $F_i^k$ the set of indices for the local patch of $\bm{x}_i^k$:
\begin{align}\label{intra_indices}
F_i^k=\{*_i^k, \left(*_i^k\right)_1,\dots,\left(*_i^k\right)_{p_1},\left(*_i^k\right)^1,\dots,\left(*_i^k\right)^{p_2}\}.
\end{align}

Combining Eq.~(\ref{intra_local_patch_sol}) with Eq.~(\ref{intra_indices}), we have:
\begin{align}
&\sum_{s=1}^{p_1+p_2}\norm{\bm{y}F_i^k\left(1\right)-\bm{y}F_i^k\left(s+1\right)}_2^2\bm{\theta}_i^k\left(s\right)\nonumber\\
=&\mathrm{tr}\left(\bm{Y}_i^k\begin{bmatrix}
-\bm{e}_{p_1+p_2}^\mathrm{T}\\
\textbf{I}_{p_1+p_2}
\end{bmatrix}\textrm{diag}\left(\bm{\theta}_i^k\right)\begin{bmatrix}
-\bm{e}_{p_1+p_2}&\bm{\textbf{I}}_{p_1+p_2}
\end{bmatrix}\left(\bm{Y}_i^k\right)^\mathrm{T}\right)\nonumber\\
=&\mathrm{tr}\left(\bm{Y}_i^k\bm{L}_i^k\left(\bm{Y}_i^k\right)^\mathrm{T}\right),
\end{align}
where $\bm{Y}_i^k\!=\!\left[\bm{y}_i^k,\left(\bm{y}_i^k\right)_1,\dots,\left(\bm{y}_i^k\right)_{p_1},\left(\bm{y}_i^k\right)^1,\dots,\left(\bm{y}_i^k\right)^{p_2}\right]$ and $\bm{L}_i^k$ can be written as:
\begin{align}
\bm{L}_i^k = \begin{bmatrix}-\bm{e}_{p_1+p_2}^\mathrm{T}\\
\textbf{I}_{p_1+p_2}
\end{bmatrix}\textrm{diag}\left(\bm{\theta}_i^k\right)\begin{bmatrix}
-\bm{e}_{p_1+p_2}&\textbf{I}_{p_1+p_2}
\end{bmatrix}.
\end{align}

We unify each $\bm{Y}_i^k$ as a whole by assuming that the coordinate for $\bm{Y}_i^k\!=\!\left[\!\bm{y}_i^k,\left(\!\bm{y}_i^k\!\right)_1,\dots,\left(\!\bm{y}_i^k\!\right)_{p_1},\left(\!\bm{y}_i^k\!\right)^1,\dots,\left(\!\bm{y}_i^k\!\right)^{p_2}\!\right]$ is selected from the global coordinate $\bm{Y}\!=\!\left[\!\bm{y}_1^1,\ldots,\bm{y}_1^m,\bm{y}_2^1,\ldots,\bm{y}_2^m,\ldots,\bm{y}_n^1,\ldots,\bm{y}_n^m\!\right]$, such that:
\begin{align}
&\bm{Y}_i^k=[\bm{Y}_1,\bm{Y}_2,\dots,\bm{Y}_n]\bm{S}_i^k=\bm{Y}\bm{S}_i^k,\nonumber\\
&\left(\bm{S}_i^k\right)_{pq}=\left\{
\begin{aligned}
1, & \quad \textrm{if}\quad p=F_i^k\left(q\right); \\
0, & \quad \textrm{otherwise},
\end{aligned}
\right.
\end{align}
where $\bm{S}_i^k\in\mathbb{R}^{\left(n\times{m}\right)\times\left(p_1+p_2+1\right)}$ is a selection matrix.
Considering the whole alignment module, we have:
\begin{align}\label{intra_sol_2}
&\sum_{i=1}^n\sum_{k=1}^m\left(\sum_{s=1}^{p_1}\norm{\bm{y}_i^k\!-\!\left(\bm{y}_i^k\right)_s}_2^2\!-\!\beta\sum_{t=1}^{p_2}\norm{\bm{y}_i^k\!-\!\left(\bm{y}_i^k\right)^t}_2^2\right)\nonumber\\
=&\sum_{i=1}^n\sum_{k=1}^m\mathrm{tr}\left(\!\bm{Y}_i^k\bm{L}_i^k\left(\bm{Y}_i^k\right)^\mathrm{T}\!\right)\!=\!\sum_{i=1}^n\sum_{k=1}^m\mathrm{tr}\left(\!\bm{Y}\bm{S}_i^k\bm{L}_i^k\left(\!\bm{S}_i^k\!\right)^\mathrm{T}\bm{Y}^\mathrm{T}\!\right)\nonumber\\
=&\mathrm{tr}\left(\bm{Y}\bm{U}\bm{Y}^\mathrm{T}\right)\!=\!\mathrm{tr}\left(\bm{W}^\mathrm{T}\bm{X}\bm{U}\bm{X}^\mathrm{T}\bm{W}\right),
\end{align}
where 
\begin{align}
\bm{U}=\sum_{i=1}^n\sum_{k=1}^m\left(\bm{S}_i^k\bm{L}_i^k\left(\bm{S}_i^k\right)^\mathrm{T}\right).
\end{align}

\section*{Acknowledgment}
This work was supported partially by the Key Science and Technology of Shenzhen~(No. CXZZ20150814155434903), in part by the Key Program for International S\&T Cooperation Projects of China~(No. 2016YFE0121200), in part by the Key Science and Technology Innovation Program of Hubei Province~(No. 2017AAA017), in part by the Macau Science and Technology Development Fund~(FDCT 024/2015/AMJ), in part by the Special Projects for Technology Innovation of Hubei Province~(No. 2018ACA135), in part by the National Natural Science Foundation of China~(No. 61571205 and 61772220), in part by the NSFC-Key Project of General Technology Fundamental Research United Fund~(No. U1736211).

\bibliographystyle{IEEEtran}
\bibliography{reference}

\end{document}